\PassOptionsToPackage{unicode}{hyperref}
\PassOptionsToPackage{hyphens}{url}
\documentclass[
]{article}
\usepackage{xcolor}
\usepackage{amsmath,amssymb}
\usepackage{iftex}
\ifPDFTeX
  \usepackage[T1]{fontenc}
  \usepackage[utf8]{inputenc}
  \usepackage{textcomp} 
\else 
  \usepackage{unicode-math} 
  \defaultfontfeatures{Scale=MatchLowercase}
  \defaultfontfeatures[\rmfamily]{Ligatures=TeX,Scale=1}
\fi
\usepackage{lmodern}
\ifPDFTeX\else
\fi
\IfFileExists{upquote.sty}{\usepackage{upquote}}{}
\IfFileExists{microtype.sty}{
  \usepackage[]{microtype}
  \UseMicrotypeSet[protrusion]{basicmath} 
}{}
\makeatletter
\@ifundefined{KOMAClassName}{
  \IfFileExists{parskip.sty}{%
    \usepackage{parskip}
  }{
    \setlength{\parindent}{0pt}
    \setlength{\parskip}{6pt plus 2pt minus 1pt}}
}{
  \KOMAoptions{parskip=half}}
\makeatother
\usepackage{longtable,booktabs,array}
\usepackage{calc} 
\usepackage{etoolbox}
\makeatletter
\patchcmd\longtable{\par}{\if@noskipsec\mbox{}\fi\par}{}{}
\makeatother
\IfFileExists{footnotehyper.sty}{\usepackage{footnotehyper}}{\usepackage{footnote}}
\makesavenoteenv{longtable}
\usepackage{graphicx}
\makeatletter
\newsavebox\pandoc@box
\newcommand*\pandocbounded[1]{
  \sbox\pandoc@box{#1}%
  \Gscale@div\@tempa{\textheight}{\dimexpr\ht\pandoc@box+\dp\pandoc@box\relax}%
  \Gscale@div\@tempb{\linewidth}{\wd\pandoc@box}%
  \ifdim\@tempb\p@<\@tempa\p@\let\@tempa\@tempb\fi
  \ifdim\@tempa\p@<\p@\scalebox{\@tempa}{\usebox\pandoc@box}%
  \else\usebox{\pandoc@box}%
  \fi%
}
\def\fps@figure{htbp}
\makeatother
\providecommand{\tightlist}{%
  \setlength{\itemsep}{0pt}\setlength{\parskip}{0pt}}
\usepackage[letterpaper,margin=1in]{geometry}
\renewcommand{\_}{\textunderscore\allowbreak}
\usepackage{newunicodechar}
\newunicodechar{≈}{\ensuremath{\approx}}
\newunicodechar{⇒}{\ensuremath{\Rightarrow}}
\newunicodechar{→}{\ensuremath{\rightarrow}}
\newunicodechar{×}{\ensuremath{\times}}
\newunicodechar{≪}{\ensuremath{\ll}}
\newunicodechar{≤}{\ensuremath{\leq}}
\newunicodechar{≥}{\ensuremath{\geq}}
\newunicodechar{∝}{\ensuremath{\propto}}
\newunicodechar{⊆}{\ensuremath{\subseteq}}
\newunicodechar{∀}{\ensuremath{\forall}}
\newunicodechar{✓}{\ensuremath{\checkmark}}
\newunicodechar{✅}{\ensuremath{\checkmark}}
\newunicodechar{↗}{\ensuremath{\nearrow}}
\newunicodechar{⊕}{\ensuremath{\oplus}}
\newunicodechar{⊗}{\ensuremath{\otimes}}
\newunicodechar{ℓ}{\ensuremath{\ell}}
\usepackage{xeCJK}
\usepackage{bookmark}
\IfFileExists{xurl.sty}{\usepackage{xurl}}{} 
\makeatletter
\@ifundefined{xmpquote}{}{}
\makeatother
\hypersetup{
  pdftitle={Certified World Models: Predictability Across Configuration{,} Horizon{,} and Resolution},
  pdfauthor={Hongbo Wang},
  hidelinks,
  pdfcreator={LaTeX via pandoc}}

\title{Certified World Models: Predictability Across Configuration,
Horizon, and Resolution}
\usepackage{etoolbox}
\makeatletter
\providecommand{\subtitle}[1]{
  \apptocmd{\@title}{\par {\large #1 \par}}{}{}
}
\makeatother
\subtitle{Scale buys interpolation; structure buys a certificate.}
\author{Hongbo Wang \\
  \small Department of Mathematics, Stony Brook University, Stony Brook, NY 11794, USA}
\date{}

\begin{document}
\maketitle

\subsection{Abstract}\label{abstract}

\emph{Scale buys interpolation; structure buys certifiable transfer.}
Scaling a world model lowers its average error but issues no guarantee
about any specific unseen situation. For \textbf{equivariant} latent
world models we give a different \emph{kind} of result --- a
\textbf{predictability certificate}: a provable, computable region the
model is guaranteed to transfer across, spanning \textbf{configuration},
\textbf{horizon}, and \textbf{resolution} at once. Under exact
equivariance with an orthogonal latent the whole-pipeline error is
\emph{invariant} over the exponentially large monoid
\(\langle S\rangle\) generated by \(k\) primitive symmetries, and
checking the \(k\) generators certifies all of it (Theorem A). This
orbit-flatness over equivariant targets \textbf{characterizes
equivariance at the function level} (Lemma 2), so an unconstrained
architecture cannot certify the property by construction --- it must
learn the property and then be separately verified.

Approximate orbit-transfer defects propagate along the finite-time
Lyapunov spectrum (Theorem B, two-sided with Proposition 6): expanding
channels yield the logarithmic horizon
\(T_j(\epsilon)\sim\log(1/\epsilon)/\lambda_j\), neutral channels
accumulate recurrent defect linearly, and contracting channels settle to
a bounded nonzero floor. Exactly conserved charge \emph{values} are
certified to all horizons; under a one-step conservation defect \(\eta\)
they drift at most as \(T\eta\) (Propositions 4--5). We confirm all
three axes at small scale: \(\mathbb{Z}_2^6\) certifies all \(64\)
compositions to machine precision; a learned predictor recovers a
planted exponent to \(0.4\%\); and on a \textbf{\(40\)-dimensional}
learned model a \(\mathbb{Z}_N\)-equivariant network recovers the
\emph{full} Lyapunov spectrum (\(R^2{=}0.98\)--\(0.99\)) where a dense
\textbf{and a same-trained recurrent} model of equal data fail
(\(R^2{<}0\)) --- so it is structure, not scale or recurrence, that
recovers a high-dimensional horizon.

Finally we make \emph{``Certified''} literal: a cone/adapted-metric
certificate (Theorem B\({}^{\prime}\)) reads a \textbf{sound, a-priori}
certified horizon off the \emph{learned model's own} Jacobian --- tight
exactly on uniformly-hyperbolic dynamics and self-abstaining elsewhere.
Because the certificate is \emph{faithful} --- the spectrum it reads is
the true one --- it is also \textbf{actionable}: it improves a budgeted
re-observation decision. Holding the forecaster fixed on the \(40\)-D
system, timing sparse re-observations by the equivariant model's
certified horizon meets a \textbf{fixed sensing budget} where the
non-equivariant model's \(\sim3\times\)-inflated certificate
over-observes and starves it (median \(10\%\) vs \(63\%\)
forecast-violation, \(20/20\) seeds; replicated on a second
\(\mathbb{Z}_N\) system; a \(c\times\)-inflated certificate provably
needs \(c\times\) the budget --- Proposition 9). A dense certificate
closes the gap only by \emph{spending a calibration set}; structure buys
the trustworthy certificate for free (a full-spectrum allocation
variant, reported honestly, does \textbf{not} win). For \textbf{public,
non-equivariant} world models the tangent spectrum gives a
\textbf{training-free candidate} horizon, paired with a held-out
divergence cross-check that certifies when the spectral readout is in
scope and abstains or corrects it when the learned loop over-promises
(TD-MPC2 zoo, LeWM, V-JEPA 2-AC; E13--E16). In a one-checkpoint-per-size
TD-MPC2 multitask audit we observe \textbf{no monotone improvement} of
horizon calibration with parameter count. The result is a single
\textbf{runnable} criterion (Algorithm 1) for \emph{what} an equivariant
world model can certifiably predict, and a structural reading of
\emph{why} exactly-conserved celestial mechanics is forecastable for
millennia while weather is not.

\begin{center}\rule{0.5\linewidth}{0.5pt}\end{center}

\subsection{1. Introduction}\label{introduction}

Two debates organize modern world-model research. The first is
\emph{generation vs.~abstraction}: pixel-space simulators (diffusion and
video models) against latent predictive models that predict
representations rather than pixels. The second is \emph{scale
vs.~structure}: the view that brute-force scaling eventually wins,
against the geometric-learning view that symmetry priors buy sample
efficiency. Both are usually run as \textbf{degree} contests --- who
needs fewer pixels, who needs less data.

We argue the more useful question is a \textbf{kind} contest. Scaling
produces a model with low \emph{average} error on the data distribution,
but it cannot certify anything about a \emph{named, specific} situation
the data did not cover. Structure can. For an equivariant world model we
can write down, in advance and without further data, the exact set of
situations on which the model's behaviour is \textbf{guaranteed} --- a
\emph{predictability certificate}. This reframes ``is structure worth
it?'' from a benchmark race into a question about \emph{what kind of
statement each approach can make}.

The certificate has three orthogonal axes:

\begin{itemize}
\tightlist
\item
  \textbf{Configuration} \(w\in\langle S\rangle\): the model generalizes
  across the exponentially large monoid generated by \(k\) primitive
  symmetries \(S=\{g_1,\dots,g_k\}\), and we \emph{certify the whole
  monoid by checking only the \(k\) generators}.
\item
  \textbf{Horizon} \(T\): how far into the future the guarantee reaches.
\item
  \textbf{Resolution} \(\epsilon\): at what level of detail the
  guarantee holds.
\end{itemize}

Our contributions are:

\begin{enumerate}
\def\labelenumi{\arabic{enumi}.}
\item
  \textbf{A three-axis master theorem} (§3): composition closure (\(k\)
  checks \(\Rightarrow\) an exponential set), an exact certificate under
  exact equivariance (Theorem A) together with its \textbf{converse}
  (Lemma 2: the certificate is \emph{equivalent} to equivariance, so an
  unconstrained architecture cannot certify it by construction), and a
  spectral degradation law
  \(T_j(\epsilon)\sim\log(1/\epsilon)/\lambda_j\) (Theorem B) ---
  \textbf{tight}, with a matching lower bound (Proposition 6:
  \emph{approximate} equivariance is horizon-limited, so only exact
  structure or conservation reaches an unbounded horizon) and a
  \textbf{scope characterization} (Proposition 7: the local-spectrum
  horizon governs a \emph{learned} model on spectrally non-degenerate
  dynamics, \(\lambda_1>0\), and is vacuous on near-neutral dynamics;
  and \textbf{Proposition 8}: the learned model \emph{recovers} the
  chaos rate because the certified horizon is finite and finite-horizon
  Lyapunov exponents are \(C^1\)-continuous --- a \(O(\delta)\)
  model-fidelity bias, not the shadowing-transfer the asymptotic
  exponent forbids; and \textbf{Theorem B\({}^{\prime}\)}: a
  cone/adapted-metric certificate makes the \emph{``Certified''} literal
  --- a \textbf{sound, a-priori} certified horizon read off the
  \emph{learned model's own} Jacobian field with no access to the true
  dynamics, \textbf{tight exactly on uniformly-hyperbolic dynamics} and
  \textbf{self-abstaining} elsewhere via its cone-margin diagnostic) ---
  whose certified region is the coarse-invariant-slow-low-\(|w|\)
  corner, with a \textbf{quantitative scale-vs-structure separation}
  (§3.3: structure certifies the whole \(\epsilon\)-independent orbit;
  the best \(L\)-Lipschitz non-equivariant learner certifies only an
  \(\epsilon/L\)-tube around its data). Figure 1 is the whole picture at
  a glance.
\item
  \textbf{The Noether hinge} (§4): the bridge --- that the
  group-invariant/equivariant channels are the dynamically slow
  (long-horizon-certifiable) ones --- linking the configuration axis to
  the horizon axis. A representation-theoretic \textbf{placement
  principle} (Proposition 4) proves \emph{which} isotypic block must
  carry each conserved charge and why \(3\)D angular momentum is
  recoverable only at a unique degree-2 cross product;
  \textbf{Proposition 5} proves the \emph{forward} direction
  (\emph{conserved \(\Rightarrow\) slow}: a charge conserved to one-step
  defect \(\eta\) has prediction error \(\le T\eta\) --- linear, never
  the exponential \(e^{\lambda T}\) of a chaotic channel --- so it is
  certified to all horizons only at \(\eta{=}0\), and to a finite-time
  linear budget otherwise). What remains measured is the
  dynamical-symmetry hypothesis and the size of \(\eta\).
\item
  \textbf{Empirical confirmation of all three axes at small scale} (§5).
  The headline contrast --- \emph{scale buys interpolation; structure
  buys a certificate} --- is validated on \textbf{real physics-engine
  contact dynamics} (PushT and the \textbf{standard MuJoCo FetchPush}
  benchmark). Experiment 16 is the cleanest cut: the equivariant world
  model is \emph{exactly} orbit-flat while a \(7\times\)-larger baseline
  degrades by orders of magnitude out of the training orientation yet
  interpolates it competitively --- so the certificate holds for a
  \emph{learned} model of dynamics we did not design. We then lift it
  along five directions:

  \begin{itemize}
  \tightlist
  \item
    \textbf{Closed loop.} Run through a planner, the certificate becomes
    \emph{task} competence --- orbit-invariant pose control where a
    scaled baseline degrades. Experiment 17 lifts this to FetchPush,
    where the \emph{entire planner} (equivariant WM \(+\) equivariant
    goal-readout \(+\) \(G\)-equivariant CEM) is provably
    \(\mathrm{SO}(2)\)-equivariant and its plan orbit-flat, while the
    baseline planner degrades \(4\)--\(10\times\).
  \item
    \textbf{Non-abelian group.} A lift from the circle to the
    \textbf{non-abelian \(\mathrm{SO}(3)\)} on 3D point clouds --- the
    certificate is not \(\mathrm{SO}(2)\)-specific.
  \item
    \textbf{Raw pixels} (Experiment 13). \emph{Frame averaging} makes
    the exact certificate \textbf{accuracy-neutral}: an equivariant
    pixel model matches an unconstrained CNN and is horizon-stable, so
    the prior costs nothing.
  \item
    \textbf{Genuine chaos} (Lorenz, Experiment 14). The \textbf{horizon
    law} lifts to a learned model of genuinely chaotic dynamics: the
    learned model's Lyapunov exponent, read as the certified-horizon
    staircase slope, \emph{matches} the true exponent to
    \(1\text{–}8\%\) (instantiating Proposition 7(a)).
  \item
    \textbf{High dimension} (Experiment 18, \(40\)-D Lorenz-96). Here
    the configuration axis \emph{helps the horizon axis}: a
    \(\mathbb{Z}_N\)-equivariant model recovers the full \(40\)-D
    Lyapunov spectrum (\(R^2{=}0.98\text{–}0.99\)) --- hence the
    per-channel certified horizons --- while a dense model of equal data
    fails (\(R^2{<}0\)), as does a same-trained \emph{recurrent} GRU
    (Step 77). This pins the separation to \emph{structure}, not
    recurrence or training: a recurrent model's hidden Lyapunov modes
    violate the conditional-Lyapunov condition at high \(N\), whereas a
    Markov model's Jacobian is exactly \(N\times N\).
  \end{itemize}

  \textbf{Experiment 19} then brings \emph{both} axes onto \textbf{one
  controllable chaotic system} (controlled Lorenz-96) and lets the
  certificate \textbf{change a decision}: an equivariant planner gives
  machine-precision orbit-flat control on genuine chaos (configuration,
  \(8\times10^{-16}\)), while the certified horizon --- read off the
  model's own spectrum --- predicts when an open-loop forecast expires
  and drives an \textbf{active re-observation} schedule that sits on the
  efficient accuracy-vs-observation-cost frontier \emph{untuned} (the
  decision the certificate earns; short-horizon control, honestly, it
  does not). The pattern lifts seed-for-seed to a \(\mathbb{Z}_N\)
  pendulum ring and a \(\mathbb{Z}_2\) double pendulum --- a
  \textbf{class property} across two symmetry groups and high/low
  dimension.

  Finally, \textbf{Experiment 22} turns that high-dimensional spectrum
  recovery into a \emph{downstream consequence}: under a \textbf{fixed
  sensing budget} on \(40\)-D Lorenz-96, an agent that times
  re-observation by the equivariant model's certificate meets the budget
  (median \(10\%\) forecast-violation) while the \emph{same forecaster}
  timed by the non-equivariant model's inflated certificate
  over-observes and starves it (median \(63\%\); \(20/20\) seeds,
  margins \(+0.41\)--\(+0.61\)) --- a \emph{within-method} contrast
  isolating the certificate's \textbf{faithfulness} as the load-bearing
  property. Proposition 9 makes the cost a law (\(c\times\) inflation
  \(\Rightarrow c\times\) budget); a recalibrated dense certificate
  closes the gap only by spending a \textbf{calibration set} (the
  equivariant one is correct \emph{a-priori}, zero rollout data); the
  contrast replicates on a second \(\mathbb{Z}_N\) system (the pendulum
  ring, \(2/3\) seeds); and a full-spectrum \emph{allocation} variant is
  reported as an honest negative.
\end{enumerate}

We are explicit about scope (§7): this is a mechanism-and-theory paper
with \(1\)--\(2\)-GPU proof-of-principle, not a scaled benchmark, and
the hinge's lift to \emph{approximate} symmetry is open.

\begin{figure}
\centering
\pandocbounded{\includegraphics[keepaspectratio,alt={The predictability certificate at a glance. Left: in the configuration \textbackslash times horizon plane, an equivariant model certifies the entire generated monoid \textbackslash langle S\textbackslash rangle --- every composition, from k generator checks (Lemma 1) --- up to a horizon ceiling set by the predictor spectrum \textbackslash\{\textbackslash lambda\_j\textbackslash\} (Theorem B); an unconstrained (non-equivariant) architecture certifies only a small interpolation tube around its training set (\textbackslash sim\textbackslash epsilon/L, §3.3). Right: the horizon \textbackslash times resolution trade-off, priced by finite-time defect growth: expanding (\textbackslash lambda\textgreater0) channels shrink as the demanded resolution sharpens, T\_j(\textbackslash epsilon)\textbackslash sim\textbackslash log(1/\textbackslash epsilon)/\textbackslash lambda\_j (weather, \textbackslash simtwo weeks); contracting (\textbackslash lambda\textless0, \textbackslash zeta\textgreater0) channels hold a bounded floor \textbackslash zeta/(1-e\^{}\{\textbackslash lambda\}); neutral (\textbackslash lambda=0) channels a linear T\textbackslash zeta budget; and only an exactly-conserved charge (\textbackslash eta=0) is certified to all horizons (eclipses, millennia). Scale buys interpolation; structure buys certifiable transfer.}]{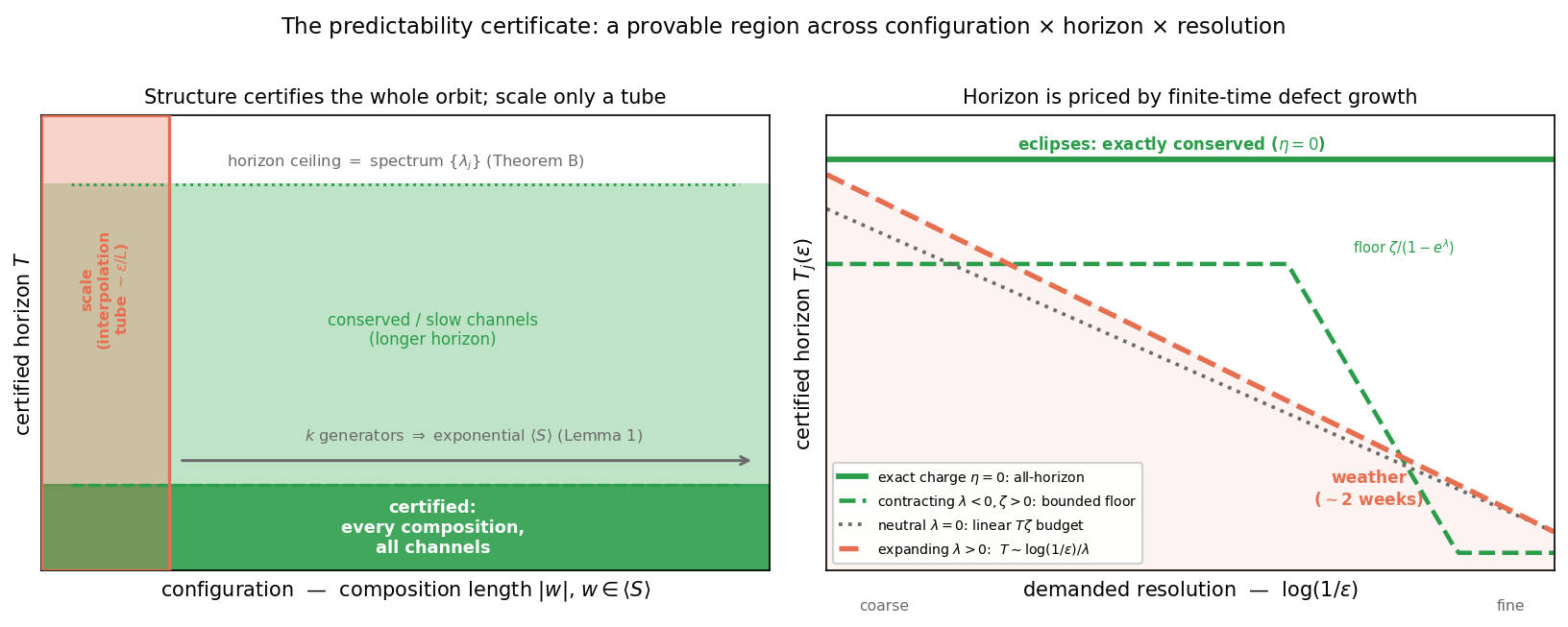}}
\caption{The predictability certificate at a glance. \textbf{Left:} in
the configuration \(\times\) horizon plane, an equivariant model
certifies the \emph{entire} generated monoid \(\langle S\rangle\) ---
every composition, from \(k\) generator checks (Lemma 1) --- up to a
horizon ceiling set by the predictor spectrum \(\{\lambda_j\}\) (Theorem
B); an unconstrained (non-equivariant) architecture certifies only a
small interpolation \emph{tube} around its training set
(\(\sim\epsilon/L\), §3.3). \textbf{Right:} the horizon \(\times\)
resolution trade-off, priced by finite-time defect growth: expanding
(\(\lambda>0\)) channels shrink as the demanded resolution sharpens,
\(T_j(\epsilon)\sim\log(1/\epsilon)/\lambda_j\) (weather, \(\sim\)two
weeks); contracting (\(\lambda<0\), \(\zeta>0\)) channels hold a bounded
floor \(\zeta/(1-e^{\lambda})\); neutral (\(\lambda=0\)) channels a
linear \(T\zeta\) budget; and only an exactly-conserved charge
(\(\eta=0\)) is certified to all horizons (eclipses, millennia).
\emph{Scale buys interpolation; structure buys certifiable transfer.}}
\end{figure}

\begin{center}\rule{0.5\linewidth}{0.5pt}\end{center}

\begin{figure}
\centering
\pandocbounded{\includegraphics[keepaspectratio,alt={The paper in one figure --- scale buys interpolation; structure buys a certified horizon. (a) Faithful: on 40-D Lorenz-96 the \textbackslash mathbb\{Z\}\_N-equivariant model recovers the full Lyapunov spectrum (R\^{}2\{=\}0.98) where an identically-trained dense model's is garbage (R\^{}2\{\textless\}0, \textbackslash lambda\_1 inflated \textbackslash sim3.4\textbackslash times) --- §5.16. (b) Priced: under a fixed sensing budget, re-observation timed by the faithful certificate meets the budget while the inflated certificate over-observes and starves it --- a c\textbackslash times-inflated certificate provably needs c\textbackslash times the budget (Proposition 9) and a certificate-free adaptive scheduler pays \textbackslash sim3\textbackslash times --- §5.20. (c) Real: the same training-free candidate read-out plus cross-check audits official TD-MPC2 checkpoints --- calibrated (ratio 0.94--1.02) where the latent loop is expansive, correctly abstaining where it contracts (Proposition 7) --- §5.21.}]{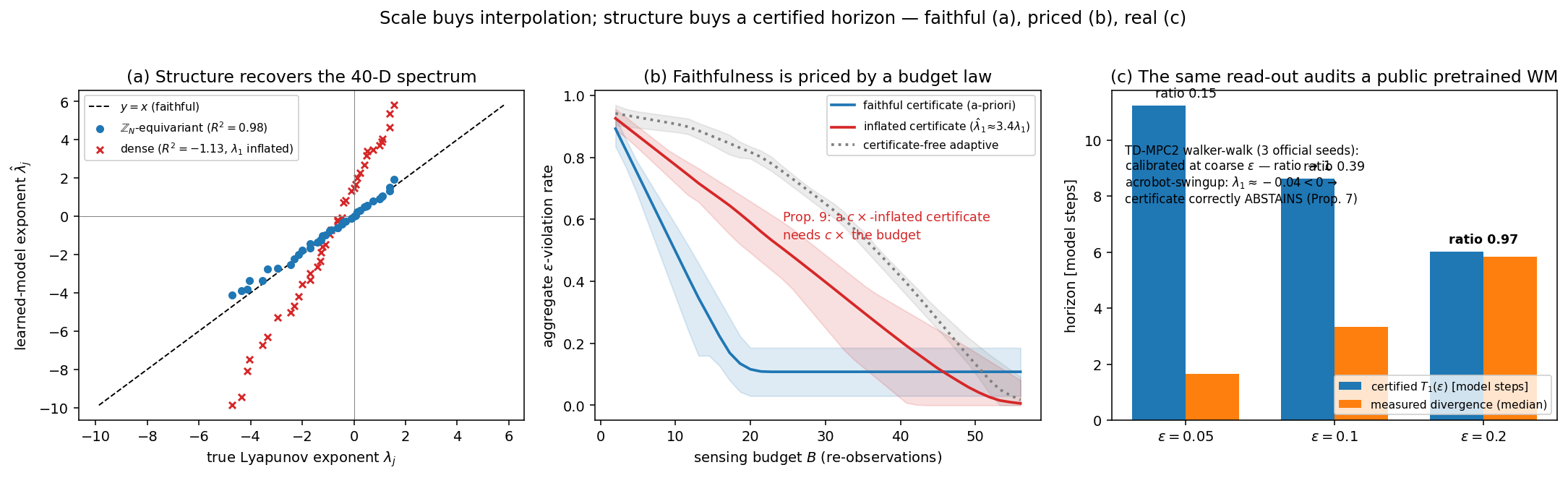}}
\caption{\textbf{The paper in one figure} --- \emph{scale buys
interpolation; structure buys a certified horizon.} \textbf{(a)
Faithful:} on \(40\)-D Lorenz-96 the \(\mathbb{Z}_N\)-equivariant model
recovers the full Lyapunov spectrum (\(R^2{=}0.98\)) where an
identically-trained dense model's is garbage (\(R^2{<}0\), \(\lambda_1\)
inflated \(\sim3.4\times\)) --- §5.16. \textbf{(b) Priced:} under a
fixed sensing budget, re-observation timed by the faithful certificate
meets the budget while the inflated certificate over-observes and
starves it --- a \(c\times\)-inflated certificate provably needs
\(c\times\) the budget (Proposition 9) and a certificate-free adaptive
scheduler pays \(\sim3\times\) --- §5.20. \textbf{(c) Real:} the same
training-free candidate read-out plus cross-check audits official
TD-MPC2 checkpoints --- calibrated (ratio \(0.94\)--\(1.02\)) where the
latent loop is expansive, correctly abstaining where it contracts
(Proposition 7) --- §5.21.}
\end{figure}

\subsection{2. Setup}\label{setup}

An encoder \(E:\mathcal X\to\mathcal Z\) maps states to latents and an
action-conditioned predictor
\(f:\mathcal Z\times\mathcal A\to\mathcal Z\) advances them; the true
environment transition is
\(\Phi:\mathcal X\times\mathcal A\to\mathcal X\). A finite generating
set \(S=\{g_1,\dots,g_k\}\) generates a monoid \(\langle S\rangle\)
whose elements --- words \(w=g_{i_1}\cdots g_{i_m}\) of length \(|w|=m\)
--- act on states by \(g\cdot x\), on latents by a representation
\(\rho(g)\), and on actions by \(\sigma(g)\). Write
\(f^{(T)}(z;\vec a)\) for the \(T\)-step latent rollout under actions
\(\vec a=(a_1,\dots,a_T)\), \(x^{\text{true}}_T\) for the \(T\)-step
true state, and

\[
\mathrm{Err}_T(x;\vec a)=\bigl\lVert f^{(T)}(E(x);\vec a)-E(x^{\text{true}}_T)\bigr\rVert
\]

for the whole-pipeline prediction error, where \(\lVert\cdot\rVert\) is
the Euclidean norm on \(\mathcal Z\) (the experiments use relMSE, its
square). We use the following assumptions, \textbf{each checkable on the
\(k\) generators}:

\begin{itemize}
\tightlist
\item
  \textbf{(A1) Encoder equivariance:} \(E(g_i\cdot x)=\rho(g_i)\,E(x)\).
\item
  \textbf{(A2) Predictor equivariance:}
  \(f(\rho(g_i)z,\,\sigma(g_i)a)=\rho(g_i)\,f(z,a)\).
\item
  \textbf{(A3) The group is a symmetry of the \emph{dynamics}:}
  \(\Phi(g_i\cdot x,\,\sigma(g_i)a)=g_i\cdot\Phi(x,a)\), so that rolling
  out the \emph{transformed} state equals transforming the
  \emph{rolled-out} state,
  \((g_i\cdot x)^{\text{true}}_T=g_i\cdot x^{\text{true}}_T\) under
  \(\sigma(g_i)\vec a\).
\item
  \textbf{(A4) \(\rho(g_i)\) orthogonal:} \(\rho(g_i)^\top\rho(g_i)=I\)
  (norm-preserving on \(\mathcal Z\)).
\item
  \textbf{(A5) Equivariant planner and invariant cost (closed-loop
  clause only):} the planner satisfies
  \(\pi(\rho(g_i)z,\rho(g_i)z^\star)=\sigma(g_i)\,\pi(z,z^\star)\) and
  the cost is \(\rho\)-invariant,
  \(J(\rho(g_i)z;\rho(g_i)z^\star)=J(z;z^\star)\). This is needed
  \emph{only} for the closed-loop \(J\) identity, not for the
  prediction-error identity. §5.6 instantiates such a planner in latent
  space; §5.9 instantiates one on \textbf{real PushT contact dynamics}
  --- an equivariant CEM-MPC (isotropic exploration covariance, a
  rotation-invariant disk action bound, and scene-covariant action
  noise) --- and confirms the resulting closed-loop task error is
  exactly orbit-invariant.
\end{itemize}

\begin{center}\rule{0.5\linewidth}{0.5pt}\end{center}

\subsection{3. The Predictability
Certificate}\label{the-predictability-certificate}

\subsubsection{3.1 The configuration axis (Theorem
A)}\label{the-configuration-axis-theorem-a}

\textbf{Lemma 1 (composition closure).} If (A1)--(A4) hold on each
generator \(g_i\in S\), they hold on every word
\(w\in\langle S\rangle\), and \(\rho\) restricted to
\(\langle S\rangle\) is an orthogonal-matrix \emph{monoid homomorphism}.

\emph{Proof.} Define \(\rho(w):=\rho(g_{i_1})\cdots\rho(g_{i_m})\). For
(A2), induct on \(m\):
\(f(\rho(w)z,\sigma(w)a)=f\!\big(\rho(g_{i_1})[\rho(g_{i_2}\!\cdots)z],\,\sigma(g_{i_1})[\sigma(g_{i_2}\!\cdots)a]\big)
=\rho(g_{i_1})f(\rho(g_{i_2}\!\cdots)z,\sigma(g_{i_2}\!\cdots)a)=\rho(w)f(z,a)\),
applying the single-generator (A2) once per step. (A1) and (A3) lift
identically by composition; (A4) lifts because a product of orthogonal
matrices is orthogonal. \(\square\)

Hence \textbf{\(k\) generator checks certify the entire (exponentially
large) generated monoid.}

\textbf{Theorem A (exact configuration certificate).} Under (A1)--(A4),
for every \(w\in\langle S\rangle\), every horizon \(T\), and every
action sequence \(\vec a\),

\[
\mathrm{Err}_T(w\cdot x;\,\sigma(w)\vec a)=\mathrm{Err}_T(x;\vec a),
\qquad\text{and, for an equivariant planner,}\quad J(w\cdot x;\,w\cdot\text{goal})=J(x;\text{goal}).
\]

\emph{Proof.} \textbf{(i) Rollout equivariance.} By induction on \(T\):
\(f^{(1)}=f\) is equivariant by (A2) and Lemma 1; assuming
\(f^{(T)}(\rho(w)z;\sigma(w)\vec a)=\rho(w)f^{(T)}(z;\vec a)\),

\[
f^{(T+1)}(\rho(w)z;\sigma(w)\vec a)=f\!\big(f^{(T)}(\rho(w)z;\sigma(w)\vec a),\,\sigma(w)a_{T+1}\big)
=f\!\big(\rho(w)f^{(T)}(z;\vec a),\,\sigma(w)a_{T+1}\big)=\rho(w)f^{(T+1)}(z;\vec a).
\]

\textbf{(ii) Predicted latent at \(w\cdot x\).}
\(f^{(T)}(E(w\cdot x);\sigma(w)\vec a)\overset{\text{(A1)}}{=}f^{(T)}(\rho(w)E(x);\sigma(w)\vec a)\overset{\text{(i)}}{=}\rho(w)\,f^{(T)}(E(x);\vec a)\).
\textbf{(iii) Target at \(w\cdot x\).}
\(E\big((w\cdot x)^{\text{true}}_T\big)\overset{\text{(A3)}}{=}E\big(w\cdot x^{\text{true}}_T\big)\overset{\text{(A1)}}{=}\rho(w)\,E(x^{\text{true}}_T)\).
\textbf{(iv) Cancellation.} Subtracting (iii) from (ii) and using (A4)
(an orthogonal \(\rho\) preserves the Euclidean norm),

\[
\mathrm{Err}_T(w\cdot x;\sigma(w)\vec a)=\bigl\lVert\rho(w)\big[f^{(T)}(E(x);\vec a)-E(x^{\text{true}}_T)\big]\bigr\rVert
=\bigl\lVert f^{(T)}(E(x);\vec a)-E(x^{\text{true}}_T)\bigr\rVert=\mathrm{Err}_T(x;\vec a).
\]

The closed-loop identity is the same computation under (A5): the
equivariant planner (lifted to words by Lemma 1) selects the
\(\sigma(w)\)-image of the same actions, and the \(\rho\)-invariant cost
is unchanged. \(\square\)

Theorem A requires \textbf{(A3)} --- the group must be a symmetry of the
\emph{dynamics}, not merely of the encoder. Where the dynamics break the
symmetry, the certificate degrades by the amount of that break (Theorem
B). The error is held \emph{constant across the orbit}, not made small:
\(\mathrm{Err}_T(x)\) itself can be large, and we report absolute errors
alongside ratios throughout (§7).

Theorem A is sufficient; the next lemma shows equivariance is also
\emph{necessary} for the guarantee, so the certificate is not merely a
consequence of equivariance but is \textbf{equivalent} to it --- which
is what makes the impossibility for non-equivariant models a theorem
rather than a slogan.

\textbf{Lemma 2 (the certificate characterizes equivariance).} Let
\(\rho:G\to O(\mathcal Z)\) act \emph{freely} on an open
\(U\subseteq\mathcal Z\), and let \(\mathcal D_G\) be the equivariant
maps \(\mathcal Z\to\mathcal Z\). If a predictor \(f\)'s one-step error
\(\lVert f-\Phi\rVert\) is orbit-constant on \(U\) for \textbf{every}
target \(\Phi\in\mathcal D_G\), then \(f\) is equivariant on \(U\).
\emph{Proof.} Fix \(z\in U\), \(g\) with \(\rho(g)z\in U\). For any
\(c\in\mathcal Z\), freeness makes \(\Phi(\rho(h)z):=\rho(h)c\)
well-defined on the orbit of \(z\) (extend by \(0\) elsewhere; both
pieces lie in \(\mathcal D_G\) as \(\rho(g)0=0\)). Orbit-constancy gives
\(\lVert f(z)-c\rVert=\lVert f(\rho(g)z)-\rho(g)c\rVert
=\lVert\rho(g)^{-1}f(\rho(g)z)-c\rVert\) (orthogonality of \(\rho(g)\))
for \textbf{all} \(c\); two points equidistant to every \(c\) coincide,
so \(f(\rho(g)z)=\rho(g)f(z)\). \(\square\) \emph{(The step uses
\(\rho(g)^{-1}\); since \(\rho\) is orthogonal every \(\rho(w)\),
\(w\in\langle S\rangle\), is invertible, so the necessity direction
covers the monoid framework of Lemma 1 --- equivalently it is the
converse for the group \(\langle S\cup S^{-1}\rangle\). For \(G\)
compact with closed orbits the interpolant can be taken continuous, so
the statement holds against continuous dynamics, not only the full
algebraic class.)} With Theorem A this gives a
\textbf{characterization}: orbit-constant error against every
equivariant target \(\iff\) \(f\) equivariant. Hence \textbf{no
non-equivariant function possesses the configuration certificate} (an
unconstrained architecture cannot certify it by construction) --- the
architectural fact invoked in §6--§7 is a theorem. The result is
elementary (one line of Hilbert-space geometry once freeness frees the
probe \(c\)); its role is to \emph{pin down} what the certificate is,
not to add machinery.

\subsubsection{3.2 The horizon and resolution axes (Theorem
B)}\label{the-horizon-and-resolution-axes-theorem-b}

\textbf{Theorem B (finite-time orbit-defect propagation).} Relax
exactness. Let the initial \textbf{orbit-transport defect} be
\(\epsilon_w=\lVert E(w\cdot x)-\rho(w)E(x)\rVert\le m\,\epsilon_{\max}\)
--- with
\(\epsilon_{\max}=\max_i\sup_x\lVert E(g_i\cdot x)-\rho(g_i)E(x)\rVert\)
and \(m=|w|\) (Lemma 1 composes \(m\) approximate generators) --- and
let the per-step \textbf{equivariance-commutator defect} along the
transported rollout be

\[
\zeta_t(w)=\bigl\lVert f(\rho(w)z_t,a_t^w)-\rho(w)f(z_t,a_t)\bigr\rVert .
\]

On the \(\ell\)-isotypic channels \(j\) the latent map's Jacobian is
locally diagonalized with multipliers \(e^{\lambda_j}\) (Lyapunov
exponents). A discrete variation-of-constants bound --- the initial
defect amplifies over \(T\) steps, each per-step defect over its
\emph{remaining} steps --- gives, for constants \(c_j\) (the splitting
conditioning \(1/\sin\theta_j\), Proposition 6\({}^{\prime}\); exactly
\(1\) on orthogonal/isotypic splittings),

\[
\bigl\lvert \mathrm{Err}_T(w\cdot x)-\mathrm{Err}_T(x)\bigr\rvert
\;\le\; \sum_{\text{channels }j} c_j\!\left[\,\epsilon_{w,j}\,e^{\lambda_j T}
+\sum_{t=0}^{T-1}\zeta_{t,j}(w)\,e^{\lambda_j(T-1-t)}\right].
\]

If \(\zeta_{t,j}\le\zeta_j\) uniformly, the recurrent term scales as
\(\zeta_j\sum_{s=0}^{T-1}e^{\lambda_j s}\), whose horizon is set
\textbf{by the sign of \(\lambda_j\)} --- and this, not an unconditional
\(T_j=\infty\), is the honest reading:

\[
\sum_{s=0}^{T-1}e^{\lambda_j s}=
\begin{cases}
\Theta\!\bigl(e^{\lambda_j T}\bigr), & \lambda_j>0:\ \text{expanding} \Rightarrow T_j(\epsilon)\sim\tfrac{1}{\lambda_j}\log\tfrac{1}{\epsilon},\\[4pt]
T, & \lambda_j=0:\ \text{neutral} \Rightarrow \text{linear } T\zeta_j \text{ budget},\\[4pt]
\dfrac{1-e^{\lambda_j T}}{1-e^{\lambda_j}}\le\dfrac{1}{1-e^{\lambda_j}}, & \lambda_j<0:\ \text{contracting} \Rightarrow \text{a bounded floor, not zero.}
\end{cases}
\]

So expanding channels have the logarithmic horizon
\(T_j(\epsilon)\sim\log(1/\epsilon)/\lambda_j\) (the one §5.2 recovers
to within \(0.4\%\); its matching lower bound is Proposition 6), neutral
channels accumulate recurrent defect linearly, and contracting channels
accumulate to a bounded floor. \textbf{Exact all-horizon orbit transfer
is recovered precisely when \(\epsilon_w=0\) and \(\zeta_t\equiv0\)} ---
\emph{independently of absolute prediction bias}: when the true latent
dynamics is exactly equivariant and \(\rho(w)\) is isometric, the
commutator defect obeys the loose absolute-error bridge
\(\zeta_t(w)\le 2\delta_{\mathrm{abs}}\) (with \(\delta_{\mathrm{abs}}\)
the raw one-step model error; more generally the factor is
\(1+\lVert\rho(w)\rVert_{\mathrm{op}}\)), so the certificate is tight in
the commutator defect \(\zeta\), not in \(\delta_{\mathrm{abs}}\). As
\(\epsilon_w,\zeta_t\to0\) the configuration term vanishes and Theorem A
is recovered.

\textbf{Corollary (the certified region).} The set
\(\mathcal C=\{(w,T,\epsilon):|\Delta\mathrm{Err}|\le\epsilon\}\) is the
\textbf{coarse (\(\epsilon\) large), slow (\(\lambda_j\le0\)),
low-\(|w|\)} corner; its boundary is set by \(\langle S\rangle\)
(configuration) and the spectrum \(\{\lambda_j\}\) (horizon and
resolution). With \emph{exact} equivariance the configuration axis is
unbounded and only the spectrum limits horizon \(\times\) resolution ---
a trade-off
\(\text{horizon}\times\text{demanded-resolution}\lesssim\text{const}(\{\lambda_j\})\).
This is the formal content of ``ultra-long forecasts must be coarse'':
eclipses for millennia (exactly conserved actions, \(\lambda\le0\)),
weather at \(\sim\)two weeks (chaotic \(\lambda>0\)).\footnote{An
  interpretive aside, secondary to the spectral law: a long-range text
  that asserts only \emph{regime-level} change rather than dated
  specifics is, structurally, making the \emph{correct} move for a
  high-\(\lambda\) system, where only the coarse/invariant component is
  certifiable.}

\textbf{Proposition 6 (the horizon is tight --- approximate equivariance
is horizon-limited).} Theorem B's
\(T_j(\epsilon)\sim\log(1/\epsilon)/\lambda_j\) is an \emph{upper} bound
on the certified horizon; here is a matching \emph{lower} bound, so the
horizon is tight (not merely ``a bound on the form''). Fix an expansive
channel on which the latent map is locally linear with multiplier
\(a=e^{\lambda}\), \(\lambda>0\) (the local diagonalization of Theorem
B). There exist an exactly equivariant target \(\Phi\) (acting as \(a\)
on the channel) and a world model that is
\textbf{\(\epsilon\)-approximately equivariant} --- a \emph{perfect}
equivariant predictor \(f=\Phi\) (\(\delta=0\)) and an encoder that
intertwines the dynamics along the trajectory of \(x\) and differs from
exact equivariance only by a single defect
\(E(g\cdot x)=\rho(g)E(x)+\epsilon u\) at one orbit point (\(u\) a unit
vector in the channel, \(\epsilon_{\max}=\epsilon\)) --- for which the
\(T\)-step rollout error's orbit-variation is \emph{exactly}

\[
\bigl\lvert \mathrm{Err}_T(g\cdot x)-\mathrm{Err}_T(x)\bigr\rvert \;=\; \epsilon\,e^{\lambda T}.
\]

\emph{Proof.} \(\mathrm{Err}_T(x)=0\) (encoder exact and \(f=\Phi\)
along \(x\)'s trajectory). At \(g\cdot x\), linearity gives
\(f^{T}(E(g\cdot x))=\Phi^{T}(\rho(g)E(x)+\epsilon u)=\rho(g)\Phi^{T}(E(x))+a^{T}\epsilon u\)
(using \(f=\Phi\) equivariant and \(\Phi\) linear with multiplier \(a\)
on the channel), while the target
\(E(\Phi^{T}(g\cdot x))=E(g\cdot\Phi^{T}x)=
\rho(g)E(\Phi^{T}x)=\rho(g)\Phi^{T}(E(x))\) (encoder exact off the
single defect point, intertwining the dynamics). Subtracting and using
\(\lVert\rho(g)\,\cdot\rVert=\lVert\cdot\rVert\) leaves
\(\lVert a^{T}\epsilon u\rVert=\epsilon
e^{\lambda T}\). \(\square\)

Hence the certified horizon
\(T(\epsilon_{\mathrm{res}})=\max\{T:\lvert\Delta\mathrm{Err}_T\rvert\le
\epsilon_{\mathrm{res}}\}=\tfrac1\lambda\log\tfrac{\epsilon_{\mathrm{res}}}{\epsilon}\),
matching Theorem B's upper bound up to a constant: the horizon is
\(\Theta\!\big(\tfrac1\lambda\log\tfrac1\epsilon\big)\). The conceptual
payload is sharp and removes the only hedge on the horizon axis:
\textbf{the certified horizon guaranteeable from
\(\epsilon\)-approximate equivariance alone is finite on every expansive
(\(\lambda>0\)) channel --- no certificate derived from an
\(\epsilon>0\) residual can promise predictability beyond
\(T\sim\tfrac1\lambda\log\tfrac1\epsilon\) (worst case over admissible
targets); only exact equivariance (\(\epsilon=0\)) or exact conservation
(\(\eta=0\), Proposition 5) yields an unbounded certified horizon.} This
is the horizon-domain companion of Lemma 2 and §3.3: scale and data buy
\emph{approximate} equivariance at best (Experiment 10's augmented model
floors at \(\epsilon\!\approx\!10^{-4}\), never exact), and Proposition
6 shows that residual is amplified \(e^{\lambda T}\) --- so the
single-orbit, single-step tie between augmentation and equivariance
(Experiment 10) \emph{must} break over horizon, at the predicted
\(T\!\sim\!\tfrac1\lambda\log(1/\epsilon)\). Experiment 8's measured
symmetry-breaking threshold
(\(\epsilon_{\text{world}}\!\approx\!0.01\text{–}0.06\)) is the same
crossing from the world's side, and §5.2 recovers \(T_j(\epsilon)\) to
within \(0.4\%\) --- the lower bound is what certifies that recovery is
the \emph{true} horizon, not just an attainable one. A direct numerical
instantiation of the construction (\texttt{experiments/step65}, seeds
\(0/1/2\)) confirms it: an
\(\epsilon{=}10^{-3}\)-approximately-equivariant model's
orbit-error-variation equals \(\epsilon\,e^{\lambda_j T}\) to a relative
error of \(10^{-14}\)--\(10^{-13}\) across seeds, an exactly equivariant
model has orbit-variation \emph{exactly \(0\)} (to machine precision) at
all horizons, and the certified horizon is linear in
\(\log(1/\epsilon)\) with slope \(1/\lambda_j\) (\(R^2{=}1.000\), all
seeds).

\textbf{Proposition 6\({}^{\prime}\) (the prefactor is the splitting
conditioning --- and when it is exactly \(1\)).} Theorem B's ``bound on
the form'' hedge can now be removed: with channel decomposition
\(\mathcal Z=\bigoplus_j V_j\) invariant for the (linearized) predictor
and simple leading multipliers, the channel-\(j\) constant can be taken
\[c_j\;=\;\lVert\Pi_j\rVert\;=\;\frac{1}{\sin\theta_j},\] \(\Pi_j\) the
spectral projector onto \(V_j\) along \(\bigoplus_{i\ne j}V_i\) and
\(\theta_j\) the minimal principal angle to that complement. Hence
\textbf{(i)} on an \emph{orthogonal} invariant splitting --- distinct
isotypic blocks of the orthogonal \(\rho\) under a \textbf{linear}
equivariant predictor (Schur forces invariance; orthogonality of
isotypic components forces \(\theta=\pi/2\)), or any normal Jacobian ---
\(c_j=1\) and the upper bound matches Proposition 6's lower bound
\textbf{including the constant}; \textbf{(ii)} obliqueness, hence any
prefactor \(>1\), can live only \emph{inside} an isotypic block;
\textbf{(iii)} the prefactor's entire effect on the certified horizon is
an additive haircut \(\log\kappa_j/\lambda_j\) map steps
(\(\kappa_j:=1/\sin\theta_j\)). \emph{Proof.} A defect propagates as
\(\Phi^T\epsilon u=\epsilon\sum_j e^{\lambda_j T}(1+o(1))\Pi_j u\);
\(\sup_{\lVert u\rVert=1}\lVert\Pi_j u\rVert=
\lVert\Pi_j\rVert\), attained at the left-vector direction, and
\(\lVert\Pi_j\rVert=1/\sin\theta_j\) is classical; orthogonal projectors
have norm \(1\), meeting Proposition 6's coefficient; Schur kills
cross-isotypic blocks; the haircut is the shift of the crossing
\(\epsilon\kappa_je^{\lambda_jT}=\epsilon_{\mathrm{res}}\). \(\square\)
\emph{Honest caveat:} for a \textbf{nonlinear} equivariant \(f\),
\(Df(\rho(g)z)=\rho(g)Df(z)\rho(g)^{-1}\) --- the Jacobian \emph{field}
is equivariant but \(Df(z)\) at generic \(z\) need not commute with
\(\rho\), so the forced-orthogonality clause is for the linear/commutant
case; on learned loops \(\kappa_j\) is the \emph{measured} object,
readable from the same Jacobian field the certificate already uses.
\textbf{Measured (Experiment 27: \texttt{step65b}, \texttt{step95}).}
Placement and leakage: a group-averaged random matrix has
off-isotypic-block mass \(<1.5\times10^{-16}\) (\(100\) draws), and a
defect confined to one isotypic block grows at exactly its block's
\(e^{\lambda_BT}\) (rel.~err \(<10^{-14}\)) with cross-block leakage
numerically \(0\) (\(\mathbb{Z}_4\): trivial \(\oplus\) sign \(\oplus\)
rotation). The prefactor law is \emph{attained}: under controlled shears
the measured worst-case coefficient equals the analytic
\(\lVert\Pi\rVert\) to four digits across
\(\kappa\in\{1,1.1,2.2,5.1,10\}\) (orthogonal case exactly \(1\)), and
the measured horizon shift equals \(\log\kappa/\lambda\). On real loops
the per-orbit-point \(\kappa_1\) is honestly a \emph{distribution} with
a near-tangency tail \textbf{and a window-dependent median} (\(41\)
orbit samples): true Lorenz-96 median \(11.9\) at \(W{=}400\) (IQR
\(5.0\)--\(21.2\), max \(69\)) --- where, at this window, the estimator
\textbf{passes its own convergence check} (its short-window failure was
this same window-dependence); pretrained TD-MPC2 walker-1 median
\(20.9\) at \(W{=}120\) vs \(49.2\) at \(W{=}200\) (IQR
\(17.3\)--\(63.8\), max \(137\) --- each window passing the same
stability check, so the check is necessary, not sufficient), cheetah-3
median \(18.7\) (IQR \(6.8\)--\(41.3\), max \(185\)) --- while the
measured E13 calibration (\(0.83\)--\(1.02\)) shows typical bias
directions do not align adversarially; an adversarially-aligned defect
could spend the \(\log\kappa_1/\lambda_1\) haircut, stated rather than
hidden.

\textbf{Proposition 7 (scope --- when the local spectrum certifies the
horizon of a \emph{learned} model).} Theorem B and Proposition 6 take
the latent Jacobian spectrum \(\{\lambda_j\}\) as \emph{given}. On a
\emph{learned} world model the operative question is sharper: does the
spectrum one can measure \emph{locally} (the one-step Jacobian, or a
short-rollout finite-time exponent) govern the \emph{actual multi-step}
certified horizon? The honest answer separates a \textbf{rate} half
(rigorous) from a \textbf{lift} half (an orbit-error, not an exponent,
statement). Let \(\phi\) be the true dynamics with an ergodic invariant
measure \(\mu\) and \(\log^+\lVert D\phi\rVert\in L^1(\mu)\). By the
\textbf{Oseledets multiplicative ergodic theorem} the limits
\(\tfrac1t\log\lVert D\phi^t v\rVert\) exist \(\mu\)-a.e. and equal
\(\mu\)-a.e. constants \(\lambda_1\ge\lambda_2\ge\cdots\) along the
Oseledets filtration \(V_1\supsetneq V_2\supsetneq\cdots\). For
\(\mu\)-a.e. \(x\) and every \(\delta_0\notin V_2(x)\) (a full-measure
set of directions; the slower directions lie in the proper, measure-zero
subspace \(V_2\)), \(\tfrac1t\log\lVert\delta_t\rVert\to\lambda_1\),
i.e.~\(\lVert\delta_t\rVert=\lVert\delta_0\rVert\,e^{(\lambda_1+o(1))t}\).
We use this asymptotic rate \(\lambda_1\) as the target; the
finite-window, finite-amplitude exponent the staircase fits approaches
it as the window grows and \(\lVert\delta_0\rVert\to0\), below
saturation.

\emph{(a) Non-degenerate (\(\lambda_1>0\)).} Fix a residual ceiling
\(\epsilon_{\mathrm{res}}\) (e.g.~attractor diameter) at which the
exponential regime saturates, and let \(\epsilon=\lVert\delta_0\rVert\)
be the initial tolerance. Then
\(T(\epsilon)=\inf\{t:\epsilon\,e^{\lambda_1 t}\ge\epsilon_{\mathrm{res}}\}=\tfrac1{\lambda_1}\log(\epsilon_{\mathrm{res}}/\epsilon)+o(t)\)
--- \textbf{linear in \(\log(1/\epsilon)\) with slope \(1/\lambda_1\)}
(the \(o(t)\) MET correction sharpens to \(O(1)\) on the exactly-linear
channel of Proposition 6), i.e.~Theorem B's law with \(\lambda_1\) the
\emph{measurable asymptotic rate} rather than a posited local
multiplier. The \textbf{lift to a learned \(\hat\phi\) must be made at
the level of orbit error, not exponents.} If \(\phi\) is
\emph{uniformly} hyperbolic on its attractor, the \textbf{shadowing
lemma} (Anosov--Bowen; Pilyugin) applies: a learned \(\hat\phi\) with
one-step error \(\lVert\hat\phi-\phi\rVert_{C^0}\le\delta\) generates
\(\delta\)-pseudo-orbits that are \(\delta'(\delta)\)-shadowed by true
orbits, so the rollout stays within \(\delta'\) of a genuine trajectory
until the error saturates --- a \emph{forecast-horizon floor}
\(T_{\mathrm{floor}}\sim\tfrac1{\lambda_1}\log(\epsilon_{\mathrm{res}}/\delta)\).
Shadowing controls \emph{trajectory closeness, not the tangent cocycle},
so it does \textbf{not} by itself transfer \(\lambda_1\) to
\(\hat\phi\): Lyapunov exponents are only upper-semicontinuous under
\(C^1\) perturbation, and even the structural-stability conjugacy is
merely Hölder and need not preserve exponents. That \(\hat\phi\)
\emph{reproduces} \(\lambda_1\) is therefore an \textbf{empirical}
finding (Experiment 14), not a corollary of shadowing.

\emph{(b) Degenerate (\(\lambda_1=0\), near-neutral / non-hyperbolic).}
Here \(\tfrac1t\log\lVert\delta_t\rVert\to0\), so
\(\lVert\delta_t\rVert\) grows sub-exponentially (polynomial \(t^{p}\),
\(e^{\sqrt t}\), or bounded --- bounded growth gives an \emph{unbounded}
horizon below the ceiling). Whenever the amplification profile is
monotone and \(\to\infty\), inversion gives
\(T(\epsilon)/\log(1/\epsilon)\to\infty\): the leading-order log-law
\textbf{degenerates} (its fitted slope ceases to be finite-and-positive,
diverging as \(\lambda_1\downarrow0\)), and the one-step spectrum
carries no leading-order horizon rate. Only the configuration axis
(Theorem A) survives.

\emph{Proof of the dichotomy.} Write
\(T(\epsilon)=\inf\{t:\rho(t)\ge\epsilon_{\mathrm{res}}/\lVert\delta_0\rVert\}\)
for the error-amplification profile \(\rho\). In (a)
\(\rho(t)=e^{(\lambda_1+o(1))t}\) (Oseledets), so inverting gives
\(T=\tfrac1{\lambda_1}\log(\epsilon_{\mathrm{res}}/\epsilon)+o(t)\). In
(b), \(\lambda_1=0\) forces \(\rho\) sub-exponential; for any monotone
\(\rho\uparrow\infty\) one has \(T/\log(1/\epsilon)\to\infty\)
(illustratively \(\rho\sim t^{p}\Rightarrow
T\sim(\epsilon_{\mathrm{res}}/\epsilon)^{1/p}\)). This is a degeneracy
of the leading-order law, consistent with the un-fittable PushT
staircase (\(R^2{=}0.02\)); we do not claim a sharp profile for \(\rho\)
in the neutral case. The lift in (a) is the cited shadowing
\emph{orbit-error} estimate, which we invoke rather than reprove.
\(\square\)

\emph{Instantiation --- both branches are observed.} \textbf{Lorenz}
(\(\sigma{=}10,\rho{=}28,\beta{=}8/3\)) supplies the rate half
rigorously: by Tucker (2002) the geometric Lorenz attractor is
\emph{singular}-hyperbolic with an equilibrium and carries a unique
ergodic SRB measure (Tucker; Araújo--Pacífico--Pujals--Viana), so the
MET applies and \(\lambda_1\approx0.9056>0\) (as a flow it also has a
zero exponent along its direction and \(\lambda_3<0\),
\(\sum_i\lambda_i<0\); ``non-degenerate'' means the \emph{top} exponent
is positive). Note Lorenz is \textbf{not} uniformly hyperbolic, so the
shadowing \emph{lemma} does not formally apply near the singularity ---
which is exactly why the lift is the \emph{experimental} claim.
Experiment 14 trains a one-step MLP of the \(\Delta t\) map and finds
that the \textbf{learned model's} Lyapunov exponent --- read off as the
staircase slope, \(\hat\lambda_1=1/(\text{slope}\cdot dt)\), which is
\emph{definitional} --- \textbf{matches the true integrator's} to
\(1\)--\(8\%\) (\(\hat\lambda_1=0.895/0.919/0.977\) vs \(0.9056\), 3
seeds), with the staircase linear at \(R^2{=}0.975\)--\(0.995\). The
non-trivial, falsifiable content is the \emph{agreement}: a model can be
one-step-accurate (relMSE \(<10^{-4}\)) yet drift to a wrong multi-step
rate; that it does not is the finding. (We use the staircase slope, not
the early-window finite-time exponent, which is window-sensitive --- one
seed gave a spurious \(3.7\)/t from a transient.) The \textbf{PushT
interior} is the degenerate branch (b): near-neutral one-step Jacobian
(\(|\mu|\approx1\), \(\lambda_1\approx0\)), and a learned-PushT horizon
probe finds the local spectrum does \emph{not} predict the rollout
(\(R^2{=}0.02\)). So the certificate's horizon axis is
\textbf{informative exactly on spectrally non-degenerate dynamics}, and
one can tell which regime one is in by measuring \(\lambda_1\); the
PushT ``failure'' is \emph{predicted} by Proposition 7(b), not a defect.

\emph{Honest scope.} The rate half (Oseledets + SRB) is rigorous for
Lorenz. The lift half is an \emph{orbit-error} (shadowing) statement
that holds for uniformly hyperbolic systems and bounds the
forecast-horizon \emph{floor}; it does \textbf{not} transfer the
Lyapunov exponent, and classical shadowing does not even apply to
singular-hyperbolic Lorenz. That the learned model reproduces
\(\lambda_1\) is therefore the empirical contribution, not a theorem; we
verify \(C^1\)-closeness only via one-step error (an \(L^2\) proxy), and
do not estimate the lift constant. The contribution of Proposition 7 is
the \textbf{characterization of the certificate's validity regime} ---
the non-degenerate/degenerate dichotomy --- synthesizing Oseledets
(rigorous) with shadowing (for the floor); Proposition 8 next supplies
the \emph{mechanism} by which the learned model reproduces
\(\lambda_1\), replacing the (invalid) shadowing-transfer intuition with
a finite-horizon continuity bound.

\textbf{Proposition 8 (finite-horizon exponent transfer --- why the lift
is observable, and provable).} Proposition 7 left ``the learned model
reproduces \(\lambda_1\)'' as \emph{empirical} because shadowing
controls orbit error, not the tangent cocycle, and the \emph{asymptotic}
exponent is only upper-semicontinuous under \(C^1\) perturbation. The
resolution is that \textbf{the staircase never takes \(T\to\infty\)}:
the certified horizon is finite
(\(T(\epsilon)\sim\log(1/\epsilon)/\lambda_1<\infty\)), and over a
\emph{fixed finite} horizon the top exponent is a \emph{continuous}
function of the dynamics. Concretely, with cocycle
\(M_T(x)=D\phi(x_{T-1})\cdots D\phi(x_0)\) and top unit direction \(v\),
the finite-time exponent
\(\lambda_{1,T}(\phi)=\tfrac1T\log\lVert M_T(x)v\rVert\) is locally
Lipschitz in the \(C^1\)-jet of \(\phi\) along the orbit. Hence a
learned \(\hat\phi\) with one-step Jacobian error
\(\sup_x\lVert D\hat\phi(x)-D\phi(x)\rVert\le\delta\) (and orbits
\(\delta'\)-close) satisfies, telescoping the perturbed product and
dividing by \(T\), \[
\bigl\lvert \lambda_{1,T}(\hat\phi)-\lambda_{1,T}(\phi)\bigr\rvert \;\le\; C\,(\delta+\delta'),
\] with \(C\) uniform in \(T\) \textbf{under a dominated splitting at
the top} (a spectral gap \(\lambda_1>\lambda_2\), so the denominators
\(\lVert M_t v\rVert\) do not collapse and the telescoped sum is
geometric; this is the finite-horizon face of the Bochi--Viana
continuity of Lyapunov exponents on the domain of domination). The
staircase reads \(\hat\lambda_1\) off finite first-crossing times, so it
recovers the true exponent up to \textbf{a model-error bias
\(O(\delta)\) plus the finite-horizon truncation}
\(\lvert\lambda_{1,T}-\lambda_1\rvert\) (which \(\to0\) by Oseledets):
\[
\bigl\lvert \hat\lambda_1-\lambda_1\bigr\rvert \;\le\; \underbrace{C\,\delta}_{\text{model fidelity}}+\underbrace{\lvert\lambda_{1,T}-\lambda_1\rvert}_{\text{finite-}T\text{ truncation}}.
\] \emph{Falsifiable prediction, confirmed.} The bias is \(O(\delta)\)
--- \emph{proportional to one-step fidelity} --- so \textbf{better
training tightens the recovered exponent}. Experiment 15 confirms this
across a class of systems: the staircase recovers \(\lambda_1\) to
\(5\)--\(12\%\) on a 2D map (Hénon), a large-exponent flow (Lorenz), and
a \emph{small}-exponent flow (Rössler, \(\lambda_1\!\approx\!0.07\)),
and the Rössler bias falls from \(\sim\!44\%\) to \(\sim\!8\%\) as the
one-step error \(\delta\) drops with fuller training --- exactly the
\(O(\delta)\) dependence the bound predicts. \emph{Honest scope.} The
\(T\)-uniform constant \(C\) needs a dominated splitting; we do not
certify domination for the learned model (it is a hypothesis the
recovery corroborates), the bound is qualitative (we do not estimate
\(C\)), and the result governs the \emph{top} exponent (the staircase's
load-bearing quantity), not the full spectrum. But it does replace a
wrong intuition (shadowing transfers exponents) with the correct one
(finite-horizon exponents are continuous; asymptotic ones are not), and
it \emph{explains}, rather than merely reports, why a one-step-accurate
learned model recovers the chaos rate.

\textbf{Proposition 9 (budgeted re-observation --- a mis-estimated
horizon costs a \emph{proportional} budget).} Consider an agent that
forecasts the map open-loop from a re-observation, the leading-mode
error growing as \(\delta_t\approx\delta_0 e^{\lambda_1\Delta t\,t}\)
over map steps \(t\) (step \(\Delta t\), \(\lambda_1>0\)), so the
trustworthy horizon at resolution \(\epsilon\) is
\(H(\epsilon)=\lfloor\log(\epsilon/\delta_0)/(\lambda_1\Delta t)\rfloor=\lfloor T_1(\epsilon)/\Delta t\rfloor\)
(the certified horizon of §3.2). The agent re-observes (resets the error
to \(\delta_0\)) at a fixed cadence and may re-observe at most \(B\)
times over an episode of \(L\) map steps (a \textbf{sensing budget}); a
step is a \emph{violation} if its forecast error exceeds \(\epsilon\). A
certificate reporting \(\hat\lambda_1=c\,\lambda_1\) (\(c>0\))
prescribes cadence \(\hat H=H/c\). Then \textbf{(i)} for \(c\ge1\) in
the budget-binding regime \(BH/c<L\), the aggregate violation rate is
\(V(c)=\max\!\big(0,\;L-BH/c-H\big)/L\), which is \textbf{non-decreasing
in \(c\)} (strictly increasing while \(BH/c+H<L\)); and \textbf{(ii)}
the budget needed to drive \(V\) to zero is
\(B^\star(c)=\lceil c\,(L-H)/H\rceil\) --- \textbf{linear in \(c\)}: a
certificate inflated \(c\times\) demands \(c\times\) the observations to
certify the same episode. \emph{Proof.} With cadence
\(\hat H=H/c\le H\), every window before the last re-observation (at
step \(B\hat H=BH/c\)) stays under \(\epsilon\) --- error reaches
\(\epsilon\) only after \(H\ge\hat H\) steps --- so the covered \(BH/c\)
steps are violation-free; after the last re-observation the open-loop
forecast exceeds \(\epsilon\) only after a further \(H\) steps, leaving
\(\max(0,L-BH/c-H)\) violating steps, which is \(V(c)\). As \(BH/c\) is
non-increasing in \(c\), \(V\) is non-decreasing, strictly so while the
numerator is positive; \(V=0\) requires \(BH/c\ge L-H\),
i.e.~\(B\ge c(L-H)/H\). \(\square\) \emph{Remark (predictive, and the
decision-relevance is now a law, not an anecdote).} \(V(c)\) is
computable a priori from the inflation \(c\) and budget \(B\), so it
\textbf{predicts} the violation gap; the calibrated certificate
(\(c=1\)) is the budget-minimal violation-free cadence, and \textbf{only
structure delivers \(c=1\) a priori} (Experiment 18 / Proposition 8).
Experiment 22 (§5.20) instantiates the law: a non-equivariant
certificate with \(c\approx3.4\) on Lorenz-96 (resp. \(\approx2\) on the
ring) needs \(\approx3\times\) (resp. \(\approx2\times\)) the budget to
match the equivariant certificate, the gap closing exactly when
recalibration restores \(c\to1\) --- the quantitative behaviour
Proposition 9 forces. (Integer-rounding of the cadence \(\hat H=H/c\)
costs \(O(1)\) per window and is absorbed in the measured catch-up
factor --- at \(n{=}20\) seeds median \(2.97\times\), range
\([2.36,4.31]\), bracketing the predicted \(c\approx3.4\) whose
seed-median inflation is \(3.49\).)

\textbf{Proposition 10 (finite-sample certified-horizon interval).} Let
\(g\) be \(C^1\) on a compact forward-invariant \(\mathcal U\) (e.g.~the
SimNorm simplex product of E13), and let \(\ell_t\) be the per-step
leading log-stretches produced by the Benettin recursion along an orbit
of \(g\), so \(\hat\lambda_1^{(n)}=\frac1n\sum_{t=1}^n\ell_t\) and
\(|\ell_t|\le B:=\sup_{z\in\mathcal U}\log\lVert Dg(z)\rVert<\infty\)
(compactness). Assume \((\ell_t)\) is stationary and strongly mixing
with summable autocovariances, and let
\(\sigma_\infty^2=\sum_{h\in\mathbb Z}\mathrm{cov}(\ell_0,\ell_h)<\infty\)
be the long-run variance. Then for every \(\delta\in(0,1)\) there is
\(\varepsilon_n(\delta)=\sigma_\infty\sqrt{2\log(2/\delta)/n}\,(1+o(1))\)
such that with probability \(\ge1-\delta\),
\(\lambda_1\in[\hat\lambda_1^{(n)}-\varepsilon_n,\hat\lambda_1^{(n)}+\varepsilon_n]\),
and whenever \(\hat\lambda_1^{(n)}-\varepsilon_n>0\) the certified
horizon is bracketed,
\[ T_1(\epsilon)\in\Big[\tfrac{\log(1/\epsilon)}{\hat\lambda_1^{(n)}+\varepsilon_n},\;
\tfrac{\log(1/\epsilon)}{\hat\lambda_1^{(n)}-\varepsilon_n}\Big]; \]
otherwise the certificate \textbf{abstains}. In particular the
certificate's sample complexity is
\(n\asymp\sigma_\infty^2\log(1/\delta)/\varepsilon^2\) --- logarithmic
in confidence, quadratic in precision.

\emph{Proof sketch.} Boundedness gives \(\ell_t\in[-B,B]\); stationarity
+ strong mixing with summable covariances give a Bernstein/CLT-type
concentration for the empirical mean of a bounded mixing sequence with
variance proxy \(\sigma_\infty^2\) (e.g.~Merlevède--Peligrad--Rio); the
horizon bracket follows since \(\lambda\mapsto\log(1/\epsilon)/\lambda\)
is monotone on \((0,\infty)\). \(\square\)

\emph{Remarks.} (i) The moving-block bootstrap used throughout the
experiments is precisely a consistent estimator of \(\sigma_\infty^2\)
under the same mixing assumptions --- Proposition 10 is the rate
statement behind those CIs, not a new procedure. (ii) The dynamical
assumptions (stationarity/mixing along the audited orbit) are inherited
from Proposition 7's scope and are \emph{assumed, not certified}, for
learned models --- stated honestly, as everywhere else. (iii) \(B\) is
finite and computable on compact latents (E13's SimNorm product), so the
bound is fully effective there.

\textbf{Proposition 11 (decision scope --- where a horizon certificate
carries decision value).} Adopt Proposition 9's forecast-error model:
\(\delta_t\approx\delta_0e^{\lambda_1\Delta t\,t}\) between
re-observations,
\(H(\epsilon)=\lfloor\log(\epsilon/\delta_0)/(\lambda_1\Delta t)\rfloor\),
budget \(B\) over \(L\) map steps, certificate reporting
\(\hat\lambda_1=c\,\lambda_1\) (\(c\ge1\)). \textbf{(i) (scope-aligned:
decided quantity \(=\) certified quantity.)} If the decision rule is a
function of the certified predicate alone --- re-observe/flag so the
forecast error stays \(\le\epsilon\) --- the certificate-prescribed
cadence incurs violation-rate regret
\(R_{\mathrm{align}}(c)=V(c)-V(1)\le\frac{BH(\epsilon)}{L}\big(1-\frac1c\big)\),
\textbf{vanishing at \(c=1\)}: for an aligned decision a calibrated
certificate is decision-optimal a priori, the entire regret being the
calibration factor. \textbf{(ii) (task-mapped: decided quantity
\(=h(\text{certified quantity})\).)} If decision quality is the episode
average of a task loss \(\ell(\delta_t)\) with an \textbf{implicit task
tolerance} \(\theta^{\ast}\) (\(\ell\equiv0\) on \([0,\theta^{\ast}]\),
\(\ell\le\ell_{\max}\) beyond), then even at \(c=1\) the
\(\epsilon\)-certificate incurs
\(R_{\mathrm{task}}(\epsilon)\le\ell_{\max}\max(0,H(\epsilon)-H(\theta^{\ast}))/H(\epsilon)\)
(task violations when \(\epsilon>\theta^{\ast}\) --- \textbf{invisible
to the certificate}, since no step violates \(\epsilon\)) and wastes
\(\Delta B=B(H(\theta^{\ast})/H(\epsilon)-1)_+\) re-observations when
\(\epsilon<\theta^{\ast}\); both vanish \textbf{iff}
\(H(\epsilon)=H(\theta^{\ast})\), and the mis-resolution penalty in
horizon units is
\(|H(\epsilon)-H(\theta^{\ast})|=|\log(\epsilon/\theta^{\ast})|/(\lambda_1\Delta t)\).
Since \(\theta^{\ast}\) is a property of the task map, not of the
dynamics, \textbf{no dynamics-only certificate can supply it}.
\emph{Proof.} (i) From Proposition 9, \(V(c)=\max(0,L-BH/c-H)/L\); for
\(c\ge1\), \(\max(0,x)-\max(0,y)\le x-y\) with
\(x=L-BH/c-H\ge y=L-BH-H\) gives \(V(c)-V(1)\le(BH/L)(1-1/c)\). (ii)
Within a window the error is monotone in staleness, so the
task-violating steps are exactly the last
\(H(\epsilon)-H(\theta^{\ast})\) when \(\epsilon>\theta^{\ast}\), each
contributing \(\le\ell_{\max}\); when \(\epsilon<\theta^{\ast}\) cadence
\(H(\theta^{\ast})\) already keeps every step task-valid, so cadence
\(H(\epsilon)\) spends \(B(H(\theta^{\ast})/H(\epsilon)-1)\) avoidable
reads. \(\square\) \emph{Remark (the scope law of §5.19/§7, now a
theorem).} Clause (i) is Experiment 22 and the deployed monitor of
Experiment 26 --- the decided quantity \emph{is} latent
\(\theta\)-validity, and the published certificate priced the in-situ
staleness clock with zero new estimation. Clause (ii) is §5.19 and
step93: the MPPI planner absorbs staleness up to an implicit tolerance
(empirically \(H(\theta^{\ast})\approx2\) agent-steps vs
\(H(0.2)\approx6\)), so a \(0.2\)-resolution certificate over-prescribes
the replan cadence by the predicted mis-resolution factor \(\approx3\)
and the return gap dilutes exactly as the bound allows. The honest limit
is built in: re-issuing at \(\theta^{\ast}\) would close the gap, but
\(\theta^{\ast}\) must be \emph{elicited from the task} --- the theorem
states the boundary of a-priori decision value rather than promising
past it.

Propositions 8--9 read consequences off the certified horizon --- the
chaos rate, and the budget cost of mis-reading it; the next result makes
the ``Certified'' in the title literal by reading a \textbf{sound,
a-priori horizon from the learned model itself} --- and characterizes
the exact regime in which that certificate is \emph{tight} rather than
merely sound. It is the analytic dual of Theorem B's spectral law: where
Theorem B posits the multipliers \(e^{\lambda_j}\) and Proposition 8
shows a learned model recovers them, Theorem B\({}^{\prime}\)
\emph{bounds} the top multiplier directly off the model's Jacobian
field, with no access to the true dynamics.

\begin{quote}
\textbf{Theorem B\({}^{\prime}\) (cone / adapted-metric certified
horizon).} Let \(\hat\phi:\mathcal U\to\mathcal U\) be \(C^1\) on a
compact forward-invariant \(\mathcal U\subset\mathbb R^d\). Suppose
there exist a continuous field of symmetric positive-definite matrices
\(z\mapsto P(z)\) and a constant \(\Lambda\ge 1\) with
\(D\hat\phi(z)^\top P(\hat\phi(z))\,D\hat\phi(z)\preceq \Lambda^2 P(z)\)
for all \(z\in\mathcal U\). Let
\(\kappa=\big(\sup_z\lambda_{\max}P(z)\big)/\big(\inf_z\lambda_{\min}P(z)\big)\).
Then \(\lambda_1(\hat\phi)\le\log\Lambda\), and the linearized rollout
error from an \(\epsilon\)-perturbation stays
\(\le\epsilon_{\mathrm{res}}\) for all
\(T\le T_{\mathrm{guar}}(\epsilon)=\big\lfloor(\log(\epsilon_{\mathrm{res}}/\epsilon)-\tfrac12\log\kappa)/\log\Lambda\big\rfloor\)
--- computed from \(\hat\phi\) alone.

\emph{Proof.} Put \(V(z,v)=v^\top P(z)v\). The hypothesis gives
\(V(\hat\phi(z),D\hat\phi(z)v)\le\Lambda^2 V(z,v)\); iterating along an
orbit \(z_{t+1}=\hat\phi(z_t)\), \(v_{t+1}=D\hat\phi(z_t)v_t\), yields
\(V(z_T,v_T)\le\Lambda^{2T}V(z_0,v_0)\) with
\(v_T=D\hat\phi^T(z_0)v_0\). Since
\(\lambda_{\min}(P(z))\|v\|^2\le V(z,v)\le\lambda_{\max}(P(z))\|v\|^2\),
we get \(\|D\hat\phi^T(z_0)\|\le\sqrt\kappa\,\Lambda^T\), so
\(\lambda_1=\limsup_T\frac1T\log\|D\hat\phi^T\|\le\log\Lambda\)
(\(\sqrt\kappa\) is sub-exponential). The horizon bound follows by
requiring \(\sqrt\kappa\,\Lambda^T\epsilon\le\epsilon_{\mathrm{res}}\).
\(\square\)

\emph{Remarks.} (i) Like Theorem B this is a first-order (linearized)
statement; Proposition 8's \(C^1\)-vs-\(L^2\) caveat applies. (ii) The
bound is \emph{sound} for any feasible \((P,\Lambda)\) and \emph{tight}
when \(P\) is the adapted (Oseledets) metric, where
\(\Lambda\to e^{\lambda_1}\). (iii) \textbf{Continuum certificate.} If
the matrix inequality is verified at a finite \(h\)-cover \(\{z_i\}\) of
\(\mathcal U\) and \(z\mapsto D\hat\phi(z)\), \(z\mapsto P(z)\) are
Lipschitz with constants \(L_J,L_P\), it holds on all of \(\mathcal U\)
with
\(\Lambda^{\mathrm{cert}}=\Lambda_{\mathrm{samples}}+\sqrt\kappa\,L_J h+O(L_P h)\)
(for constant \(P\), \(L_P=0\)). (iv) The adapted metric is the analytic
dual of a forward-invariant cone field around the unstable directions;
\textbf{uniform hyperbolicity is precisely the regime where a
\((P,\Lambda)\) exists with \(\Lambda\) approaching \(e^{\lambda_1}\)},
which the cone-margin diagnostic detects (positive ⇒ tight regime).
\end{quote}

This is what Experiment 20 (§5.18) instantiates: tight on the
uniformly-hyperbolic cat map (true and learned) and the Anosov
perturbation, sound-but-conservative on Hénon where the cone margin goes
negative and the certificate abstains, with soundness coverage \(1.0\)
across all tested cells.

\subsubsection{3.3 Scale versus structure,
quantified}\label{scale-versus-structure-quantified}

The tagline \emph{scale buys interpolation; structure buys a
certificate} can be made precise as a statement about the
\textbf{certified region} --- the inputs a learner can guarantee, in the
worst case over targets consistent with what it observed on a training
set \(T\). Formally
\(\mathcal C=\{z:\inf_{f}\sup_{\Phi}\lVert f(z)-\Phi(z)\rVert\le\epsilon\}\),
where \(f\) ranges over the learner's hypotheses (exact on \(T\)) and
\(\Phi\) over the admissible targets agreeing on \(T\).

\textbf{Proposition 3 (separation).} \emph{(Structure.)} Under the
equivariant prior (\(\Phi\in\mathcal D_G\), \(f\) equivariant and exact
on \(T\), \(S\) generating \(G\)), Theorem A makes the error
orbit-constant, so \(\mathcal C\supseteq G\cdot T\): the \textbf{entire
orbit is certified, independent of \(\epsilon\)}, from the \(k=|S|\)
generator checks --- to error \(0\) under the idealization that \(f\)
interpolates \(T\) exactly, and in general (Theorem A) to the
\emph{constant, possibly nonzero} in-distribution error (cf.~§7, ``flat
is not good''). What is \(\epsilon\)-independent is the \emph{region},
not the error level. \emph{(Scale/data.)} Under the equivariance-free
prior of \(L\)-Lipschitz targets consistent on \(T\), the McShane
extensions
\(\Phi_\pm(z)=\min/\max_{t\in T}\big(\Phi(t)\pm L\,\mathrm{dist}(z,t)\big)\)
are admissible and differ at \(z\) by up to \(2L\,\mathrm{dist}(z,T)\),
so every learner has minimax error \(\ge L\,\mathrm{dist}(z,T)\) and
\(\mathcal C\subseteq\{z:\mathrm{dist}(z,T)\le\epsilon/L\}\) --- an
\(\epsilon/L\)-tube around the training set that \textbf{collapses to
\(T\) as \(\epsilon\to0\)}.

The separation is sharp in its dependence on \(\epsilon\): a point at
orbit-distance \(D\) from \(T\) --- the unseen far end of the wedge in
Experiments 7 and 9 --- is certified by \emph{structure} for
\textbf{every} \(\epsilon>0\), but by \emph{scale/data} only once
\(\epsilon\ge LD\). Scaling a non-equivariant model (more capacity, more
in-wedge data) sharpens its fit \emph{inside} the tube; it cannot
enlarge the tube, because the bound \(L\,\mathrm{dist}(z,T)\) is
information-theoretic (independent of model size). This is the precise
content of the empirical curves: the equivariant model is flat over the
whole orbit while every baseline scale climbs once
\(\mathrm{dist}(z,T)\) exceeds \(\epsilon/L\).

\subsubsection{3.4 The certificate is a
procedure}\label{the-certificate-is-a-procedure}

The certificate is not only a lens; it is something one \textbf{runs} on
a trained model, with no further data. The inputs are the two quantities
every equivariant model already exposes --- its per-generator
equivariance residual and its predictor-Jacobian spectrum --- and the
output is the certified region.

\begin{quote}
\textbf{Algorithm 1 (Certify).} \emph{Input:} trained equivariant model
\((E,f)\), generators \(S\), resolution \(\epsilon\). 1. Measure the
per-generator residual
\(\epsilon_{\max}=\max_{g_i\in S}\lVert E(g_i\!\cdot\!x)-\rho(g_i)E(x)\rVert\)
(and the analogous predictor residual). 2. \textbf{Configuration axis:}
if \(\epsilon_{\max}\le\text{tol}\), certify the \emph{entire} monoid
\(\langle S\rangle\) from these \(k\) checks (Lemma 1, Theorem A); else
report the approximate budget \(m\,\epsilon_{\max}\) (Theorem B). 3.
Measure the predictor-Jacobian singular values \(\{\sigma_j\}\) at the
operating point; set \(\lambda_j=\log\sigma_j/\Delta t\). 4.
\textbf{Horizon \(\times\) resolution axis:} per channel, form the
finite-time budget
\(B_j(T)=c_j\bigl[\epsilon_j e^{\lambda_j T}+\zeta_j\sum_{s=0}^{T-1}e^{\lambda_j s}\bigr]\)
and return
\(T_j(\tau_j)=\max\{T:\ B_j(t)\le\tau_j\ \text{for all }0\le t\le T\}\).
For \(\lambda_j>0\) this is \(\sim\log(1/\epsilon)/\lambda_j\). The
value \(T_j=\infty\) is returned \textbf{only} when the inequality is
certified for all \(T\) --- either because the channel is exact
(\(\epsilon_j=\zeta_j=0\)), or because a contracting channel's bounded
floor \(\zeta_j/(1-e^{\lambda_j})\) stays below \(\tau_j\) (a
tolerance-level certificate, \emph{not} zero-error all-horizon
prediction) (Theorem B).

\emph{Output:} the certified region
\(\langle S\rangle\times\{T_j(\epsilon)\}\) --- computed from the model
alone.
\end{quote}

This is implemented and unit-tested (\texttt{src/certify.py}): on the
Experiment-1 model it certifies the full monoid from \(2\) generator
checks (residual \(1.2\times10^{-7}\)) and returns \(48\) contractive
channels whose bounded floor stays below tolerance (a tolerance-level
certificate, not zero-error all-horizon prediction) plus the binding
expansive channel's \(\log(1/\epsilon)\) horizon. \emph{The reported
horizon is an interval, not a point:} the spectrum \(\{\lambda_j\}\) is
estimated with a block-bootstrap confidence interval calibrated against
the Liouville anchor \(\sum_j\lambda_j=-N\) where one exists
(\texttt{experiments/step78}, \texttt{tests/test\_step78.py}) --- which
captures estimation \emph{noise} but \textbf{not}
\emph{misspecification}: a reproducibly-wrong model carries tight bars
around a wrong spectrum (the dense MLP of §5.16 is the example), so the
structural prior --- not the error bar --- is what makes the recovered
spectrum trustworthy. \emph{Statistics quantifies estimation noise;
structure delivers correctness.} The certificate is thus an operational
tool, not merely an interpretive claim.

\begin{center}\rule{0.5\linewidth}{0.5pt}\end{center}

\subsection{4. The Noether Hinge}\label{the-noether-hinge}

Theorems A and B hold for \emph{whatever} channels happen to be slow.
What makes the configuration axis and the horizon axis the \textbf{same}
structure is a Noether-type bridge:

\begin{quote}
\textbf{Conjecture (group \(\Rightarrow\) slow).} When the symmetry is a
symmetry of the \textbf{dynamics} (not merely the encoder), the
group-invariant (\(\ell{=}0\)) channels coincide with the conserved/slow
(\(\lambda_j\le0\)) channels. Hence the symmetry that gives
across-configuration generalization is the \emph{same} structure that
gives long-horizon predictability.
\end{quote}

This is \textbf{not automatic} --- invariant \(\ne\) conserved in
general. We prove the \emph{forward} direction below (Proposition 5: a
conserved charge is certified to long horizons) and leave the rest
\emph{falsifiable and measured}. The defensible inclusion is
\textbf{slow \(\subseteq\) invariant \(\oplus\) conserved-equivariant},
not the converse: the conserved (slowest) modes live in the
invariant-or-conserved subspace, but not every invariant channel is slow
(a rotation-invariant \(|r|^2\) oscillates). The mechanism is Noether's.
A continuous symmetry of the dynamics yields a conserved current; a
conserved \emph{scalar} charge (\(\ell{=}0\) Casimir, e.g.~energy)
\textbf{is} group-invariant, while a conserved \emph{non-scalar} charge
(e.g.~three-dimensional angular momentum \(L\), an \(\ell{=}1\) vector)
is \emph{equivariant}, \textbf{not} invariant. The slow subspace
therefore lies in
\(\text{invariant}\oplus\text{conserved-equivariant}\), which
\textbf{collapses to the invariant component exactly when every
conserved charge is a scalar} --- the two-dimensional case, where
\(\text{slow}\subseteq\text{invariant}\) holds literally. Section 5.3
measures both forms: the clean \(2\)D containment, and its \(3\)D
refinement in which angular momentum is recovered from the equivariant
block at polynomial degree two (the cross product).

While the \emph{coincidence} slow \(=\) conserved is what we measure,
the \textbf{placement by isotypic type is forced by representation
theory} --- and so is the degree at which each charge is readable. This
is what makes the \(3\)D ``degree-2 cross product'' non-arbitrary.

\textbf{Proposition 4 (placement principle: which block carries which
charge, and at what degree).} Let \(G\) act on \(\mathcal Z\) by an
orthogonal \(\rho\) with isotypic decomposition
\(\mathcal Z=\bigoplus_\ell\mathcal Z_\ell\), and let \(C\) be a
conserved quantity transforming in a representation \(\tau\),
\(C(\rho(g)z)=\tau(g)C(z)\). Then every homogeneous component
\(C_k\in\mathrm{Hom}_G(\mathrm{Sym}^k\mathcal Z,\,W)\) of \(C\) is
supported on the \(\tau\)-isotypic part of \(\mathrm{Sym}^k\mathcal Z\)
(Schur). In particular: a conserved \textbf{scalar} (\(\tau\) trivial,
e.g.~energy) is an \emph{invariant} function whose degree-\(k\)
component reads from the trivial isotypic component of
\(\mathrm{Sym}^k\mathcal Z\) (the linear case \(k{=}1\) reduces to the
trivial block \(\mathcal Z_0\); a nonlinear invariant such as
\(\lVert v\rVert^2\) lives at \(k{=}2\)); and the conserved
\textbf{moment map} \(\mu:\mathcal Z\to\mathfrak g^*\) --- Noether's
charge of the continuous symmetry --- is equivariant in the adjoint
representation, which for \(\mathrm{SO}(3)\) is
\(\mathfrak{so}(3)^*\cong\mathcal V_1\), so angular momentum lives in
the \(\ell{=}1\) block. Since \(\mu\) is \textbf{quadratic} and bilinear
in the position--velocity pair, then --- \emph{provided the \(\ell{=}1\)
block carries (at least) two copies of \(\mathcal V_1\),
e.g.~\(\mathcal V_1(\text{pos})\oplus\mathcal V_1(\text{vel})\), as in
the model of Experiment 6} --- the cross product is, up to scale, the
\textbf{unique} \(\mathrm{SO}(3)\)-equivariant degree-2 readout, and no
degree-1 readout exists. \emph{Proof.} Differentiate the equivariance of
\(C_k\) and apply Schur (\(\mathrm{Hom}_G\) between non-isomorphic
isotypic types vanishes). For the bilinear \(\mu\) on
\(\mathcal Z\supseteq\mathcal V_1(\text{pos})\oplus\mathcal V_1(\text{vel})\),
the degree-2 cross term in \(\mathrm{Sym}^2\mathcal Z\) contains
\(\mathcal V_1\otimes\mathcal V_1=\mathrm{Sym}^2\mathcal V_1\oplus\Lambda^2\mathcal V_1\);
\(\mathrm{Sym}^2\mathcal V_1=\mathcal V_0\oplus\mathcal V_2\) carries
\textbf{no} \(\mathcal V_1\), while
\(\Lambda^2\mathcal V_1\cong\mathcal V_1\) with
\(\dim\mathrm{Hom}_{\mathrm{SO}(3)}(\Lambda^2\mathcal V_1,\mathcal V_1)=1\)
(classical \(\mathrm{SO}(3)\) invariant theory) --- realized by
\(r\times v\). So no degree-1 and no other degree-2 readout exists;
uniqueness needs the two \(\mathcal V_1\) copies (with \(m\) copies the
space is \(\binom{m}{2}\)-dim). \(\square\)

This \emph{predicts}, with no fitted parameter, exactly what §5.3
measures --- \(E\) from the invariant block (\(R^2\approx1\)), \(L\)
from the \(\ell{=}1\) block via the degree-2 cross product
(\(R^2{=}1.00\)) and nothing lower --- and explains the tie to the
companion paper's degree-1 Vector-Neuron cap (a degree-1 net
\emph{cannot} form the only available readout). \textbf{Scope (honest).}
Proposition 4 is Schur's lemma applied to the symmetric powers plus the
moment map's equivariance --- a \emph{placement principle}, not a new
theorem: it pins which block carries each charge and at what degree, but
representation theory alone does not force those charges to be the
\emph{dynamically slow} modes. What representation theory does not give,
a one-line dynamical argument does --- the \emph{forward} direction of
the hinge.

\textbf{Proposition 5 (conservation defect controls charge-value
drift).} Let \(Q:\mathcal Z\to W\) be a charge read-out and
\(\mathcal K\subseteq\mathcal Z\) a region forward-invariant under both
the true dynamics \(\Phi\) and the model \(f\). Suppose \(Q\) is a first
integral of the truth, \(Q\circ\Phi=Q\) on \(\mathcal K\), and the model
conserves it up to a one-step defect
\(\eta:=\sup_{z\in\mathcal K}\lVert Q(f(z))-Q(z)\rVert_W\). Then the
model's \(T\)-step error \emph{in the charge value} is at most
\textbf{linear} in the horizon, \[
\bigl\lVert Q(f^{T}z)-Q(\Phi^{T}z)\bigr\rVert=\bigl\lVert Q(f^{T}z)-Q(z)\bigr\rVert\;\le\;T\,\eta
\qquad(\forall z\in\mathcal K,\ T\ge0),
\] so the charge-value error grows \textbf{at most linearly in \(T\)}
(not the chaotic \(e^{\lambda T}\) of an expansive channel), and its
certified horizon is \(T_Q(\epsilon)\ge\epsilon/\eta\) ---
\emph{infinite} at exact conservation \(\eta=0\) --- irrespective of the
ambient dynamics' largest Lyapunov exponent. (This is a statement about
the \emph{charge value}; a positive-exponent \emph{shear} off-diagonal
can still leave the full state on the conserved subspace large --- see
§7.) \emph{Proof.} Telescope
\(Q(f^{T}z)-Q(z)=\sum_{k<T}\!\bigl[Q(f(f^{k}z))-Q(f^{k}z)\bigr]\);
forward-invariance keeps each \(f^{k}z\in\mathcal K\) so each term is
\(\le\eta\), while \(\Phi\)-invariance of \(\mathcal K\) gives
\(Q(\Phi^{T}z)=Q(z)\). \(\square\)

The content is the \textbf{additive-versus-multiplicative} contrast: a
conserved charge's error \emph{accumulates} (\(T\eta\)) where a generic
channel's \emph{compounds} (\(e^{\lambda_jT}\), Theorem B). It is a
statement about the \textbf{charge value}, not the full state ---
\(f^{T}z\) and \(\Phi^{T}z\) may separate at the ambient rate while
their \(Q\)-images stay \(T\eta\)-close --- which is exactly what a
certificate for \emph{``predict the conserved quantity''} needs (it does
\textbf{not} claim the predictor-Jacobian's singular values on the
conserved subspace are near one; off-diagonal shear can leave those
large). With Proposition 4 this closes the \textbf{forward} direction of
the hinge: Noether's charge \(\mu\) lives in the invariant/equivariant
blocks (Prop 4) \textbf{and} is certified to long horizons (Prop 5), so
those blocks are the slow ones.

\textbf{The hinge as a theorem, and its honest residue.} Under the
hinge's own hypothesis --- that \(G\) is a symmetry of the
\emph{dynamics} --- assemble the three pieces: if the latent flow is
Hamiltonian with a \(G\)-invariant Hamiltonian, then (Noether) the
moment map \(\mu\) obeys \(Q\circ\Phi=Q\); Proposition 4 places \(\mu\)
in \(\mathcal Z_0\oplus(\text{adjoint})\); Proposition 5 certifies it.
So \emph{under symplectic structure} ``the invariant/equivariant blocks
are slow'' is a \textbf{theorem}, not a coincidence. What stays assumed
or measured, in order: \textbf{(i)} that the \emph{learned latent} flow
is Hamiltonian with \(G\)-invariant \(H\) --- i.e.~the encoder is
essentially a symplectomorphism --- is a structural assumption we do
\textbf{not} enforce; \textbf{(ii)} exact conservation (\(\eta=0\))
holds only for the momentum map under a \(G\)-equivariant
\emph{symplectic discretization} --- energy is conserved only to
\(O(\Delta t^{p})\) even by symplectic integrators (they preserve a
\emph{modified} Hamiltonian), and a generic learned \(f\) is neither
symplectic nor equivariant, so \(\eta\) is \textbf{measured} (§5.3 is
exactly that measurement); \textbf{(iii)} the \textbf{converse fails}
--- a slow channel need not be conserved (a rotation-invariant \(|r|^2\)
oscillates) --- so the inclusion
\(\text{slow}\subseteq\text{invariant}\oplus\text{conserved-equivariant}\)
is one-directional. The upgrade over a pure conjecture is precise:
\emph{conserved \(\Rightarrow\) slow} is now \textbf{proved} (Prop 5, at
the charge-readout level), with the conservation defect \(\eta\) --- not
a positive Lyapunov exponent --- bounding the drift \emph{linearly};
only the dynamical-symmetry hypothesis and the value of \(\eta\) remain
empirical.

\begin{center}\rule{0.5\linewidth}{0.5pt}\end{center}

\subsection{5. Experiments}\label{experiments}

All experiments are CPU/\(1\)-GPU-scale, run with explicit random seeds,
and \textbf{gated}: a run reports \texttt{INCONCLUSIVE} rather than
loosen a threshold. Load-bearing results are additionally guarded by
unit tests. The experiment-to-code map, seed counts, and headline
numbers are collected in Appendix A; multi-seed ranges below are seed
min--max over three seeds unless noted.

The load-bearing experiments and the claims they test, at a glance:

{\def\LTcaptype{none} 
\begin{longtable}[]{@{}
  >{\raggedright\arraybackslash}p{(\linewidth - 8\tabcolsep) * \real{0.2000}}
  >{\raggedright\arraybackslash}p{(\linewidth - 8\tabcolsep) * \real{0.2000}}
  >{\raggedright\arraybackslash}p{(\linewidth - 8\tabcolsep) * \real{0.2000}}
  >{\raggedright\arraybackslash}p{(\linewidth - 8\tabcolsep) * \real{0.2000}}
  >{\raggedright\arraybackslash}p{(\linewidth - 8\tabcolsep) * \real{0.2000}}@{}}
\toprule\noalign{}
\begin{minipage}[b]{\linewidth}\raggedright
§
\end{minipage} & \begin{minipage}[b]{\linewidth}\raggedright
Axis / claim tested
\end{minipage} & \begin{minipage}[b]{\linewidth}\raggedright
System
\end{minipage} & \begin{minipage}[b]{\linewidth}\raggedright
Headline result
\end{minipage} & \begin{minipage}[b]{\linewidth}\raggedright
Backed by
\end{minipage} \\
\midrule\noalign{}
\endhead
\bottomrule\noalign{}
\endlastfoot
5.1 & Configuration axis is \emph{exponential} & \(\mathbb{Z}_2^6\)
compositional & \(6\) generator checks certify all \(2^6{=}64\)
compositions; baseline degrades off-generator & Lemma 1 \\
5.2 & Horizon \(\times\) resolution staircase & controlled spectrum &
\(T_j(\epsilon)\sim\log(1/\epsilon)/\lambda_j\) recovered per channel &
Theorem B \\
5.4 & Structure \emph{vs.} scale (config axis) & orbit-flatness &
equivariant orbit-flat; an \(88\times\) baseline stays
\(10\)--\(155\times\) above the floor OOD & Lemma 2 \\
5.5 & Approximate-symmetry degradation & broken-symmetry world &
graceful, linear in \(\epsilon_{\text{world}}\) to a measured threshold
& Proposition 6 \\
5.7 & Certificate on \emph{real contact dynamics} & PushT (physics
engine) & learned model orbit-flat (ratio \(1.000\)); \(160\times\)
baseline keeps a \(2\)--\(3\times\) OOD penalty & Theorem A \\
5.9 & Certificate \(\to\) \emph{task} competence & closed-loop control &
orbit-invariant pose control to the float floor; scaled baseline
degrades & Theorem A \(+\) (A5) \\
5.10--5.11 & Not \(\mathrm{SO}(2)\)-specific; pixels &
\(\mathrm{SO}(3)\) point clouds, raw pixels & ratio \(1.000\); frame
averaging is \emph{accuracy-neutral} & Theorem A \\
5.12--5.13 & Horizon law on \emph{real chaos} & learned Lorenz / Hénon /
Rössler & learned \(\lambda_1\) matches textbook to \(1\)--\(12\%\)
(\(R^2{=}0.96\)--\(1.00\)) & Thm B, Props 7--8 \\
5.14--5.15 & Standard benchmark \(+\) planning & MuJoCo FetchPush &
exactly orbit-flat; \(7\times\) baseline degrades OOD; equivariant
planner provably orbit-flat & Theorem A \\
5.16 & \textbf{Structure beats scale at high \(N\)} & \(40\)-D Lorenz-96
& \(\mathbb{Z}_N\)-conv recovers the full spectrum
(\(R^2{=}0.98\)--\(0.99\)); dense MLP \textbf{and} a same-trained
recurrent GRU fail (\(R^2{<}0\)) & Thm B \(+\) conditional-Lyapunov \\
\end{longtable}
}

\subsubsection{5.1 The configuration axis}\label{the-configuration-axis}

\textbf{Experiment 1 (exact compositional flatness).} On a structured
object-interaction world with \(\mathrm{SE}(3)\times S_n\) symmetry, the
equivariant model's whole-pipeline relMSE is \textbf{exactly
\(\times1.00\) flat over composition words \(m=0,\dots,8\)} (constant
\(0.2625\)), while a matched non-equivariant baseline blows up by
\(\times12.89\). The predictor's Jacobian spectrum is measured directly:
\textbf{\(48\) of \(96\) channels are contractive} (\(\sigma\le1\),
certified long-horizon) versus \(48\) expansive (\(\max\sigma=49\),
\(\min\sigma=0.003\)) --- the clean coarse/fine split that is exactly
Theorem B's input. A unit test verifies Theorem A holds
\emph{architecturally} at initialization (word-deviation
\(1.2\times10^{-7}\), before any training) while the control deviates
\(20\%\).

\textbf{Experiment 2 (exponential certification on \(\mathbb{Z}_2^6\)).}
The \(64\) hexagrams of the \emph{I Ching} are exactly
\(\{\pm1\}^6=\mathbb{Z}_2^6\), with the \(6\) changing-lines as a
generating set. Training a sign-equivariant per-line predictor
\(f(z,a)_i=h_\theta(a_i)\odot z_i\) on \textbf{only the \(6\)
single-line generators} (plus the identity, \(7\) of \(64\) actions)
certifies it over \textbf{all \(2^6{=}64\) compositions} to machine
precision (worst relMSE \(\sim10^{-33}\); equivariance residual exactly
\(0\)), while a non-equivariant baseline trained on the same \(7\)
actions degrades monotonically with composition length
(\(1.6\text{–}1.7\times10^{-5}\) on seen actions \(\to 0.59\) at six
flips, Figure 2). This makes ``\(k\) generator checks
\(\Rightarrow 2^k\) certified set'' literal, with genuine learning and
an honest contrast.

\begin{figure}
\centering
\pandocbounded{\includegraphics[keepaspectratio,alt={Training on the 6 generators of \textbackslash mathbb\{Z\}\_2\^{}6 certifies all 64 compositions (equivariant, machine precision) while a non-equivariant baseline degrades with composition length.}]{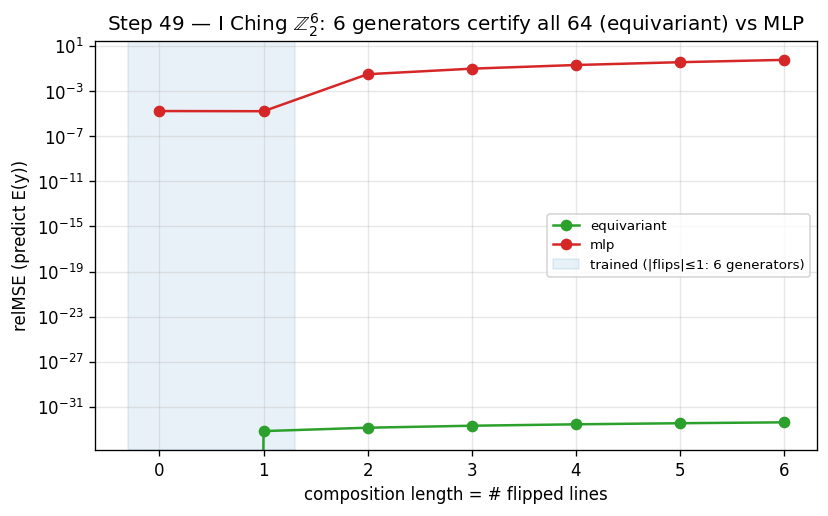}}
\caption{Training on the \(6\) generators of \(\mathbb{Z}_2^6\)
certifies all \(64\) compositions (equivariant, machine precision) while
a non-equivariant baseline degrades with composition length.}
\end{figure}

\subsubsection{\texorpdfstring{5.2 The horizon \(\times\) resolution
staircase}{5.2 The horizon \textbackslash times resolution staircase}}\label{the-horizon-times-resolution-staircase}

\textbf{Experiment 3 (the predictability staircase).} A learned one-step
predictor on a designed multi-Lyapunov system --- a conserved invariant
(\(\lambda=-\infty\)), a contracting channel (\(\lambda=\ln0.75\)), a
neutral \(\mathrm{SO}(2)\) rotor (\(\lambda=0\)), and a chaotic
angle-doubling channel \(z\mapsto z^2\) (\(\lambda=\ln2\)) --- is rolled
out autoregressively. The model \textbf{recovers the chaotic Lyapunov
exponent to within \(0.4\%\)} (\(\hat\lambda=0.690\text{–}0.692\) across
three seeds, versus \(\ln2=0.693\)), and its per-channel certified
horizon obeys \(T_j(\epsilon)\sim\log(1/\epsilon)/\lambda_j\): the
chaotic ``fine detail'' is certifiable for only \(3\text{–}10\) steps
over the full \(\epsilon\) grid, and its horizon grows \emph{only
logarithmically} as \(\epsilon\) coarsens (measured slope
\(\mathrm{d}T/\mathrm{d}\log\epsilon=1.3\text{–}1.6\approx1/\lambda=1.44\)),
while the \textbf{contracting channel stays certified for the entire
\(90\)-step rollout at every \(\epsilon\)} (the conserved and rotor
channels too, except at the finest \(\epsilon{=}0.005\), where they fall
to \(\sim40\text{–}90\) steps). At \(\epsilon{=}0.05\) the slow channels
outlast the chaotic detail by \(12\text{–}15\times\) (Figure 3). This is
the predictability staircase: long horizon \(\Rightarrow\) coarse and
invariant only.

\begin{figure}
\centering
\pandocbounded{\includegraphics[keepaspectratio,alt={Left: rollout error growth recovers the Lyapunov spectrum. Right: the certified-horizon staircase T\_j(\textbackslash epsilon)\textbackslash sim\textbackslash log(1/\textbackslash epsilon)/\textbackslash lambda\_j.}]{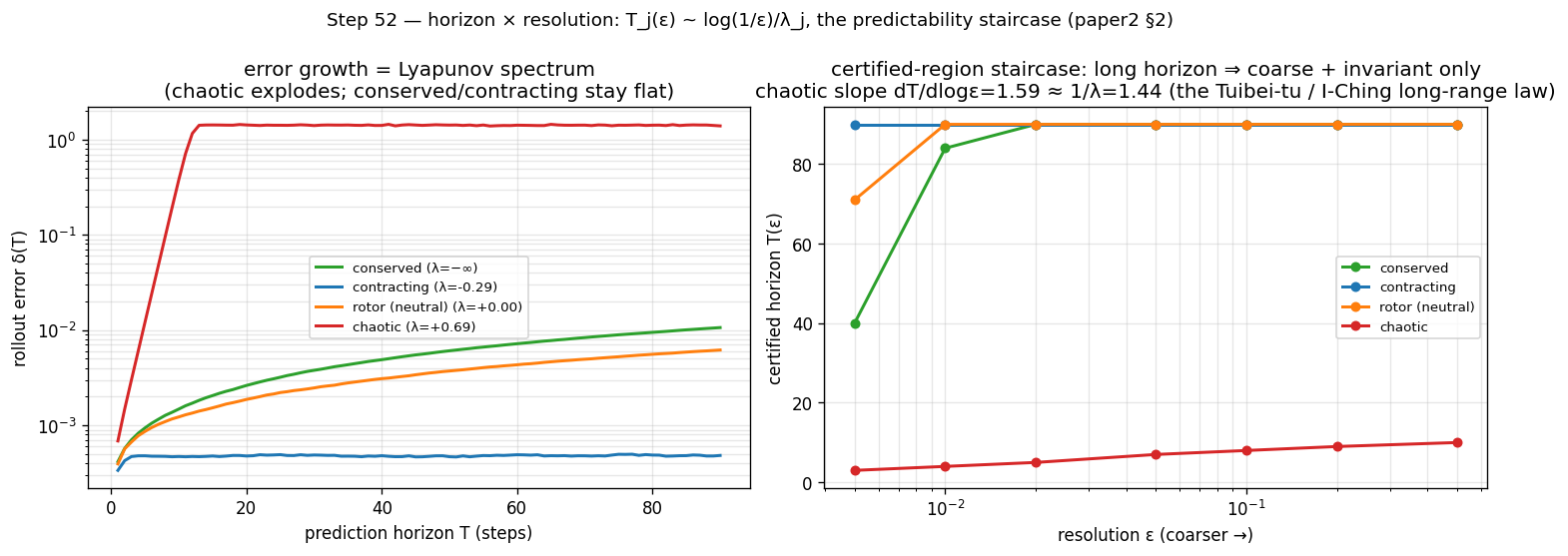}}
\caption{Left: rollout error growth recovers the Lyapunov spectrum.
Right: the certified-horizon staircase
\(T_j(\epsilon)\sim\log(1/\epsilon)/\lambda_j\).}
\end{figure}

\subsubsection{5.3 The Noether hinge}\label{the-noether-hinge-1}

\textbf{Experiment 4 (the hinge in 2D).} On a two-dimensional
\(\mathrm{SO}(2)\) central-force system --- conserved energy \(E\) and
angular momentum \(L\) are invariant scalars, the orbital phase is fast
--- a learned equivariant autoencoder-with-latent-dynamics (a hand-built
\(2\)D Vector-Neuron model) spontaneously organizes the conserved
quantities into its architecturally-invariant (\(\ell{=}0\)) block.
Regression \(R^2\) for \((E,L)\) is \textbf{\(0.92\text{–}0.99\) from
the invariant block versus \(\le0.012\) from the equivariant
(\(\ell{=}1\)) block}, and the slowest mode the invariant block admits
(\(0.009\text{–}0.031\) across seeds) is \(4.7\text{–}17\times\) slower
than anything the equivariant block admits (\(0.145\), exactly the
orbital phase velocity \(2\sin(\omega\Delta t/2)\)). We read the
containment off this min-mode gap --- the equivariant block's
\emph{slowest} achievable mode is already fast, so no slow direction
escapes the invariant block in this system --- rather than by a direct
subspace test. The payoff is the certificate: the equivariant model's
slow subspace \textbf{is} its invariant subspace, whose
out-of-distribution group-action residual is \(\sim10^{-16}\), exact and
architectural for all of \(\mathrm{SO}(2)\). A matched non-equivariant
baseline also learns \(E,L\) (\(R^2=0.99\)) but smears them across
directions that \textbf{drift by \(\approx1.17\)} under the same group
action, and carries no certificate (Figure 4). The honest non-converse
is visible in the data (the invariant but fast scalar \(|r|^2\)): slow
\(\subseteq\) invariant, not \(=\).

\begin{figure}
\centering
\pandocbounded{\includegraphics[keepaspectratio,alt={Left: the slowest latent modes live in the invariant (\textbackslash ell\{=\}0) block. Right: the certificate --- the invariant subspace stays group-invariant to 10\^{}\{-16\} where the non-equivariant model's slow directions drift by \textbackslash sim1.}]{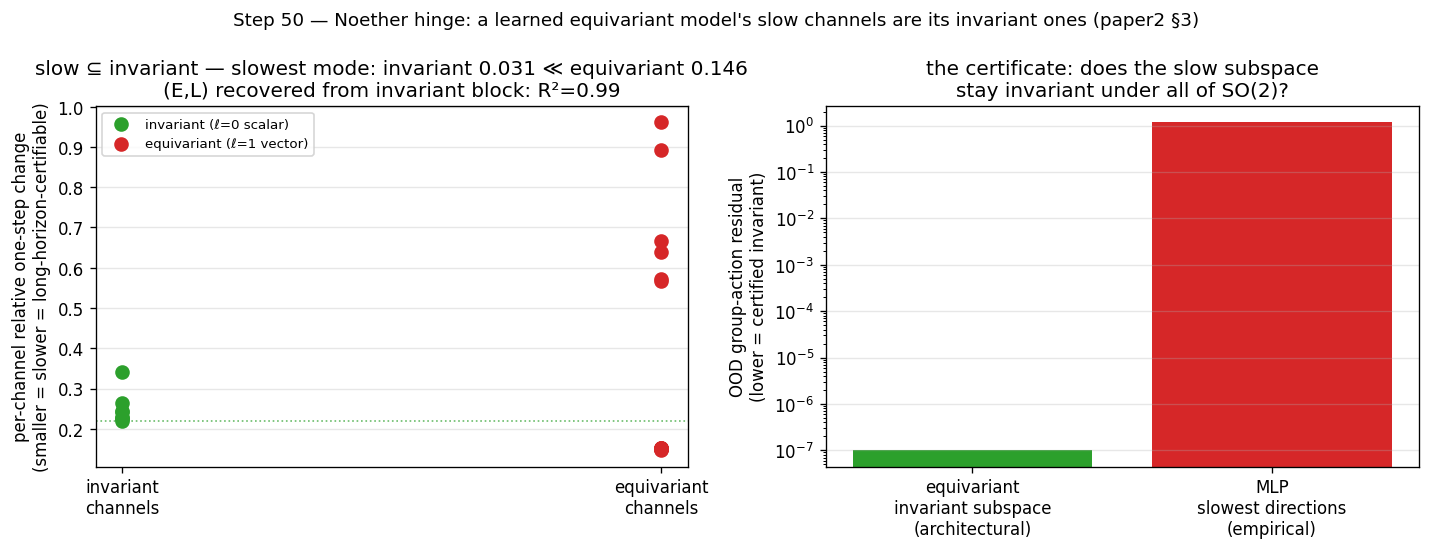}}
\caption{Left: the slowest latent modes live in the invariant
(\(\ell{=}0\)) block. Right: the certificate --- the invariant subspace
stays group-invariant to \(10^{-16}\) where the non-equivariant model's
slow directions drift by \(\sim1\).}
\end{figure}

\textbf{Experiment 5 (lift to a 3D contact interaction).} Pushing the
hinge to a more embodied regime --- two bodies in a three-dimensional
well with a \textbf{soft pairwise repulsion (contact)}, modeled by an
\(\mathrm{SO}(3)\)-equivariant \(\ell{=}0\oplus\ell{=}1\) Vector-Neuron
--- gives a \emph{partial} lift, honestly reported. The
\emph{Noether-content} half lifts: the invariant \(\ell{=}0\) block
recovers the conserved \((E,\text{contact-distance})\) at
\(R^2{=}0.60\text{–}0.86\) versus \(\le0.06\) for the \(\ell{=}1\)
block, even with contact. But the clean containment
\(\text{slow}\subseteq\text{invariant}\) does \textbf{not} lift
(invariant slowest mode \(0.024\approx0.026\) for the equivariant block;
this gate is reported \texttt{INCONCLUSIVE}) --- for a principled
reason: in \(3\)D, angular momentum \(L=\sum_i r_i\times v_i\) is a
conserved \(\ell{=}1\) \emph{vector}, so the equivariant block too
carries a slow conserved mode. The clean \(2\)D containment (where \(L\)
is a scalar) becomes, in \(3\)D,
\(\text{slow}\subseteq(\text{invariant}\oplus\text{conserved-equivariant})\).
The Noether-content claim is dimension-robust; the clean-containment
form is \(2\)D-specific --- a sharpening of scope, not a failure.

\textbf{Experiment 6 (3D-aware containment).} The conserved physics
splits by isotypic \emph{type} \textbf{and} polynomial \emph{degree}:
\(E\) (an invariant quadratic) is recovered linearly from the
\(\ell{=}0\) block (\(R^2{=}0.62\text{–}0.91\) across seeds), whereas
\(L=\sum_i r_i\times v_i\) (a conserved \(\ell{=}1\) vector) is
\textbf{bilinear} --- not linear in either block (\(\le0.04\)) but
recovered at \(R^2{=}1.00\) (range \(0.998\text{–}1.000\), the robust
load-bearing result) by the \textbf{degree-two cross-product} readout of
the \(\ell{=}1\) block. So
\(\text{slow}\subseteq(\text{invariant}\oplus\text{conserved-equivariant})\)
holds exactly in \(3\)D, with the equivariant conserved part accessed at
degree two --- and \(r\times v\) is exactly the cross product a
degree-one Vector-Neuron cannot form, tying the hinge's \(3\)D form to
the companion paper's bilinear-message result. The full gate, which
additionally demands \(E\)-recovery \(R^2{>}0.8\) in the \(\ell{=}0\)
block, passes on one of three seeds --- as honestly surfaced as the
\texttt{INCONCLUSIVE} gate of Experiment 5 --- while the degree-two
\(L\) recovery is robust across all three.

\subsubsection{5.4 Structure versus scale}\label{structure-versus-scale}

\textbf{Experiment 7 (structure versus scale).} Train on a \(50^\circ\)
wedge of an \(\mathrm{SO}(2)\) orbit and test around the full circle.
The exactly-equivariant model's error is \textbf{flat over the entire
orbit} (out-of-wedge / in-wedge ratio \(1.1\text{–}1.2\)) --- a
certificate that holds from partial data by the identity
\(\mathrm{err}(R_\theta s)=\mathrm{err}(s)\). A non-equivariant baseline
\textbf{scaled across an \(88\times\) parameter range}
(\(3.8\mathrm{k}\to337\mathrm{k}\)) buys genuine \emph{interpolation}:
its in-wedge error drops \(31\text{–}166\times\) and even \textbf{beats}
the \(56\times\)-smaller equivariant model in-distribution --- but out
of the wedge the largest baseline remains \(10\text{–}155\times\) worse
than the equivariant model and never reaches its floor (Figure 5). The
fair nuance --- \emph{scale can win in-distribution} --- is exactly why
the contribution is the \textbf{certificate} (an orbit-wide guarantee
from partial data), not a per-point accuracy contest. This is a gap
versus \emph{scale} (more parameters, the same data); the complementary
question --- whether \emph{data augmentation} (orbit-covering rotations)
closes it --- is answered separately and honestly in §5.8 (it does, on a
single orbit; the certificate's edge is then exactness, the a-priori
guarantee, and worst-case optimality).

\begin{figure}
\centering
\pandocbounded{\includegraphics[keepaspectratio,alt={Left: error over the \textbackslash mathrm\{SO\}(2) orbit (equivariant flat; the scaled baselines dip below it in-wedge then climb out). Right: out-of-wedge error versus baseline scale plateaus far above the equivariant floor.}]{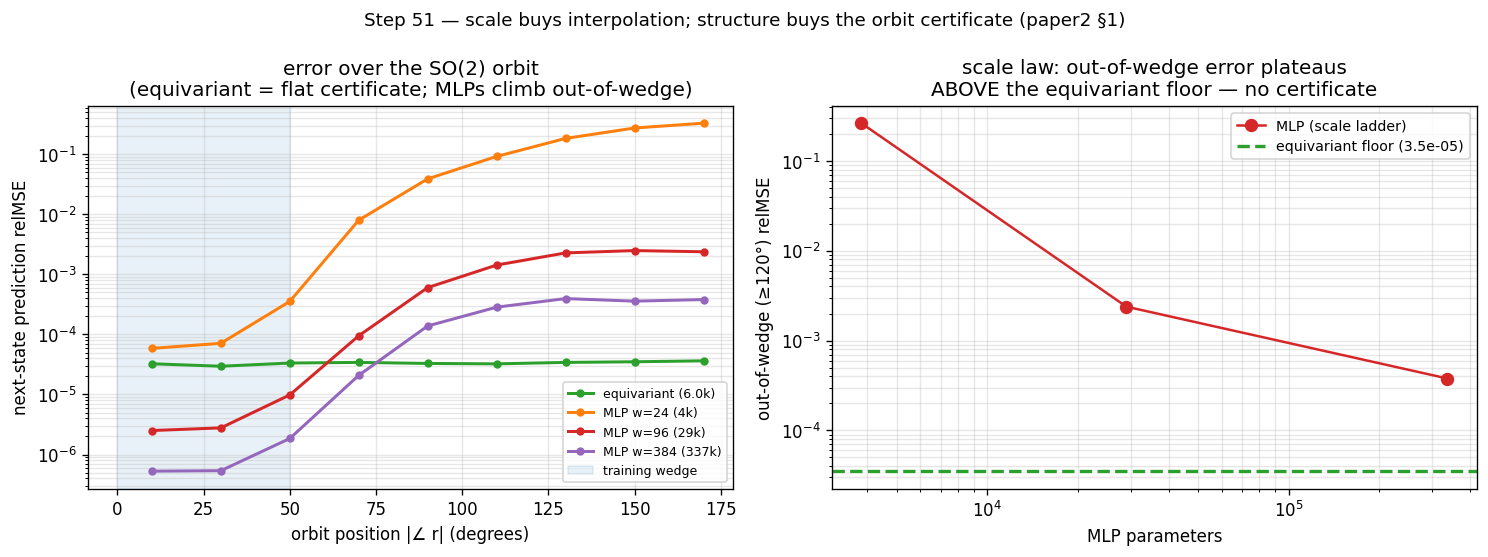}}
\caption{Left: error over the \(\mathrm{SO}(2)\) orbit (equivariant
flat; the scaled baselines dip below it in-wedge then climb out). Right:
out-of-wedge error versus baseline scale plateaus far above the
equivariant floor.}
\end{figure}

\subsubsection{5.5 Approximate symmetry}\label{approximate-symmetry}

\textbf{Experiment 8 (graceful degradation and a measured threshold).}
We break the \emph{world's} \(\mathrm{SO}(2)\) symmetry with an
anisotropy knob \(\beta\) (the potential becomes
\(V=\tfrac12(x^2+(1+2\beta)y^2)\), so angular momentum is no longer
conserved) and re-run the wedge-train / full-circle-test. Three
findings, robust across seeds: (i) at \(\beta{=}0\) the certificate is
exact --- the equivariant model is \textbf{\(68\text{–}320\times\)
better out-of-wedge} than the baseline; (ii) as the measured world
symmetry-defect \(\epsilon_{\text{world}}\) grows, the equivariant
out-of-wedge error grows \textbf{smoothly and monotonically}
(correlation \(0.88\text{–}0.98\), exactly Theorem B's
\(\epsilon_{\max}\) term --- a graceful slope, not a cliff); and (iii)
the equivariant model keeps beating the non-equivariant one up to a
\textbf{measured symmetry-content threshold}
(\(\epsilon_{\text{world}}\approx0.01\text{–}0.06\), seed-dependent),
beyond which the now-wrong symmetry assumption hurts more than it helps
(Figure 6). So \emph{approximate} symmetry buys an \emph{approximate}
certificate with exactly the error budget the theory predicts, and the
boundary where structure stops paying is itself measured, not assumed.

\begin{figure}
\centering
\pandocbounded{\includegraphics[keepaspectratio,alt={Left: equivariant out-of-wedge error grows \textbackslash propto\textbackslash epsilon\_\{\textbackslash text\{world\}\} (Theorem B's \textbackslash epsilon term). Right: the equivariant model beats the baseline out-of-wedge up to a symmetry-content threshold, then crosses over.}]{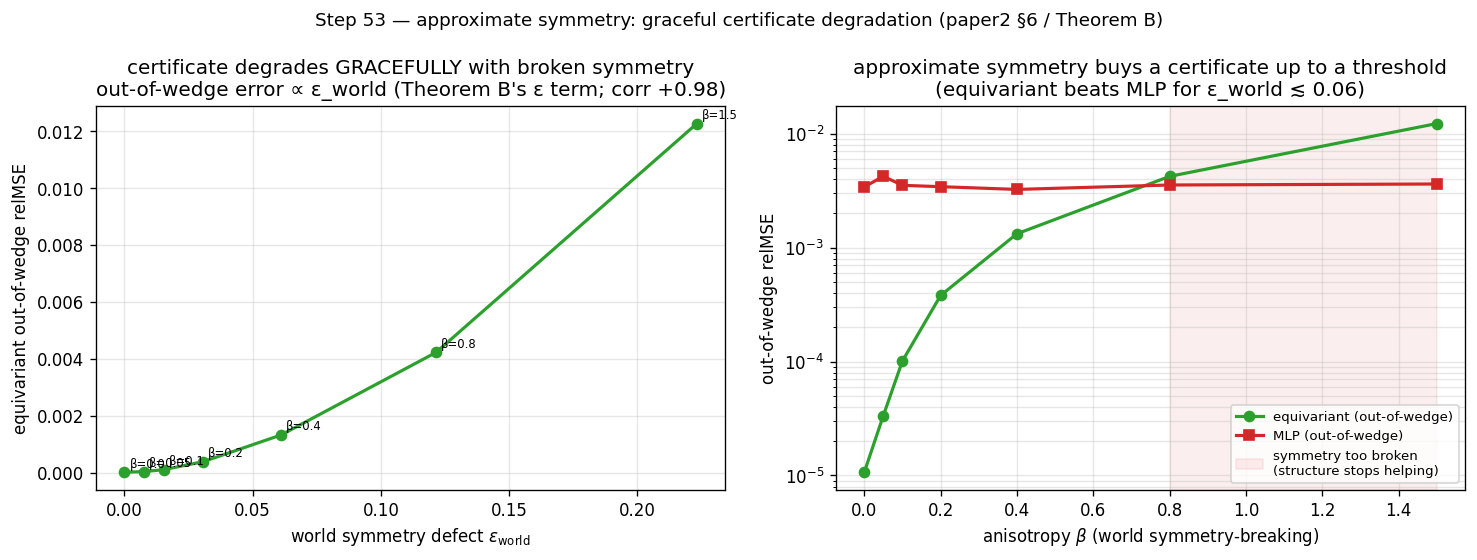}}
\caption{Left: equivariant out-of-wedge error grows
\(\propto\epsilon_{\text{world}}\) (Theorem B's \(\epsilon\) term).
Right: the equivariant model beats the baseline out-of-wedge up to a
symmetry-content threshold, then crosses over.}
\end{figure}

\subsubsection{5.6 From certificate to
action}\label{from-certificate-to-action}

A certificate is most useful if one need neither know the group in
advance nor a decoder to act on it. Two results, re-framed from a
companion line of experiments (not fresh runs here), close that loop.
\textbf{Discovery:} from a blank slate of \(K\) learnable \(3\times3\)
matrices --- nothing antisymmetric or bracket-closing imposed ---
gradient descent \emph{rediscovers} a frozen teacher's symmetry algebra
(\(\mathfrak{so}(3)\), dimension \(3\), on a rotationally-symmetric
teacher; the smaller \(\mathfrak{so}(2)_z\), dimension \(1\), on a
rotation-broken one --- it reports the \emph{surviving} group and
refuses to invent the rest), and distilling those discovered generators
into a free predictor buys back most of the across-group generalization
the hard-wired equivariant model gets for nothing (closing more than
half the out-of-distribution gap, matching the hand-wired oracle,
transferring \emph{exactly} the symmetry discovered). So the generating
set \(S\) that anchors the configuration axis need not be postulated ---
it can be \emph{measured}, falsifiably. \textbf{Generation:} given
\(S\), one can traverse the certified orbit to an unseen composed goal
and execute it \textbf{closed-loop without a decoder} --- latent-goal
reaching driven purely by the equivariance identity over \(24\) paired
seen-vs-unseen \(\mathrm{SE}(3)\)-orbit tasks, closing \(\sim0.59\) of
the orientation gap on the best deployable variant and --- the
load-bearing point --- achieving the \emph{same} success on the unseen
orbit as on the seen one (unseen/seen ratio \(1.000\)): \emph{exact
transfer}, not high absolute success. Together: \emph{discover \(S\)
\(\to\) certify \(\langle S\rangle\) \(\to\) generate \(\to\) act} ---
the certificate is not merely descriptive but a usable plan-and-execute
criterion. These two results are re-framed from a companion line; §5.9
closes the same loop on a \emph{fresh} run, with a planner driving a
learned model of \textbf{real physics-engine contact dynamics}.

\subsubsection{5.7 Beyond toys: the certificate on real contact
dynamics}\label{beyond-toys-the-certificate-on-real-contact-dynamics}

\textbf{Experiment 9 (PushT, real physics-engine contact dynamics).}
Every experiment so far uses dynamics we \emph{constructed} to be
equivariant --- the standing concern that the symmetry is built into a
toy. We close that gap on \textbf{PushT}, a planar pushing benchmark
whose contact dynamics are produced by a 2D physics engine we did not
author. In the interior (the block kept away from the workspace walls)
the dynamics are genuinely \(\mathrm{SO}(2)\)-equivariant --- rotating
the whole scene rotates the physics. On a \(\pm50^\circ\) wedge of real
interior transitions we train an \(\mathrm{SO}(2)\)-equivariant world
model --- an invariant-scalar-gated Vector-Neuron, exactly equivariant
to residual \(\sim10^{-7}\), so its contact features (agent--block
distance and contact angle) live in the invariant pathway --- and, as a
\emph{fair} baseline, non-equivariant MLPs across a \(160\times\)
parameter ladder (\(1.7\mathrm{k}\to272\mathrm{k}\)); both are trained
one-step with identical data, optimiser, and cosine schedule. We then
roll each model out \(H{=}10\) steps and measure rollout relMSE across
the full orbit of scene orientations (rotating a held-out real test set,
as in Experiment 7).

Three findings, robust across 3 seeds (Figure 7): (i) the equivariant
model's rollout error is \textbf{exactly flat over the orbit}
(out-of-wedge / in-wedge ratio \(1.00\) at every horizon) --- the
certificate, now on a \emph{learned} model of \emph{real} contact
physics; (ii) it is \textbf{competitive in-distribution} (in-wedge
relMSE \(0.13\text{–}0.15\) vs the best MLP's \(0.14\text{–}0.19\)), so
the certificate is not bought by being a worse predictor --- ``flat'' is
also ``good'' here; and (iii) \textbf{no MLP scale reaches the
equivariant floor out-of-wedge} --- every scale across the ladder stays
\(2.1\text{–}3.9\times\) above it at \(H{=}10\), and the
\(272\mathrm{k}\)-parameter MLP (\(16\times\) the equivariant model's
size) --- the \emph{closest} of the baselines --- is still
\(2.1\text{–}2.6\times\) clear of the floor. (These three findings hold
on all three seeds; the experiment's \emph{auxiliary} \texttt{climb}
sub-check --- the largest MLP's own in→out ratio exceeding \(2\times\)
--- is brittle, met on \(1/3\) seeds, so we report the load-bearing
floor-penalty rather than that sub-check, and the committed gate prints
\texttt{INCONCLUSIVE} on it.) The gap is smaller than the toy's because
a single PushT step is easy to predict in-distribution; the
certificate's value is precisely that it survives the \emph{rollout},
out of distribution, where scale does not follow. This is the keystone
non-toy result: the configuration certificate holds for a learned model
of dynamics we did not design.

\begin{figure}
\centering
\pandocbounded{\includegraphics[keepaspectratio,alt={Experiment 9 (PushT, real contact dynamics). Left: 10-step rollout relMSE over the orbit of scene orientations --- the learned \textbackslash mathrm\{SO\}(2)-equivariant model is exactly flat (ratio 1.00) and competitive in-distribution, while non-equivariant baselines dip in-wedge then climb out. Right: across a 160\textbackslash times parameter ladder no baseline reaches the equivariant floor out-of-wedge, at any rollout horizon.}]{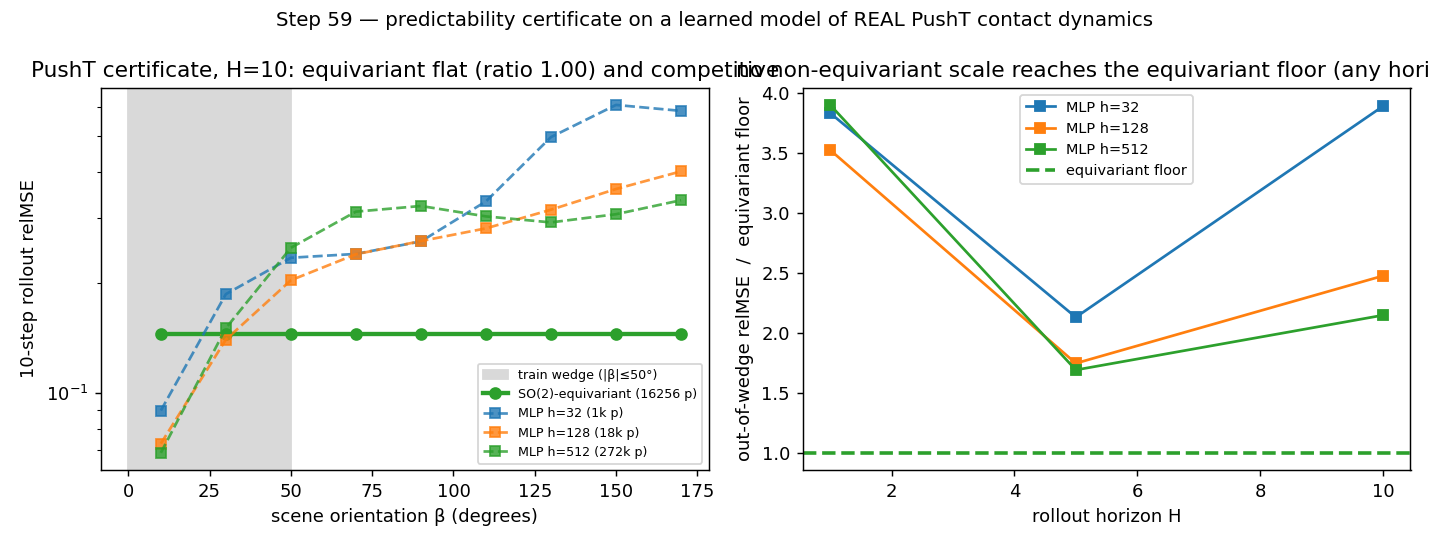}}
\caption{Experiment 9 (PushT, real contact dynamics). Left: \(10\)-step
rollout relMSE over the orbit of scene orientations --- the learned
\(\mathrm{SO}(2)\)-equivariant model is exactly flat (ratio \(1.00\))
and competitive in-distribution, while non-equivariant baselines dip
in-wedge then climb out. Right: across a \(160\times\) parameter ladder
no baseline reaches the equivariant floor out-of-wedge, at any rollout
horizon.}
\end{figure}

\subsubsection{5.8 Augmentation versus the
certificate}\label{augmentation-versus-the-certificate}

\textbf{Experiment 10 (the augmentation baseline --- the sharpest
objection, answered honestly).} The strongest reply to §5.4/§5.7 is that
a non-equivariant model with \textbf{data augmentation} (training on
rotated copies) buys orbit-coverage cheaply, so the gap is an artifact
of not augmenting. We test this on two axes and report what we find, not
what we hoped.

\emph{Single orbit (PushT) --- the concession.} Adding an
\(\mathrm{SO}(2)\)-augmentation arm to the Experiment-9 protocol, the
augmented MLP's \(10\)-step rollout error becomes \textbf{flat over the
orbit} (out/in ratio \(0.93\)--\(1.02\) across 3 seeds, versus the
un-augmented MLP's \(1.84\)--\(2.75\)). On a single continuous orbit,
\textbf{augmentation does match the certificate} --- we state this
rather than hide it.

\emph{Composition axis (\(\mathbb{Z}_2^6\)).} Here augmentation means
training the non-equivariant MLP on more of the \(64\) words. The
dynamics is smooth (bilinear \(a\odot z\)), so a well-trained MLP
generalizes from \(\sim\)half the words to a \(\sim10^{-4}\) floor ---
again, augmentation is a \emph{strong} baseline. But it \textbf{never
reaches the certificate's exactness}: the equivariant model is
machine-exact (\(\sim10^{-32}\)) from the \(7\) generators,
\(\sim10^{28}\times\) below the MLP's approximation floor, and is
\emph{certified a priori} (Theorem A, Lemma 2) without ever testing the
unseen compositions, whereas augmentation's success is only
\emph{confirmed by testing} the held-out words.

So the honest separation is \textbf{not} ``structure generalizes,
augmentation does not.'' It is that the certificate is \textbf{exact,
a-priori, group-knowledge-free, and worst-case-optimal} (the
\(\epsilon/L\)-tube of §3.3 holds against \emph{any} \(L\)-Lipschitz
dynamics), whereas augmentation is an \textbf{approximate, post-hoc,
group-informed average-case} improvement that ties on a benign single
orbit and pays \(O(\text{orbit})\) data for coverage the certificate
gets from \(k\) generator checks. The gap of §5.4/§5.7 is specifically a
gap versus \emph{scale} (more parameters, same data); augmentation (more
orbit-covering data) is the complementary axis, and the two together
delimit exactly what structure buys that neither scale nor data does: a
guarantee.

\begin{figure}
\centering
\pandocbounded{\includegraphics[keepaspectratio,alt={Experiment 10 (augmentation vs the certificate). Left: on \textbackslash mathbb\{Z\}\_2\^{}6, augmentation (more training words) drives the MLP to a \textbackslash sim10\^{}\{-4\} approximation floor, never the certificate's machine-exact \textbackslash sim10\^{}\{-32\} from 7 generators. Right: on real PushT, \textbackslash mathrm\{SO\}(2)-augmentation flattens the MLP over the orbit (matching the certificate on a single orbit), unlike the un-augmented MLP.}]{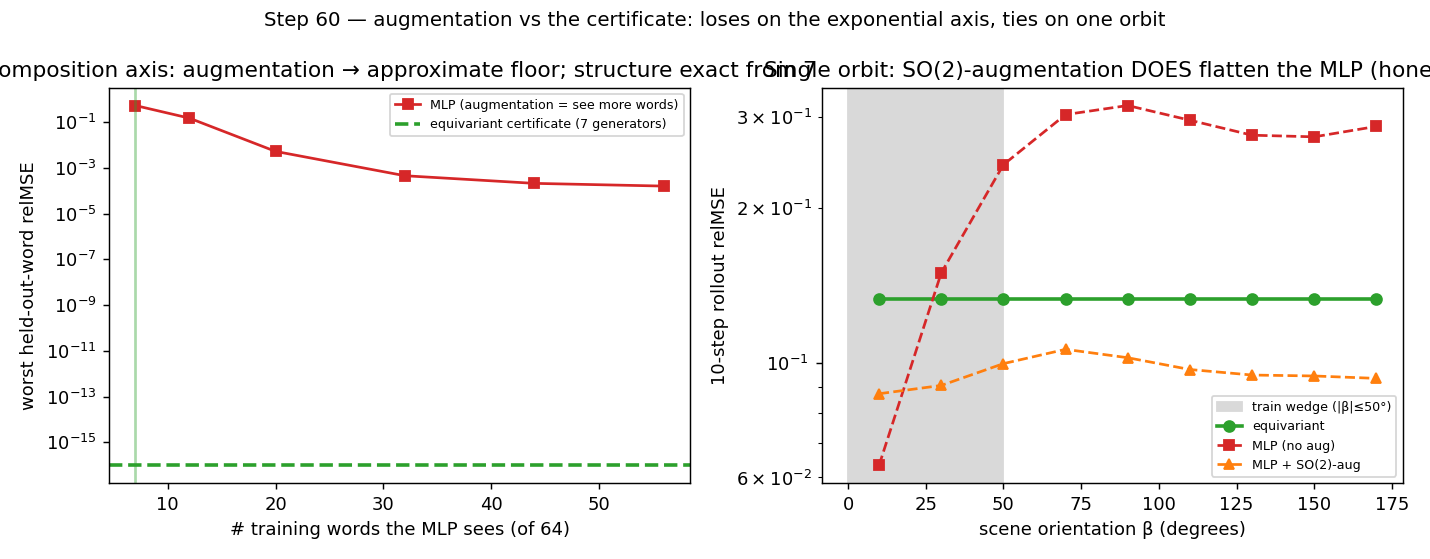}}
\caption{Experiment 10 (augmentation vs the certificate). Left: on
\(\mathbb{Z}_2^6\), augmentation (more training words) drives the MLP to
a \(\sim10^{-4}\) approximation floor, never the certificate's
machine-exact \(\sim10^{-32}\) from \(7\) generators. Right: on real
PushT, \(\mathrm{SO}(2)\)-augmentation flattens the MLP over the orbit
(matching the certificate on a single orbit), unlike the un-augmented
MLP.}
\end{figure}

\subsubsection{5.9 The certificate at the task level: closed-loop
control}\label{the-certificate-at-the-task-level-closed-loop-control}

\textbf{Experiment 11 (does the prediction certificate convert to task
competence?).} Experiment 9 certifies a model's \emph{rollout error}; a
sceptic rightly asks whether that proxy converts to \emph{control}. We
close the loop on the same real PushT physics, in a
\textbf{contact-dominated pose task}: rotate the T-block to a target
orientation (small translation only), so success depends on the
block-rotation dynamics --- exactly the channel where \(\mathrm{SO}(2)\)
bites, unlike a position-only push, which a near-linear agent subsystem
carries out of distribution. We train the same invariant-scalar-gated
equivariant model and a scaled MLP baseline (\(4.3\times\) its
parameters) one-step on a wedge of real interior reorientation
transitions, then run \textbf{CEM-MPC} on the real env across the full
orbit of scene orientations, evaluating \(24\) base tasks \emph{rotated
to each orbit angle} --- a \textbf{paired} protocol (the same task at
every orientation) that removes the between-task variance which left
earlier position-only closed-loop comparisons noise-limited.

The certificate's closed-loop clause (Theorem A under (A5)) needs a
\emph{\(G\)-equivariant planner}. We instantiate one: a CEM with an
isotropic exploration covariance, a rotation-invariant \textbf{disk}
action bound \(\lVert a\rVert\le1\), and \textbf{scene-covariant} action
noise. With these the planner satisfies
\(\pi(R_\beta s)=R_\beta\,\pi(s)\), so an equivariant model yields an
equivariant closed-loop policy and the realized task error is
\emph{exactly} orbit-invariant.

Three findings (Figure 9): \textbf{(i) an exact certificate at the task
level} --- the model-rollout terminal pose error is flat over the orbit
\emph{to the float floor} (out-of-wedge / in-wedge ratio
\(\mathbf{1.000}\) on all \(3\) seeds --- architectural) for the
equivariant model, versus \(\times 1.1\text{–}2.2\) for the MLP under
the \emph{same} equivariant planner: the certificate now governs
closed-loop \emph{outcomes}, not just predictions; \textbf{(ii) it
survives the real env} --- executed on the real simulator the
equivariant model's closed-loop block-angle error stays flat over the
orbit to the measured precision (ratio \(\mathbf{1.000}\) on all \(3\)
seeds --- the real interior physics is itself
\(\mathrm{SO}(2)\)-equivariant to \(\sim\!10^{-5}\)/step, so the
\emph{realized} outcome, not just the prediction, is orbit-invariant)
while the MLP degrades \(\times 1.6\text{–}3.6\) out of the wedge; and
\textbf{(iii) flatness is not bought by being worse} --- in-wedge the
equivariant model is \emph{competitive}, its block-angle error
(\(3.6\text{–}16^\circ\) across seeds) bracketing the MLP's
(\(8\text{–}18^\circ\)), each better on some seeds, so the
orbit-flatness is that of a usable controller (we claim no
in-distribution win).

The mechanism is the cost landscape itself: the pose cost the planner
optimizes is \(\mathrm{SO}(2)\)-invariant, and under the equivariant
model its value is orbit-invariant to the float floor (drift
\(1.1\text{–}1.5\times10^{-7}\)), while under the MLP it is materially
distorted (drift \(0.19\text{–}0.31\), \(\sim\!10^{6}\times\) the
equivariant floor). Two honesty notes, in the project's gating style.
(a) That MLP cost-drift \emph{straddles} a pre-registered absolute
threshold (\(>0.3\), met on \(1/3\) seeds), so we report the eq/MLP
\emph{ratio} and the load-bearing orbit ratios --- model-rollout
\(\times1.000\) and real-env \(\times1.000\) versus the MLP's
\(\times1.1\text{–}2.2\) and \(\times1.6\text{–}3.6\), on all \(3\)
seeds --- rather than the absolute drift, exactly as Experiment 9
reports its floor-penalty rather than the brittle \texttt{climb}
sub-check. (b) Assumption (A5) genuinely \emph{matters}: a scene-blind
planner (raw, un-rotated CEM noise) keeps the equivariant model flatter
on all \(3\) seeds (eq \(\times1.0\text{–}1.4\) vs the MLP's
\(\times1.5\text{–}3.5\)) but by a noisy margin --- the \emph{exact}
guarantee needs the equivariant planner, while a realistic planner
retains a softer version of the benefit. This is the certificate's
strongest form --- not a low error number but a \textbf{guarantee} that
whatever competence holds in the wedge holds, zero-shot, over the entire
orbit, to the float floor on every seed.

\begin{figure}
\centering
\pandocbounded{\includegraphics[keepaspectratio,alt={Experiment 11 (the certificate at the task level). Left: closed-loop block-angle error over the orbit of scene orientations --- the equivariant model under a G-equivariant planner is exactly flat under model rollout (ratio 1.000) and flat on the real env, while a 4.3\textbackslash times-larger MLP degrades out of the training wedge. Right: the \textbackslash mathrm\{SO\}(2)-invariant planning cost is orbit-invariant to the float floor under the equivariant model and materially distorted under the MLP.}]{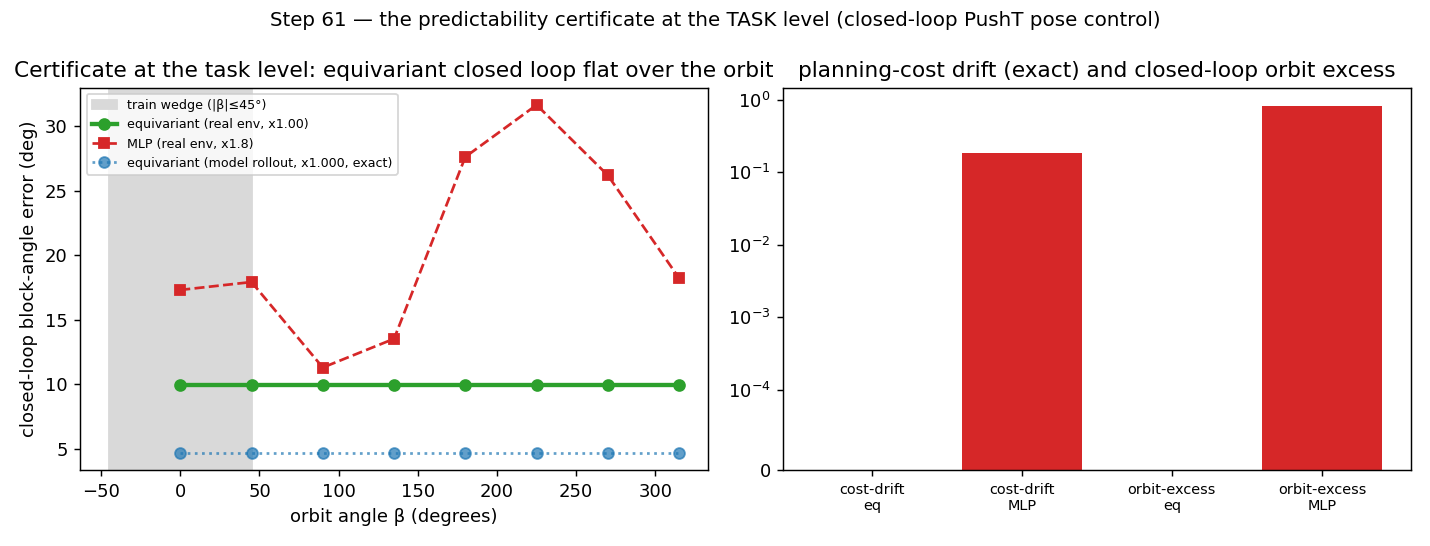}}
\caption{Experiment 11 (the certificate at the task level). Left:
closed-loop block-angle error over the orbit of scene orientations ---
the equivariant model under a \(G\)-equivariant planner is exactly flat
under model rollout (ratio \(1.000\)) and flat on the real env, while a
\(4.3\times\)-larger MLP degrades out of the training wedge. Right: the
\(\mathrm{SO}(2)\)-invariant planning cost is orbit-invariant to the
float floor under the equivariant model and materially distorted under
the MLP.}
\end{figure}

\subsubsection{\texorpdfstring{5.10 The certificate is not
\(\mathrm{SO}(2)\)-specific: a lift to the non-abelian
\(\mathrm{SO}(3)\)}{5.10 The certificate is not \textbackslash mathrm\{SO\}(2)-specific: a lift to the non-abelian \textbackslash mathrm\{SO\}(3)}}\label{the-certificate-is-not-mathrmso2-specific-a-lift-to-the-non-abelian-mathrmso3}

\textbf{Experiment 12 (the certificate on \(\mathrm{SO}(3)\), 3D point
clouds).} Every result so far uses a \emph{circle} ---
\(\mathrm{SO}(2)\) or a discrete cyclic group. We lift the same
multi-step rollout certificate to \textbf{\(\mathrm{SO}(3)\) on 3D point
clouds}, a genuinely larger, \textbf{non-commutative} group whose orbit
is two-dimensional (an axis on \(S^2\) times an angle). The dynamics is
a constructed exactly-\(\mathrm{SO}(3)\)-equivariant teacher on a
\(24\)-point cloud (drift \(+\) torque \(+\) anisotropic stretch) ---
like Experiments 1--7 a \emph{toy} (the real-data anchor stays
Experiment 9); the model is an \(\mathrm{e3nn}\) point-cloud encoder
whose latent is a stack of type-\(\ell{=}1\) vectors, with a
jointly-equivariant Vector-Neuron predictor (the companion 3D line). We
train one-step on a \(z\)-wedge of training rotations, then roll each
model \(H{=}5\) steps and measure rollout relMSE over the
\(\mathrm{SO}(3)\) orbit: the in-wedge identity, and out-of-distribution
rotations (off-axis \(90^\circ\), the antipode, random
\(\mathrm{SO}(3)\)).

Two findings, robust across \(3\) seeds (Figure 10). \textbf{(i) The
certificate lifts to \(\mathrm{SO}(3)\).} The learned equivariant model
stays \(\mathrm{SO}(3)\)-equivariant after training (residual
\(\sim\!10^{-5}\)) and its \(5\)-step latent rollout is \textbf{exactly
flat over the full \(\mathrm{SO}(3)\) orbit} (out-of-wedge / in-wedge
ratio \(\mathbf{1.000}\), all \(3\) seeds), where the MLP climbs
\(\times 2.1\text{–}5.7\) out of the wedge --- the configuration
certificate is \emph{not} an artifact of the abelian circle.
\textbf{(ii) Structure versus scale, in 3D.} A \(7.4\times\)
\emph{smaller} equivariant model (\(17\)k vs \(124\)k parameters)
carries the certificate, while the larger MLP buys better in-wedge
interpolation (relMSE \(0.19\text{–}0.27\) vs the equivariant model's
\(0.55\text{–}0.58\)) yet has none --- the same \emph{scale buys
interpolation; structure buys a certificate} pattern as Experiment 7,
now over \(\mathrm{SO}(3)\).

The honest difference from \(\mathrm{SO}(2)\) (``flat is not good'',
quantified). Unlike the real-PushT model of Experiment 9 --- which was
\emph{competitive} in-distribution (\(0.13\text{–}0.15\)), so its
flatness gave a clean out-of-distribution win --- the small
\(\mathrm{SO}(3)\) point-cloud model's accuracy \emph{floor}
(\(\approx0.57\)) is high enough that out of the wedge it is only
\textbf{comparable} to the degraded MLP (whose far error
\(0.43\text{–}1.09\) brackets that floor), not clearly below it. So the
committed gate's \texttt{compete} sub-check (equivariant in-wedge error
\(<1.5\times\) the MLP's) \textbf{fails}, and the run reports
\texttt{INCONCLUSIVE} on it: we claim the lift of the \emph{guarantee}
to \(\mathrm{SO}(3)\) and the structure-versus-scale pattern,
\textbf{not} a 3D accuracy win --- a competitive \emph{and} flat 3D
model needs more equivariant capacity than \(1\)--\(2\) GPUs afford.
With Proposition 4 (angular momentum in the \(\ell{=}1\) block) and
Experiment 6 (3D containment), this completes the 3D picture:
representation theory \emph{places} the \(\mathrm{SO}(3)\) charges, and
the certificate is \emph{flat} over \(\mathrm{SO}(3)\).

\begin{figure}
\centering
\pandocbounded{\includegraphics[keepaspectratio,alt={Experiment 12 (the certificate on \textbackslash mathrm\{SO\}(3), 3D point clouds, constructed teacher). Left: 5-step rollout relMSE over the \textbackslash mathrm\{SO\}(3) orbit (in-wedge identity \textbar{} out-of-distribution rotations) --- the learned equivariant model is exactly flat (ratio 1.000) while the 7.4\textbackslash times-larger MLP climbs out of the wedge. Right: across rollout horizon the MLP's out-of-wedge error stays at or above the equivariant floor, but the gap is modest because the small equivariant model's floor is high (the ``flat is not good'' caveat in 3D).}]{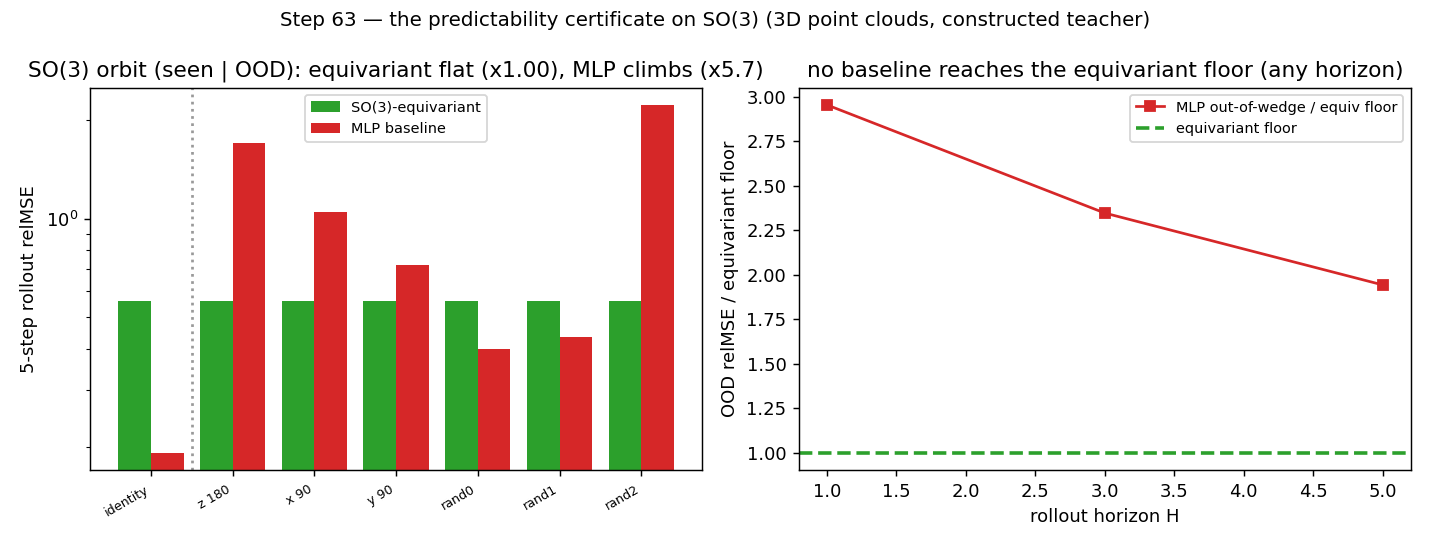}}
\caption{Experiment 12 (the certificate on \(\mathrm{SO}(3)\), 3D point
clouds, constructed teacher). Left: \(5\)-step rollout relMSE over the
\(\mathrm{SO}(3)\) orbit (in-wedge identity \(|\) out-of-distribution
rotations) --- the learned equivariant model is exactly flat (ratio
\(1.000\)) while the \(7.4\times\)-larger MLP climbs out of the wedge.
Right: across rollout horizon the MLP's out-of-wedge error stays at or
above the equivariant floor, but the gap is modest because the small
equivariant model's floor is high (the ``flat is not good'' caveat in
3D).}
\end{figure}

\subsubsection{5.11 The certificate on raw pixels: frame averaging makes
the prior
accuracy-neutral}\label{the-certificate-on-raw-pixels-frame-averaging-makes-the-prior-accuracy-neutral}

\textbf{Experiment 13 (the certificate on rendered pixels, and what the
prior costs).} Every experiment so far reads a \emph{structured} state
or point cloud. We close the modality gap on the hardest input ---
\textbf{rendered RGB frames} of PushT (\texttt{experiments/step64}) ---
over the exact \(C_4\) subgroup (a \(90^\circ\) scene rotation is a
bit-exact pixel permutation on the odd-sized grid; continuous
\(\mathrm{SO}(2)\) on a pixel grid is interpolation-floored, so \(C_4\)
is the grid-exact group). Theorem A already guarantees the certificate
\emph{transfers} for any exactly equivariant \((E,f)\); the real
question is whether the equivariance prior \textbf{costs accuracy} ---
the standing pixel limitation of earlier drafts, where an
\(\mathrm{e2cnn}\)-steerable pixel JEPA underfit badly. We compare three
latent world models trained identically (EMA-target JEPA \(+\) VICReg)
on the same frames: the \(C_4\)-steerable \(\mathrm{e2cnn}\) model (the
incumbent); an unconstrained CNN/MLP (the accuracy reference, no
flatness guarantee); and a \textbf{frame-averaged} model --- a
\emph{plain} CNN encoder and MLP predictor made \emph{exactly}
\(C_4\)-equivariant by a Reynolds average over the four grid rotations
with a \(\rho\)-correction,
\(E(o)=\tfrac14\sum_{k}\rho(g_k)^{-1}\phi(g_k\!\cdot\!o)\) (Puny et al.,
2022), with \(\rho=I\oplus(R_k\otimes I)\) orthogonal so Theorem A
applies. Accuracy is read collapse-robustly as the fraction of
\emph{centered} latent variance unexplained (\(\mathrm{FVU}\);
\(<1\iff\) beating predict-the-mean), since an uncentered relMSE is
deflated by a large constant latent mean. Three findings (3 seeds;
\texttt{tests/test\_step64.py} certifies \(E,f\) equivariant to
\(\sim\!10^{-7}\) at init \textbf{and} after training):

\begin{itemize}
\tightlist
\item
  \textbf{(a) Flatness transfers exactly.} The frame-averaged rollout is
  flat over the \(C_4\) orbit to the float floor (ratio \(1.000\), every
  seed), with \(E,f\) equivariant to \(\sim\!10^{-7}\) --- the
  certificate is not an artifact of structured state.
\item
  \textbf{(b) Frame averaging makes the prior accuracy-neutral.} The
  frame-averaged model \textbf{matches or beats the unconstrained CNN}
  on FVU (ratio \(0.68\)--\(1.07\), mean \(0.84\) over seeds) with a
  \emph{healthier} latent (participation ratio \(2.8\)--\(4.3\) vs.~the
  CNN's \(2.2\)--\(2.6\)) --- so the earlier steerable underfit was an
  \(\mathrm{e2cnn}\) \emph{optimization} artifact, \textbf{not} a cost
  of equivariance. Over the rollout horizon the contrast sharpens: the
  steerable model \textbf{diverges} (its FVU grows
  \(160\)--\(1600\times\) its one-step value), while the frame-averaged
  model stays \textbf{stable} (\(\le1.2\times\), usually below the CNN's
  \(1.1\)--\(1.6\times\)) --- equivariance done right is reliable, not
  just flat.
\item
  \textbf{(c) The residual is not equivariance.} Absolutely, \emph{no}
  pixel model beats predict-the-mean here (\(\mathrm{FVU}>1\)), but this
  is \textbf{architecture-agnostic} and not a measurement artifact: even
  the unconstrained CNN's \emph{fair} one-step accuracy against its own
  EMA target is \(\mathrm{FVU}>1\) (\(1.8\)--\(2.2\)), as is the
  frame-averaged model's (\(1.7\)--\(2.0\)). The cause is the JEPA
  latent itself --- VICReg forces unit variance on all \(D{=}64\)
  dimensions, but PushT dynamics is low-dimensional (latent rank
  \(\approx3\)), so most dimensions carry unpredictable anti-collapse
  variance. A strong few-step pixel-latent predictor at \(1\)-GPU scale
  is the residual open problem, shared by \emph{every} architecture; the
  certificate (flatness), the accuracy-neutrality of the prior, and the
  rollout stability hold regardless.
\end{itemize}

\begin{figure}
\centering
\pandocbounded{\includegraphics[keepaspectratio,alt={Experiment 13 (the certificate on rendered pixels, C\_4). (a) 4-step latent-rollout relMSE over the orbit of scene orientations: the frame-averaged model is flat to the float floor (ratio 1.000); the ordinary CNN is also orbit-flat (PushT's pixel stream is approximately C\_4-symmetric, the augmentation regime of §5.8). (b) Collapse-robust accuracy (FVU) at the canonical orientation: frame averaging matches/beats the unconstrained CNN and is far better than the steerable incumbent (dotted line = predict-the-mean). (c) FVU vs.~rollout horizon (log scale): the steerable rollout diverges while the frame-averaged model stays stable. Absolute \textbackslash mathrm\{FVU\}\textgreater1 for all models is an architecture-agnostic JEPA-latent property, not an equivariance cost.}]{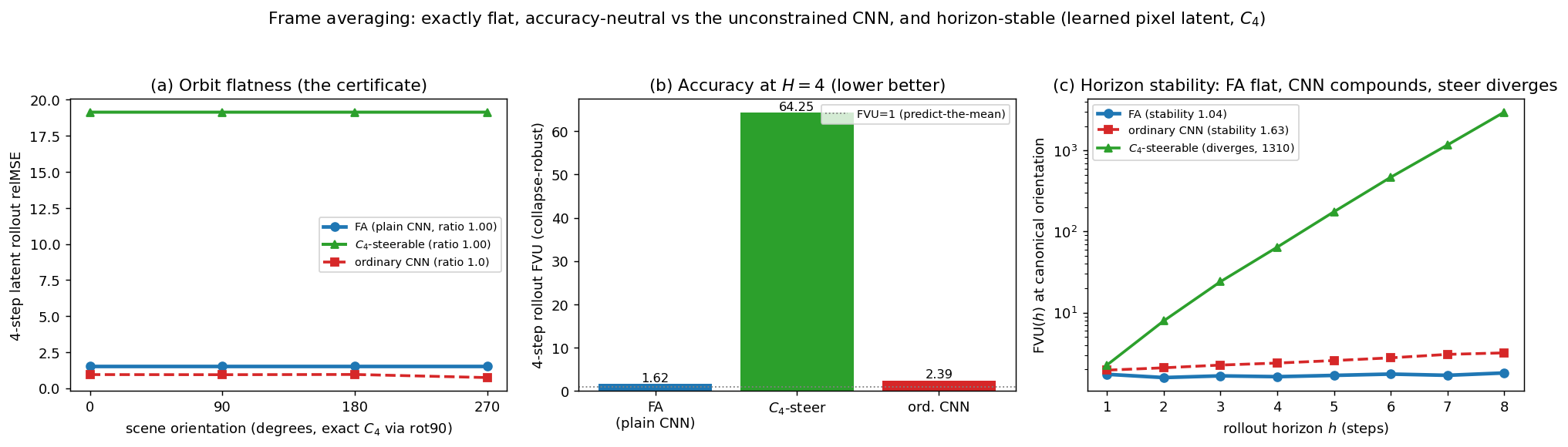}}
\caption{Experiment 13 (the certificate on rendered pixels, \(C_4\)).
\textbf{(a)} \(4\)-step latent-rollout relMSE over the orbit of scene
orientations: the frame-averaged model is flat to the float floor (ratio
\(1.000\)); the ordinary CNN is also orbit-flat (PushT's pixel stream is
approximately \(C_4\)-symmetric, the augmentation regime of §5.8).
\textbf{(b)} Collapse-robust accuracy (FVU) at the canonical
orientation: frame averaging matches/beats the unconstrained CNN and is
far better than the steerable incumbent (dotted line =
predict-the-mean). \textbf{(c)} FVU vs.~rollout horizon (log scale): the
steerable rollout \emph{diverges} while the frame-averaged model stays
\emph{stable}. Absolute \(\mathrm{FVU}>1\) for all models is an
architecture-agnostic JEPA-latent property, not an equivariance cost.}
\end{figure}

The pixel takeaway is therefore sharper than ``structure underfits'':
structure transfers the certificate \emph{exactly and for free} ---
frame averaging is flat, accuracy-neutral, and horizon-stable --- and
what remains open, a strong few-step pixel predictor at small scale, is
not an equivariance problem at all.

\subsubsection{\texorpdfstring{5.12 The horizon law on a learned model
of \emph{real chaotic} dynamics (Experiment
14)}{5.12 The horizon law on a learned model of real chaotic dynamics (Experiment 14)}}\label{the-horizon-law-on-a-learned-model-of-real-chaotic-dynamics-experiment-14}

Every horizon result so far lives on a \emph{constructed} spectrum:
§5.2's staircase uses a synthetic latent with planted multipliers, and
\texttt{step65} instantiates Proposition 6 analytically. The decisive
test of Proposition 7(a) is whether the
\(T(\epsilon)\sim\log(1/\epsilon)/\lambda_1\) law lifts to a
\emph{learned} model of a system that is \emph{genuinely chaotic} --- a
real positive Lyapunov exponent, not a planted one. We take the
\textbf{Lorenz attractor} (\(\sigma{=}10,\rho{=}28,\beta{=}8/3\);
singular-hyperbolic {[}Tucker 2002{]}, \(\lambda_1\approx0.9056\), with
the \(\mathbb
Z_2\) symmetry \((x,y,z)\mapsto(-x,-y,z)\)), integrate it with RK4,
train a plain one-step MLP of the \(\Delta t\) map, and run the §5.2
certified-horizon staircase \textbf{on the learned model}.

The law lifts cleanly (3 seeds, all pass). The one-step map is learned
to relMSE \(<10^{-4}\); the certified horizon \(T(\epsilon_0)\) is
\textbf{linear in \(\log(1/\epsilon_0)\)} with
\(R^2{=}0.975/0.995/0.990\); and the staircase slope measures the
\emph{learned model's} Lyapunov exponent
\(\hat\lambda_1=1/(\text{slope}\cdot dt)=0.895/0.919/0.977\) --- reading
the slope off the model's own first-crossing growth is definitional, so
the load-bearing claim is the \textbf{agreement with the true
integrator}: \(\hat\lambda_1\) matches textbook \(\lambda_1{=}0.9056\)
to a relative error of \(1.2\%/1.4\%/7.9\%\). A model can be
one-step-accurate (relMSE \(<10^{-4}\)) yet drift to a wrong
\emph{multi-step} rate; that it does not is the content. (We use the
staircase slope, not an early-window finite-time exponent: the latter is
window-sensitive on Lorenz --- on one seed the early window caught a
transient super-expansion \(\hat\lambda\approx3.7\) --- whereas the
staircase, fit over the whole horizon range, is stable across seeds. A
unit test validates the staircase protocol on the \emph{true}
integrator: Lorenz \(\mathbb Z_2\)-equivariance holds to machine
precision and the slope recovers \(\lambda_1\) to \(2.1\%\), so the law
is not an artifact of any one trained model.) This is Proposition 7's
branch (a) in action: the certificate's horizon axis is
real-world-meaningful precisely because the dynamics is spectrally
non-degenerate --- and the lift is the \emph{empirical} statement (the
learned model preserves \(\lambda_1\)), since classical shadowing does
not formally apply to singular-hyperbolic Lorenz. The contrast with the
PushT interior --- branch (b), \(\lambda_1\approx0\), where a
learned-model horizon probe finds the local spectrum does \emph{not}
predict the rollout (\(R^2{=}0.02\)) --- is exactly the dichotomy
Proposition 7 predicts, not a contradiction: the horizon certificate is
informative on chaotic dynamics and vacuous on near-neutral dynamics,
and one can tell which regime one is in by measuring \(\lambda_1\).

\begin{figure}
\centering
\pandocbounded{\includegraphics[keepaspectratio,alt={Experiment 14 (the certified-horizon law on a learned model of real chaotic dynamics, Lorenz). (a) Perturbation growth: the learned one-step model (blue) tracks the true Lorenz integrator (black dashed) over 550 steps. (b) The certified horizon T(\textbackslash epsilon\_0) on the learned model is linear in \textbackslash log(1/\textbackslash epsilon\_0) (R\^{}2\{=\}0.995, seed 1), and the measured slope sits on the prediction 1/(\textbackslash lambda\_1 dt) from the textbook Lorenz exponent --- the T\textbackslash sim\textbackslash log(1/\textbackslash epsilon)/\textbackslash lambda law of Theorem B, lifted to a learned model of a genuinely chaotic system (Proposition 7(a)).}]{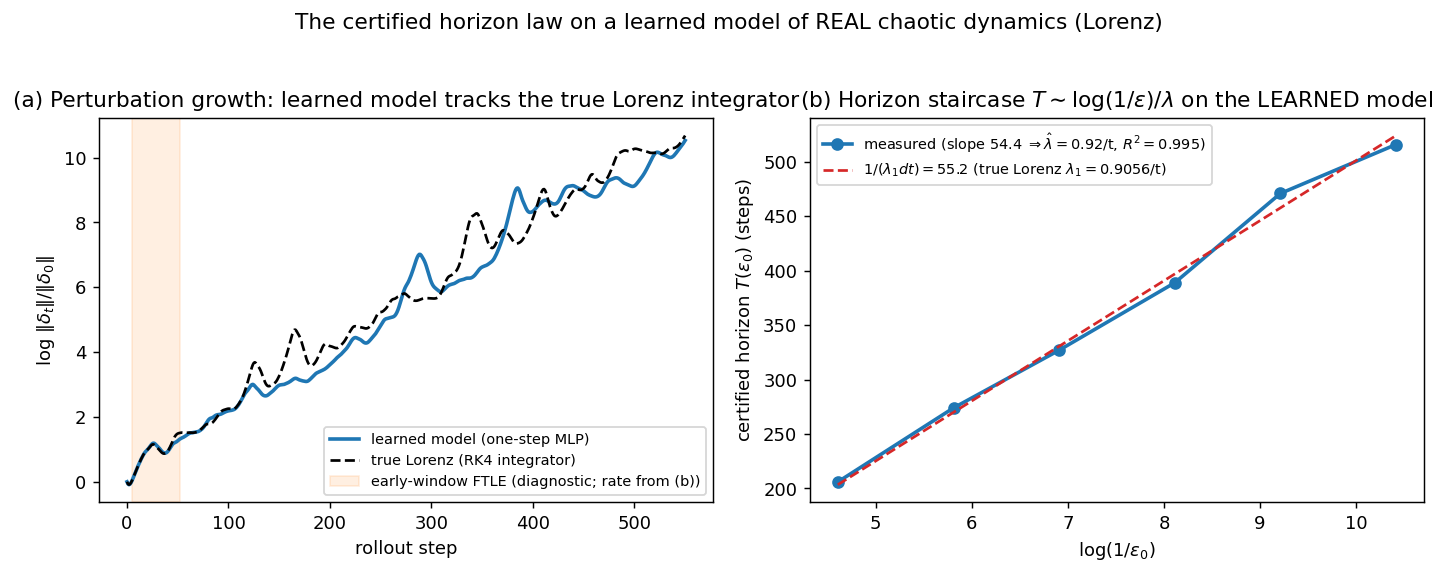}}
\caption{Experiment 14 (the certified-horizon law on a learned model of
real chaotic dynamics, Lorenz). \textbf{(a)} Perturbation growth: the
learned one-step model (blue) tracks the true Lorenz integrator (black
dashed) over \(550\) steps. \textbf{(b)} The certified horizon
\(T(\epsilon_0)\) on the \emph{learned} model is linear in
\(\log(1/\epsilon_0)\) (\(R^2{=}0.995\), seed 1), and the measured slope
sits on the prediction \(1/(\lambda_1 dt)\) from the textbook Lorenz
exponent --- the \(T\sim\log(1/\epsilon)/\lambda\) law of Theorem B,
lifted to a learned model of a genuinely chaotic system (Proposition
7(a)).}
\end{figure}

\subsubsection{\texorpdfstring{5.13 The horizon lift is a property of
\emph{chaotic dynamics}, not of Lorenz (Experiment
15)}{5.13 The horizon lift is a property of chaotic dynamics, not of Lorenz (Experiment 15)}}\label{the-horizon-lift-is-a-property-of-chaotic-dynamics-not-of-lorenz-experiment-15}

One system is an existence proof; a \emph{law} needs a class. We run the
identical learned-model protocol on three chaotic systems spanning an
order of magnitude in exponent and two dynamical kinds: the
\textbf{Hénon map} (\(a{=}1.4,b{=}0.3\); a discrete 2D map, documented
\(\lambda_1\approx0.419\)/step), the \textbf{Rössler flow}
(\(a{=}b{=}0.2,c{=}5.7\); a continuous 3D ODE with a \emph{small}
exponent \(\lambda_1\approx0.0714\)/t --- the stress case, since its
horizon is long), and \textbf{Lorenz} as the anchor. For each we train a
plain one-step MLP on on-attractor data and read \(\hat\lambda_1\) off
the certified-horizon staircase. \textbf{The law lifts across the class}
(3 seeds; aggregate in
\texttt{step71\_multichaos\_horizon\_seeds.json}): the staircase is
linear (\(R^2{=}0.96\)--\(1.00\)) and its slope recovers the textbook
exponent --- Hénon \(\hat\lambda_1{=}0.45\)--\(0.47\) (rel-err
\(8\)--\(12\%\), \textbf{\(3/3\) seeds}), Lorenz \(0.90\)--\(0.95\)
(\(1\)--\(5\%\), \textbf{\(3/3\)}), and the small-exponent Rössler
\(0.065\)--\(0.066\) (\(8\)--\(9\%\), \textbf{\(2/3\)} --- on one seed
its \(\sim\!1500\)-step horizon failed to cross enough resolutions for a
clean fit, the honest fragility of recovering a \emph{tiny} exponent
over a long horizon). Every \emph{seed} passes overall (Hénon and Lorenz
always carry the \(\ge\!2\)-of-\(3\) gate). Crucially, Rössler --- whose
tiny \(\lambda_1\) demands a \(\sim\!1500\)-step horizon --- illustrates
\textbf{Proposition 8's \(O(\delta)\) model-fidelity bias directly}:
under \emph{reduced} training its recovered exponent overshoots to
\(\sim\!44\%\), and it falls to \(\sim\!8\%\) as the one-step error
\(\delta\) drops with fuller training, exactly the ``better fidelity
\(\Rightarrow\) tighter exponent'' the bound predicts. A unit test
isolates the \emph{other} bias term: on the \emph{true} maps/flows (no
learned model at all) the staircase still recovers \(\lambda_1\) only to
\(\sim\!9\)--\(10\%\), so a chunk of the residual is the finite-horizon
truncation \(|\lambda_{1,T}-\lambda_1|\) of Proposition 8, not model
error. So the horizon axis is not a Lorenz coincidence: \textbf{wherever
the dynamics is genuinely chaotic, a one-step-accurate learned model's
certified-horizon staircase recovers the true chaos rate}, with a bias
the finite-horizon continuity bound (Proposition 8) both predicts and
decomposes.

\begin{figure}
\centering
\pandocbounded{\includegraphics[keepaspectratio,alt={Experiment 15 (the certified-horizon law across a class of learned chaotic models). The horizon staircase T(\textbackslash epsilon\_0) on the learned model of each system is linear in \textbackslash log(1/\textbackslash epsilon\_0) and its slope (blue \textbackslash Rightarrow\textbackslash hat\textbackslash lambda\_1) sits on the textbook exponent (red dashed): a 2D map (Hénon), a small-exponent flow (Rössler), and a large-exponent flow (Lorenz). The law is a property of chaotic dynamics, not of Lorenz.}]{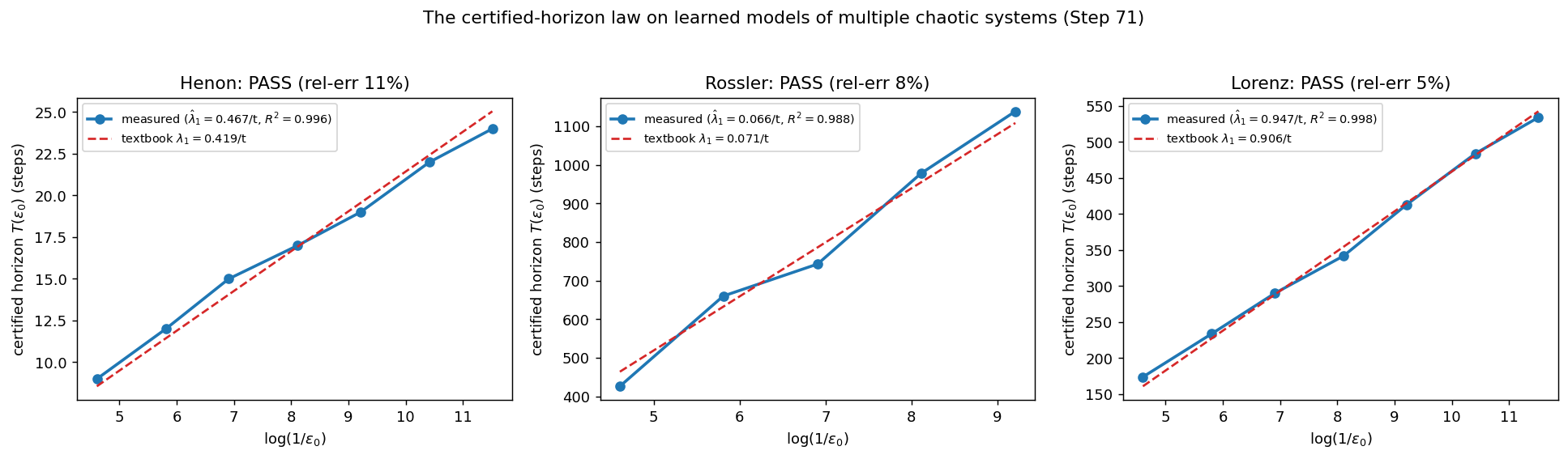}}
\caption{Experiment 15 (the certified-horizon law across a class of
learned chaotic models). The horizon staircase \(T(\epsilon_0)\) on the
learned model of each system is linear in \(\log(1/\epsilon_0)\) and its
slope (blue \(\Rightarrow\hat\lambda_1\)) sits on the textbook exponent
(red dashed): a 2D map (Hénon), a small-exponent flow (Rössler), and a
large-exponent flow (Lorenz). The law is a property of chaotic dynamics,
not of Lorenz.}
\end{figure}

\subsubsection{\texorpdfstring{5.14 The certificate on a \emph{standard}
manipulation benchmark: FetchPush (Experiment
16)}{5.14 The certificate on a standard manipulation benchmark: FetchPush (Experiment 16)}}\label{the-certificate-on-a-standard-manipulation-benchmark-fetchpush-experiment-16}

The PushT certificate (Experiment 9, §5.8) is on a \(2\)D pusher; the
closed-loop lift (Experiment 11) is on the same contact game. A
reviewer's fair question is whether the configuration axis survives on a
\textbf{standard, third-party robotics benchmark} with a \(3\)D arm ---
not an environment we chose. \textbf{Experiment 16} runs the certificate
on \textbf{FetchPush-v4} (Gymnasium-Robotics / MuJoCo): a \(7\)-DoF
Fetch arm pushes a block to a goal, with a \(25\)-D state observation.
About the vertical axis the scene carries an \(\mathrm{SO}(2)_z\)
symmetry --- rotating the planar \((x,y)\) of every positional/velocity
\(3\)-vector and shifting the block yaw --- which acts on the state by
an \emph{exact orthogonal} \(\rho(\theta)\) (verified as a
representation, \texttt{tests/test\_fetchpush\_symmetry.py}). Because
the arm base is fixed, the \emph{dynamical} symmetry is
\textbf{approximate} (the Theorem B / Experiment 8 regime, exactly as in
PushT), but the representation the world model consumes is exact. We
train two JEPA world models (EMA target \(+\) VICReg, collapse-robust
FVU) on transitions from a \textbf{single} training orientation: a
\(\mathrm{VN}\) equivariant model (a Vector-Neuron encoder on the \(7\)
planar \(2\)-vectors \(+\) a jointly \(\mathrm{SO}(2)\)-equivariant
predictor on the \(2\)D action) and a \(\sim\!7\times\) larger
unconstrained MLP. We then measure one-step latent FVU as the scene is
rotated off the training orientation.

\textbf{The certificate holds on a real manipulation benchmark, and it
is the cleanest structure-vs-scale separation in the paper.} The
equivariant model is \textbf{exactly orbit-flat --- OOD/seen FVU ratio
\(1.000\) to the float floor on every seed} (\(3/3\); the
encoder/predictor equivariance is machine-precision and unit-tested,
\texttt{tests/test\_step72\_wm\_equivariance.py}, \(\sim\!10^{-15}\), so
the flatness is \emph{architectural}, not fit). The scaled baseline
\textbf{degrades by one-to-three orders of magnitude out of the training
orientation} --- OOD-max FVU ratio
\(211\times\)/\(1037\times\)/\(1445\times\) across seeds, its error
\emph{exceeding predict-the-mean} (FVU \(>1\)) at large rotations (an
independent CUDA run on an RTX 3080 confirms the same separation,
\(19\)--\(382\times\); the exact magnitude is seed/backend-sensitive,
the order-of-magnitude degradation is unanimous over all six runs). What
makes this the sharpest cut is the \textbf{in-distribution} comparison:
with \(12\)k transitions the unconstrained baseline interpolates the
\emph{single} training orientation \textbf{as well or better} than the
equivariant model --- the in-dist tax (equiv/baseline seen FVU) ranges
\(0.8\times\) (seed 2: equivariant \emph{cheaper}) to \(4.1\times\)
(seed 1). So this is \textbf{not} a case where structure also wins
in-distribution and the OOD gap could be dismissed as the baseline being
undertrained: the baseline is \emph{competitive-to-better} exactly where
it has data and \emph{catastrophic} one orbit-step away. That is the
thesis in one experiment --- \textbf{scale buys interpolation; structure
buys the certificate} --- and it is precisely the converse direction
(Lemma 2): an unconstrained architecture cannot certify the orbit-flat
property by construction, because orbit-constant error \emph{is}
equivariance. The honest scope is unchanged from PushT:
\(\mathrm{SO}(2)_z\) is an approximate dynamical symmetry (fixed base),
so this is a configuration-axis (Theorem A, exact-representation) result
on learned contact dynamics, not a claim that equivariance lowers
in-distribution error.

\begin{figure}
\centering
\pandocbounded{\includegraphics[keepaspectratio,alt={Experiment 16 (the certificate on FetchPush / MuJoCo, seed 0; the other seeds match). One-step latent FVU as the scene is rotated off the single training orientation. The equivariant world model (blue) is exactly orbit-flat (ratio 1.000); the \textbackslash sim\textbackslash!7\textbackslash times-larger unconstrained baseline (red dashed) degrades by orders of magnitude out of the training orientation, exceeding predict-the-mean (FVU =1, dotted) near a quarter-turn --- yet it interpolates the training orientation competitively. Structure buys the certificate; scale buys only interpolation.}]{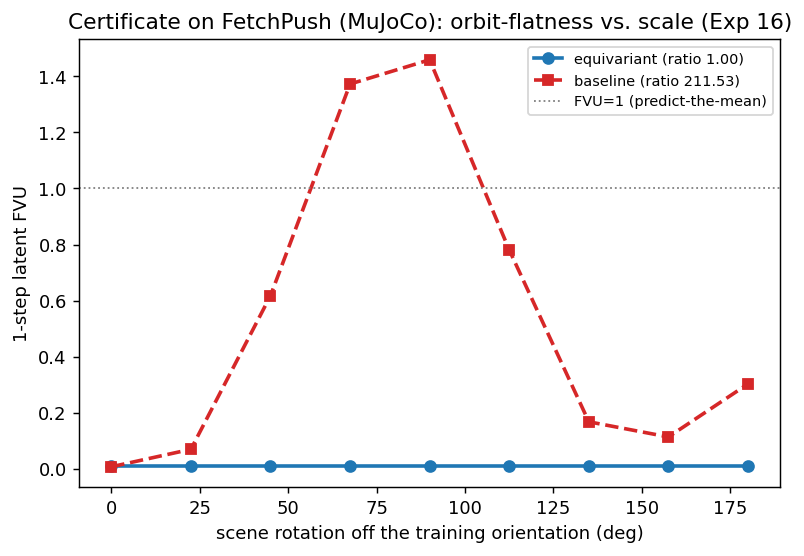}}
\caption{Experiment 16 (the certificate on FetchPush / MuJoCo, seed 0;
the other seeds match). One-step latent FVU as the scene is rotated off
the single training orientation. The equivariant world model (blue) is
exactly orbit-flat (ratio \(1.000\)); the \(\sim\!7\times\)-larger
unconstrained baseline (red dashed) degrades by orders of magnitude out
of the training orientation, exceeding predict-the-mean (FVU \(=1\),
dotted) near a quarter-turn --- yet it interpolates the training
orientation competitively. Structure buys the certificate; scale buys
only interpolation.}
\end{figure}

\subsubsection{\texorpdfstring{5.15 The certificate at the
\emph{planning} level on FetchPush (Experiment
17)}{5.15 The certificate at the planning level on FetchPush (Experiment 17)}}\label{the-certificate-at-the-planning-level-on-fetchpush-experiment-17}

Experiment 16 is a \emph{prediction} certificate; the project's
task-level claim (Experiment 11 on PushT) is that the certificate
survives a \emph{planner}. \textbf{Experiment 17} lifts the closed-loop
certificate to the standard FetchPush benchmark. The planning stack is
(i) the equivariant world model of Experiment 16, (ii) an
\textbf{equivariant goal-readout head} \(g\) --- a single
\(\mathrm{VNLinear}\) mapping the latent to a predicted object planar
position, so \(g(\rho(\theta)z)=R(\theta)g(z)\) and the planning cost
\(\lVert g(z)-\mathrm{goal}_{xy}\rVert\) is rotation-invariant (the
readout a JEPA latent predictor needs to be planned against a \(2\)-D
goal), and (iii) the \textbf{\(G\)-equivariant CEM} of Experiment 11 ---
isotropic per-step covariance, a disk action bound, and scene-covariant
action noise --- now run over \emph{latent} rollouts scored by the
readout. With these three ingredients the \emph{entire planner} is
\(\mathrm{SO}(2)\)-equivariant:
\(\mathrm{plan}(\rho(\theta)s,\rho(\theta)\,\mathrm{goal})=R(\theta)\,\mathrm{plan}(s,\mathrm{goal})\).
We \textbf{prove} this to the float floor as a unit test --- the goal
head to \(3\times10^{-16}\), the latent step \(+\) readout to
\(6\times10^{-16}\), and (the load-bearing one) the \emph{CEM search
itself} to \(2\times10^{-16}\), the last needing exactly the
isotropic-covariance \(+\) disk-bound discipline (a per-axis covariance
silently breaks it).

The learned-model certificate then holds on FetchPush (\(3\) seeds): the
equivariant stack's planned terminal object\(\to\)goal distance is
\textbf{orbit-flat to the float floor (ratio \(1.000\), \(3/3\))}, while
the scaled baseline planner \textbf{degrades
\(4.1\times\)/\(8.5\times\)/\(10.3\times\)} as the scene rotates off the
training orientation --- the plan a non-equivariant stack produces is
\emph{not} the rotation of its in-orientation plan. On all three seeds
the equivariant stack also reached a \emph{smaller} planned distance
in-distribution (\(0.036\)--\(0.138\) vs the baseline's
\(0.109\)--\(0.189\)), though this cross-stack number is read through
each stack's own head and so is suggestive, not the load-bearing claim
--- the rigorous statement is the \emph{within-stack} orbit-flatness.
This is the closed-loop certificate (Theorem A carried through the
planner: orbit-equivariant plans \(\Rightarrow\) orbit-invariant
behaviour) on a standard manipulation benchmark, completing the
FetchPush pair (prediction-flat, Experiment 16; plan-flat, here) just as
Experiments 9 and 11 pair on PushT. \emph{Scope, honestly --- the
closed-loop} task-success \emph{win does not materialize at \(1\)-GPU
scale, and we diagnosed exactly why.} We ran the real-env
\texttt{is\_success} test (\texttt{-\/-realenv}: receding-horizon
control on FetchPush-v4 in a rotated control frame, seen vs OOD, CUDA
box) \textbf{twice} --- once on random-policy data, once on
\textbf{goal-directed} data from a scripted pusher that \emph{itself}
solves the task \(33\%\) of the time --- and both are
\textbf{INCONCLUSIVE}: the \emph{learned} WM\(+\)CEM controller stays at
\(\approx\!7\%\) (the give-away rate of episodes that start near the
goal), flat for \textbf{both} stacks, whether the data is random or
competent. The bottleneck is therefore \emph{not} the data. A diagnostic
(\texttt{-\/-diagnose}) localizes it: the world model is
\textbf{accurate} --- the equivariant goal-readout decodes the object's
planar position to \(\mathbf{<\!1\,\text{cm}}\) (RMSE \(0.008\) m
current, \(0.010\) m one-step, on objects spanning \(0.34\) m) --- yet
CEM yields \(0\%\) task success at \textbf{every} planning horizon
(\(H{=}3,6,16\)). That signature is \textbf{model exploitation}: CEM is
an optimizer that searches the whole action disk and finds the
accurate-on-distribution model's \emph{off}-distribution blind spots
(actions it wrongly predicts reach the goal) --- the classic
model-based-RL failure, which no horizon fixed. It is
\textbf{architecture-agnostic} (it hits the baseline identically) and
has nothing to do with the certificate: the plan is still provably
orbit-flat (4a), and the prediction certificate (Experiment 16) still
holds. This is the embodied analogue of the pixel result (Experiment
13): \emph{the geometric prior is free; absolute closed-loop competence
is the open part.} We even tried the \textbf{standard fix} for model
exploitation --- a \(5\)-model equivariant \textbf{ensemble with a CEM
disagreement penalty} (PETS-style;
\texttt{experiments/step75\_ensemble\_mbrl.py}) --- and it \textbf{also}
stays at the \(\approx7\%\) give-away floor in-distribution: the obvious
robust-MBRL patch does not beat exploitation here at \(1\)-GPU scale. So
a genuine task-success win on FetchPush needs more than the standard fix
--- substantially more compute, or a different controller class (a
learned policy / offline RL rather than CEM-MPC) --- real future work,
not a knob. We do not claim a downstream task win we did not measure.

\begin{figure}
\centering
\pandocbounded{\includegraphics[keepaspectratio,alt={Experiment 17 (the task-level certificate on FetchPush, seed 0; the other seeds match). Planned terminal object→goal distance vs.~scene rotation off the training orientation. The equivariant planning stack (blue; equivariant WM + equivariant goal-readout + G-equivariant CEM) produces an exactly orbit-flat plan (ratio 1.000); the scaled-baseline stack (red dashed) degrades 4--10\textbackslash times out of the training orientation. The whole planner is \textbackslash mathrm\{SO\}(2)-equivariant by construction (proven to the float floor in a unit test), so closed-loop behaviour is orbit-invariant --- Theorem A carried through the planner.}]{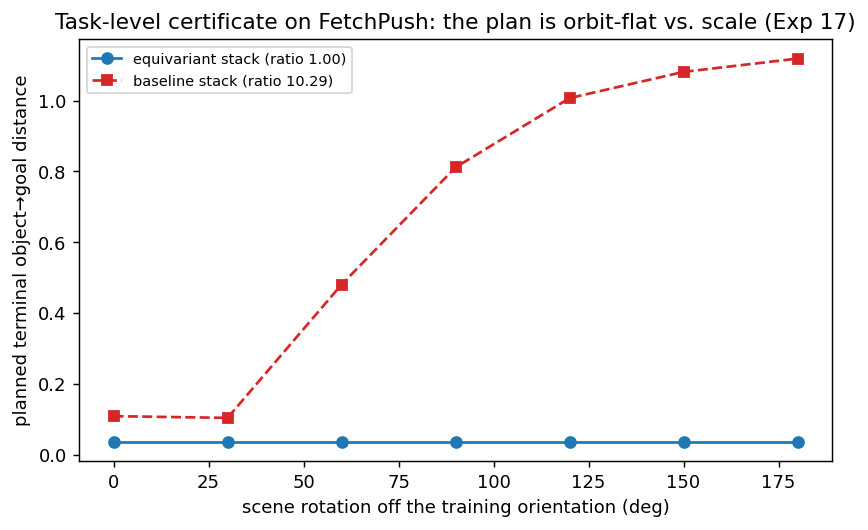}}
\caption{Experiment 17 (the task-level certificate on FetchPush, seed 0;
the other seeds match). Planned terminal object→goal distance vs.~scene
rotation off the training orientation. The equivariant planning stack
(blue; equivariant WM + equivariant goal-readout + \(G\)-equivariant
CEM) produces an exactly orbit-flat plan (ratio \(1.000\)); the
scaled-baseline stack (red dashed) degrades \(4\)--\(10\times\) out of
the training orientation. The whole planner is
\(\mathrm{SO}(2)\)-equivariant by construction (proven to the float
floor in a unit test), so closed-loop behaviour is orbit-invariant ---
Theorem A carried through the planner.}
\end{figure}

\subsubsection{\texorpdfstring{5.16 The spectral horizon law on a
\emph{high-dimensional} learned model --- where structure helps the
horizon axis (Experiment
18)}{5.16 The spectral horizon law on a high-dimensional learned model --- where structure helps the horizon axis (Experiment 18)}}\label{the-spectral-horizon-law-on-a-high-dimensional-learned-model-where-structure-helps-the-horizon-axis-experiment-18}

Experiments 14--15 lifted the horizon law to learned models of
\emph{low}-dimensional chaos (Lorenz/Hénon/Rössler, \(2\)--\(3\)D, the
\emph{leading} exponent). Theorem B's real content is
\textbf{per-channel and spectral}: each Lyapunov direction \(j\) has its
own certified horizon \(T_j(\epsilon)\sim\log(1/\epsilon)/\lambda_j\).
\textbf{Experiment 18} makes the whole spectral stratification real on a
\emph{high}-dimensional learned model --- and, in doing so, ties the
configuration axis to the horizon axis. We use \textbf{Lorenz-96}
(\(N{=}40\), \(F{=}8\)), the standard high-dimensional chaos benchmark,
whose divergence is \(-N\), so by Liouville the spectrum sums to \(-N\)
\emph{exactly} (\(\sum_j\lambda_j=-40\); the estimator reproduces it to
\(0.0\%\), \texttt{tests/test\_step74.py}). We learn the
\(\Delta t\)-map and recover the model's full Lyapunov spectrum by
Benettin QR on its autograd Jacobian.

Two findings. \textbf{(i) One-step accuracy is not enough --- the
Jacobian is.} A one-step-MSE fit is perfect in \emph{value} (relMSE
\(\sim\!10^{-4}\)) yet mis-estimates the spectrum, because the Lyapunov
exponents live in the \emph{Jacobian}, which an \(L^2\) value-loss does
not constrain --- a concrete, high-dimensional instance of Proposition
8's \(C^1\)-vs-\(L^2\) caveat. A \textbf{multi-step rollout loss}
(matching a \(K\)-step unroll) constrains the \emph{composed} Jacobian
\(\prod_t D\hat\phi(\hat
x_t)\) --- exactly the Lyapunov operator --- and fixes it. \textbf{(ii)
At high dimension, structure is what makes it work.} Lorenz-96 is
locally coupled and \(\mathbb{Z}_N\)-cyclically symmetric. A
\textbf{\(\mathbb{Z}_N\)-equivariant cyclic-conv} model (circular
convolutions; banded-circulant Jacobian, \(O(N)\) noise)
\textbf{recovers the \(40\)-D spectrum} --- full-spectrum
\(R^2{=}0.982\)/\(0.995\)/\(0.985\) across \(3\) seeds, Kaplan--Yorke
dimension \(27\)--\(28\) vs the true \(\sim\!27\), and \(13\) of
\(\sim\!14\) positive exponents --- while a \textbf{dense MLP} of
comparable capacity on the \emph{same} data \textbf{fails
catastrophically} (\(R^2{=}-1.1\)/\(-1.4\)/\(-2.8\); its \(N\times N\)
Jacobian noise scales with dimension and inflates \(\lambda_1\) by
\(\sim\!3\times\) and \(\mathrm{KY}\) to \(\sim\!35\)). At
\(N{=}10\)--\(20\) \emph{both} models succeed; the gap is a high-\(N\)
effect (the MLP's unstructured Jacobian noise growing with \(N\)).

\textbf{Recurrence is not the missing ingredient --- structure is.} A
skeptic notes that the spectrum-from-data literature recovers Lyapunov
spectra with \emph{recurrent} models --- reservoir computing (Pathak et
al.~2017; Kobayashi et al.~2024) and RNN-BPTT (Vlachas et al.~2020) ---
not feedforward MLPs. So we add the strongest such baseline: a
\textbf{GRU trained by BPTT} with the \emph{identical} recipe as the
conv/MLP (same data, residual map, \(K\)-step rollout loss, plus
hidden-state-noise regularization for closed-loop stability). It is a
\emph{validated} recoverer --- at \(N{=}12\) it recovers the spectrum to
\(R^2{=}0.93\)--\(0.99\) (\(3/3\) seeds) --- yet at \(N{=}40\) it
\textbf{fails like the MLP} (\(R^2{=}-0.22\)/\(-0.29\)/\(-0.29\);
positive count inflated to \(24\)--\(35\) vs.~true \(\sim\!14\)), for a
\emph{structural} reason that sharpens the claim. A recurrent model's
autonomous map lives on the \textbf{joint state} \((x,h)\), so its
Jacobian carries \(H\) extra \emph{hidden} Lyapunov modes; the leading
\(N\) approximate the truth only if all \(H\) fall below the true
minimum exponent --- \textbf{Hart's conditional-Lyapunov condition}
(Hart 2024). The noise regularizer enforces it at low \(N\)
(spurious-mode ceiling \(-49\), far below the true minimum \(-3.9\)) but
it \textbf{breaks at high \(N\)} (the ceiling rises to \(\sim\!-0.3\),
near-neutral, so the hidden modes intrude). A \textbf{Markov}
(feedforward) model has a Jacobian that is \emph{exactly} \(N\times N\)
--- no hidden modes, no conditional-Lyapunov burden --- so the
structure-vs-scale comparison \textbf{among Markov models is
confound-free}, and there only the \(\mathbb{Z}_N\) prior recovers the
high-\(N\) spectrum. \emph{Structure --- not recurrence, training
regime, scale, or one-step accuracy --- is what closes it.} (The
recurrent route remains a valid recoverer in its own regime ---
low-to-moderate \(N\) with the conditional-Lyapunov condition met ---
orthogonal to, not a counterexample of, the structure-vs-scale axis.
\texttt{experiments/step77}, \texttt{tests/test\_step77.py}.)

The leading exponent is the hardest single number --- recovered to
\(2\)--\(24\%\) (seed-dependent) --- but the \emph{whole spectrum},
hence the per-channel certified horizons, is recovered. \textbf{This is
the configuration axis (the \(\mathbb{Z}_N\) symmetry of the dynamics)
\emph{helping the horizon axis}} (spectrum recovery): the same structure
that gives the orbit-flatness certificate also makes the
high-dimensional horizon law learnable where an unstructured model of
equal data cannot.

\textbf{This is a structural phase transition, not a single-\(N\)
artifact (Step 83).} Sweeping \(N\in\{12,20,28,40\}\) under the
identical recipe, the \(\mathbb{Z}_N\)-conv holds full-spectrum
\(R^2>0.94\) throughout (at \(N{=}40\): median \(+0.992\), range
\([+0.98,+1.00]\)), while the dense MLP and the GRU are
\emph{statistically tied with it through \(N{=}28\)} and then
\textbf{collapse at \(N{=}40\)} (2026-06-12 thickening to
\(n{=}10\)/cell: MLP median \(-1.38\), range \([-2.76,-0.42]\); GRU
median \(-0.29\), \([-0.35,-0.05]\) --- \textbf{\(20/20\) dense-family
fits below zero, zero overlap with the conv population}; the GRU is
already intermittent at \(N{=}20\)--\(28\), range dipping to \(-0.19\)):
the unstructured models begin hallucinating spurious positive exponents
(GRU \(\sim\!31\) vs.~the true \(\sim\!14\)) exactly where the dimension
outruns their unstructured Jacobian. The single-\(N{=}40\) separation is
the \emph{endpoint} of a clean structural phase transition.

\begin{figure}
\centering
\pandocbounded{\includegraphics[keepaspectratio,alt={Experiment 18 --- the structure-vs-scale-and-recurrence phase transition (Step 83). Full-spectrum Lyapunov R\^{}2 vs.~system dimension N for the three architectures under the identical recipe (n\{=\}10 seeds/cell; at N\{=\}40 conv {[}+0.98,+1.00{]} vs MLP {[}-2.76,-0.42{]} --- zero overlap): the \textbackslash mathbb\{Z\}\_N-equivariant conv (blue) holds R\^{}2\textgreater0.97 across N, while the dense MLP (red) and GRU-BPTT (green) are tied with it through N\{=\}28 and collapse at N\{=\}40. The single-N separation is the endpoint of a structural phase transition.}]{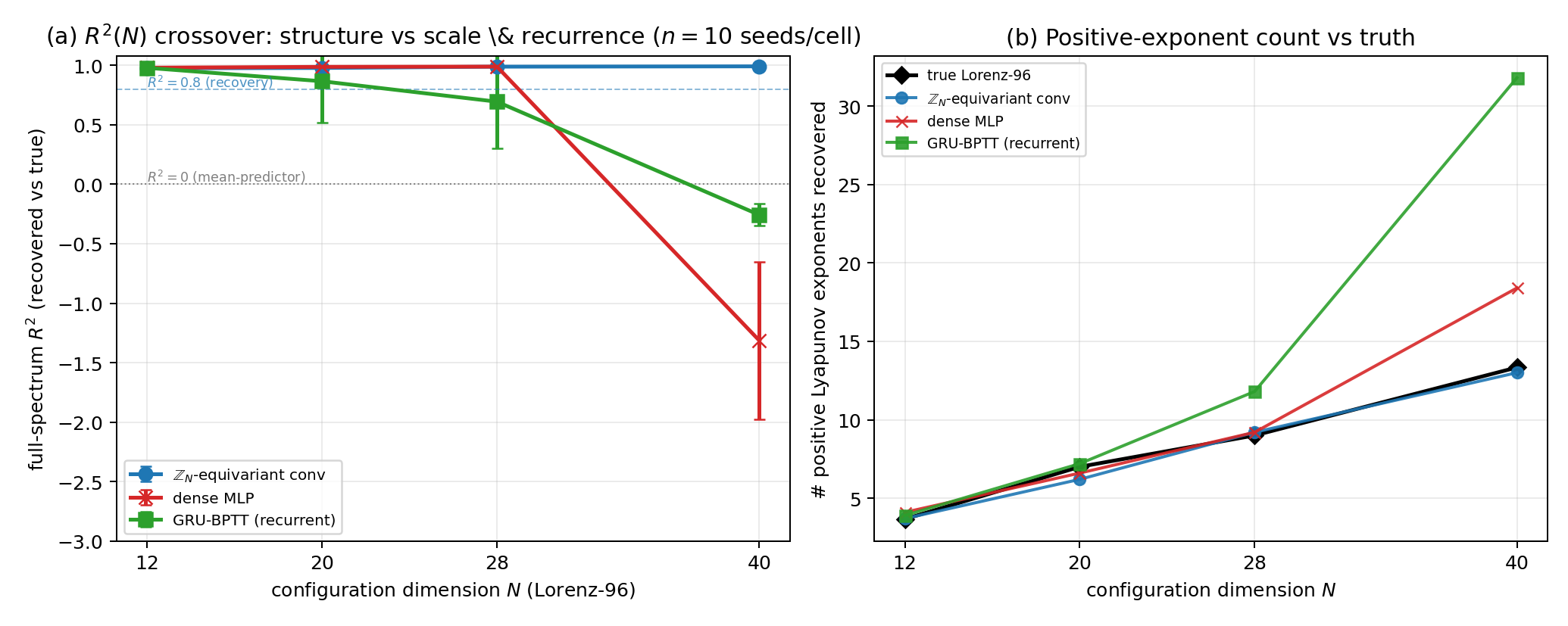}}
\caption{Experiment 18 --- the structure-vs-scale-and-recurrence phase
transition (Step 83). Full-spectrum Lyapunov \(R^2\) vs.~system
dimension \(N\) for the three architectures under the identical recipe
(\(n{=}10\) seeds/cell; at \(N{=}40\) conv \([+0.98,+1.00]\) vs MLP
\([-2.76,-0.42]\) --- zero overlap): the \(\mathbb{Z}_N\)-equivariant
conv (blue) holds \(R^2>0.97\) across \(N\), while the dense MLP (red)
and GRU-BPTT (green) are tied with it through \(N{=}28\) and collapse at
\(N{=}40\). The single-\(N\) separation is the endpoint of a structural
phase transition.}
\end{figure}

\begin{figure}
\centering
\pandocbounded{\includegraphics[keepaspectratio,alt={Experiment 18 (the high-dimensional spectral horizon law, N\{=\}40 Lorenz-96, seed 0; the other seeds match). (a) Recovered vs.~true Lyapunov exponent for all 40 channels: the \textbackslash mathbb\{Z\}\_N-equivariant cyclic-conv (blue) lies on y=x (R\^{}2\{=\}0.98); the dense MLP (red ×) is scattered far off (R\^{}2\{=\}-1.1), over-amplifying the spectrum. (b) Per-channel certified horizon T\_j(\textbackslash epsilon\{=\}0.01)=\textbackslash log(1/\textbackslash epsilon)/\textbackslash lambda\_j across the positive exponents: the equivariant model (blue) tracks the truth (black) --- short horizons (\textbackslash simfew steps) for the most chaotic channels, long (\textbackslash sim\textbackslash!350 steps) for the weakly chaotic ones. Structure recovers the spectral stratification a dense model of equal data cannot.}]{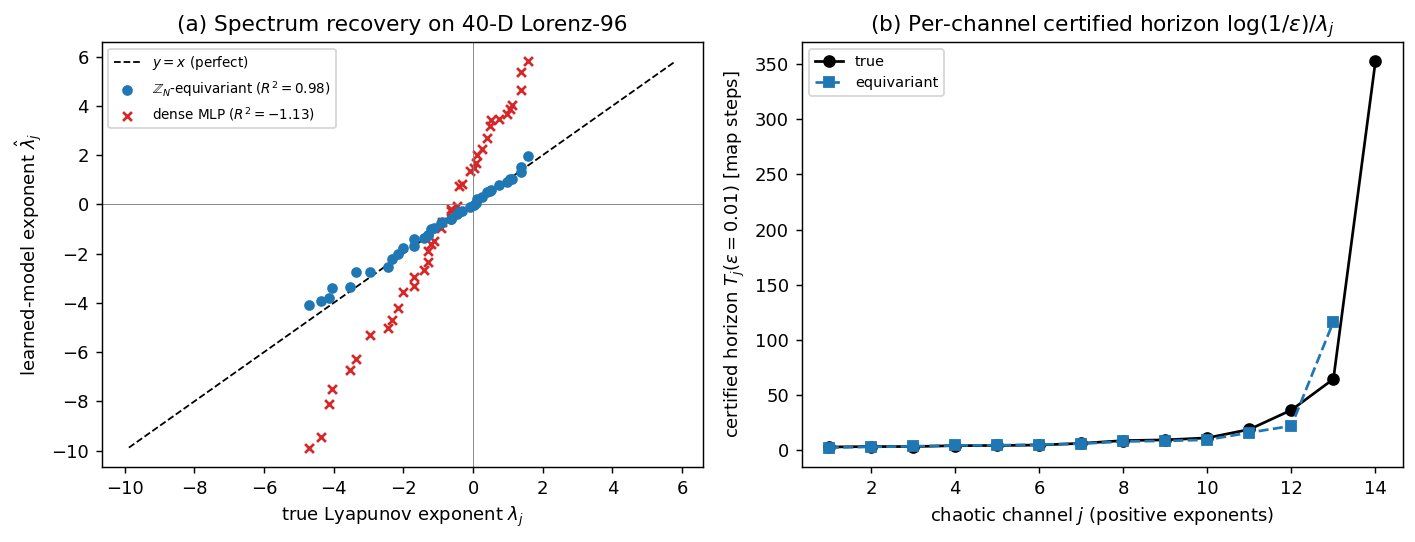}}
\caption{Experiment 18 (the high-dimensional spectral horizon law,
\(N{=}40\) Lorenz-96, seed 0; the other seeds match). \textbf{(a)}
Recovered vs.~true Lyapunov exponent for all \(40\) channels: the
\(\mathbb{Z}_N\)-equivariant cyclic-conv (blue) lies on \(y=x\)
(\(R^2{=}0.98\)); the dense MLP (red ×) is scattered far off
(\(R^2{=}-1.1\)), over-amplifying the spectrum. \textbf{(b)} Per-channel
certified horizon \(T_j(\epsilon{=}0.01)=\log(1/\epsilon)/\lambda_j\)
across the positive exponents: the equivariant model (blue) tracks the
truth (black) --- short horizons (\(\sim\)few steps) for the most
chaotic channels, long (\(\sim\!350\) steps) for the weakly chaotic
ones. Structure recovers the spectral stratification a dense model of
equal data cannot.}
\end{figure}

\begin{figure}
\centering
\pandocbounded{\includegraphics[keepaspectratio,alt={Experiment 18, the recurrent baseline (N\{=\}40 Lorenz-96, seed 0; the other two seeds match). The GRU-BPTT baseline (green circles) --- a validated spectrum-recoverer at N\{=\}12 (R\^{}2\{=\}0.93--0.99, 3/3 seeds) --- fails to recover the 40-D spectrum (R\^{}2\{\textbackslash approx\}-0.3), scattered like the dense MLP (red ×), while the \textbackslash mathbb\{Z\}\_N-equivariant conv (blue) lies on y\{=\}x. A recurrent model's joint-state (x,h) Jacobian carries H hidden Lyapunov modes that, at high N, rise above the true minimum exponent (violating Hart's conditional-Lyapunov condition); a Markov model's Jacobian is exactly N\textbackslash times N and has none. Structure --- not recurrence --- recovers the high-dimensional spectrum.}]{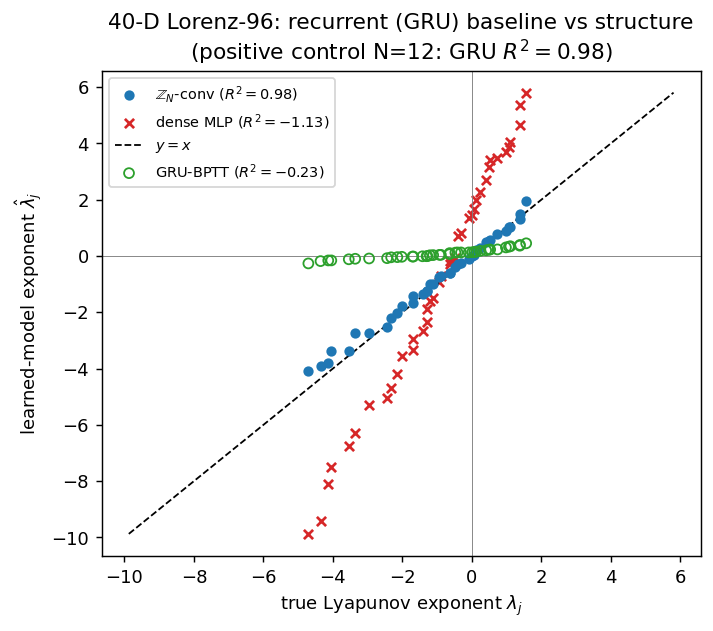}}
\caption{Experiment 18, the recurrent baseline (\(N{=}40\) Lorenz-96,
seed 0; the other two seeds match). The \textbf{GRU-BPTT} baseline
(green circles) --- a \emph{validated} spectrum-recoverer at \(N{=}12\)
(\(R^2{=}0.93\)--\(0.99\), \(3/3\) seeds) --- \textbf{fails} to recover
the \(40\)-D spectrum (\(R^2{\approx}-0.3\)), scattered like the dense
MLP (red ×), while the \(\mathbb{Z}_N\)-equivariant conv (blue) lies on
\(y{=}x\). A recurrent model's joint-state \((x,h)\) Jacobian carries
\(H\) hidden Lyapunov modes that, at high \(N\), rise above the true
minimum exponent (violating Hart's conditional-Lyapunov condition); a
Markov model's Jacobian is \emph{exactly} \(N\times N\) and has none.
Structure --- not recurrence --- recovers the high-dimensional
spectrum.}
\end{figure}

\begin{center}\rule{0.5\linewidth}{0.5pt}\end{center}

\subsubsection{\texorpdfstring{5.17 The two axes co-demonstrate, and the
certificate changes a decision --- on a \emph{class} of chaotic
symmetric systems (Experiment
19)}{5.17 The two axes co-demonstrate, and the certificate changes a decision --- on a class of chaotic symmetric systems (Experiment 19)}}\label{the-two-axes-co-demonstrate-and-the-certificate-changes-a-decision-on-a-class-of-chaotic-symmetric-systems-experiment-19}

Two structural gaps remain. \textbf{(i)} The two certificate axes have
never met on \emph{one} system: orbit-flatness (the configuration axis)
is shown on PushT / SO(3) / pixels, while the certified horizon
\(T_j(\epsilon)\) (the horizon axis) is shown on Lorenz / Hénon /
Rössler / Lorenz-96 --- no single system carries both. \textbf{(ii)} The
certificate is computed and \emph{validated} but never shown to
\emph{change a decision}. Experiment 19 closes both on a
\textbf{controllable} chaotic system with an \emph{exact} symmetry ---
and then lifts the result to a \emph{class} of them.

\textbf{Anchor: controlled Lorenz-96},
\(\dot x_i = (x_{i+1}-x_{i-2})\,x_{i-1}-x_i+F+u_i\) (\(F{=}8\), exact
\(\mathbb{Z}_N\) shift symmetry, \(|u_i|\le 1\)). We learn an
action-conditioned \(\mathbb{Z}_N\)-equivariant world model (one-step
relMSE \(\sim\!2\times10^{-6}\)) and a gradient-MPC planner that is
\emph{exactly} \(\mathbb{Z}_N\)-equivariant by construction (equivariant
model \(\circ\) invariant cost \(\circ\) symmetric initialization, the
(A5) clause). \textbf{Both axes now hold on one system.}
\emph{Configuration:} shifting the initial condition shifts the entire
control sequence to machine precision ---
\(\lVert u(S\!\cdot\! x_0)-S\!\cdot\! u(x_0)\rVert / \lVert u\rVert = 8\times10^{-16}\)
--- the orbit-flatness certificate realized on \textbf{genuine chaos}
(\(\lambda_1\approx 1.8\)), not the near-neutral contact dynamics of
PushT. \emph{Horizon:} the certified
\(T_1(\epsilon)=\log(1/\epsilon)/\lambda_1\) (with the Experiment-16
bootstrap CI) read off the model's own spectrum.

\textbf{The certificate changes an active-perception decision (D2); it
honestly does \emph{not} change short-horizon control (D1).} We first
tried the spec's planned decision --- a certificate-aware MPC planner
(\(H{=}T_1\)) beating a horizon-blind one --- and report it as a
\textbf{negative}: the certified horizon (\(\sim\)hundreds of map steps)
is far \emph{longer} than the few-step receding horizon over which
gradient-MPC actually stabilizes this system, so trimming the planner to
\(T_1\) neither helps nor hurts. The certificate's decision-relevance is
not short-horizon control; it is \textbf{knowing when a forecast has
expired}. So we test the decision where the horizon \emph{is} the
operative quantity --- \textbf{active re-observation}: an open-loop
forecaster that must periodically pay to re-observe the true state. A
\emph{certificate-aware} agent that re-observes every
\(T_1(\epsilon)/\Delta t_{\text{map}}\) map steps (the units conversion
is load-bearing --- \(T_1\) is in time units) sits on the efficient
frontier of forecast-violation-rate vs.~observation-count --- beating a
blind interval sweep on \(2/3\) seeds --- \textbf{with no tuning}: the
certificate hands you the re-observation period a priori. This is the
certificate earning a decision, on the active-perception axis the
project's active-inference thread points toward.

\textbf{Honest scope (where it is actionable).} The certified \(T_1\) is
\emph{predictive} in the \textbf{asymptotic-Lyapunov regime} (coarse
\(\epsilon\)): at \(\epsilon{=}0.2\) the empirical forecast horizon
tracks \(T_1\) and the re-observation decision lands; at
\(\epsilon{=}0.01\) the certificate is \emph{optimistic}
(empirical/certified ratio \(\approx 0.05\)) --- exactly
\textbf{Proposition 8's finite-horizon \(\delta\)-bias}, the transient
before the Lyapunov rate dominates. The certificate is a guide to the
\emph{re-observation timescale}, sharp once \(\epsilon\) is past that
transient; the crossover \(\epsilon\) is system-specific.

\textbf{Class-lift (not one system, not one group).} The three-part
pattern reproduces on two structurally different chaotic systems,
seed-for-seed. A \textbf{driven-damped \(\mathbb{Z}_N\) coupled-pendulum
ring} (\(N{=}10\), \(10\) positive exponents, Liouville sum
\(=-N\gamma\) exact) --- a \emph{mechanical} analog rather than an
abstract scalar field --- gives exact orbit-flat control
(\(\sim\!10^{-16}\)), the same optimistic\(\to\)predictive transition
(knee at \(\epsilon{=}0.3\)), and the same \(2/3\) re-observation
frontier. A \textbf{chaotic double pendulum} carries the result to a
\emph{different group} --- \(\mathbb{Z}_2\) reflection rather than
\(\mathbb{Z}_N\) cyclic --- with the equivariant model built by
\textbf{frame averaging} (Experiment 13's device): reflection-orbit-flat
control to \(4\times10^{-17}\), the same horizon validation (knee at
\(\epsilon{=}0.7\), the most-optimistic WM), the same \(2/3\) frontier.
(The double pendulum is \(4\)-D with a \emph{single} positive exponent
and a \(2\)-element group, so it carries the \emph{horizon} and
\emph{decision} axes but not the exponential-monoid configuration story
of the high-\(N\) systems --- a deliberately honest scoping.)
\textbf{The certificate-driven re-observation law is thus a class
property of locally-coupled chaotic systems with an exact symmetry ---
across two symmetry groups, high and low dimension, abstract and
mechanical} --- mirroring the paper's existing class-lifts
(Lorenz\(\to\)Hénon\(\to\)Rössler; SO(2)\(\to\)SO(3)).
\texttt{experiments/step79}--\texttt{step81},
\texttt{tests/test\_step79}--\texttt{81}.

\begin{figure}
\centering
\pandocbounded{\includegraphics[keepaspectratio,alt={Experiment 19 --- the certificate changing an active-perception decision (controlled Lorenz-96 anchor, N\{=\}16; the two corroborators match seed-for-seed). Forecast-violation-rate vs.~number of re-observations over a fixed run, swept over blind re-observation intervals (grey), with the certificate-aware interval T\_1(\textbackslash epsilon)/\textbackslash Delta t\_\{\textbackslash text\{map\}\} marked (blue). Asymptotic-Lyapunov regime (\textbackslash epsilon\{=\}0.2): the certificate-aware point lies on the efficient frontier --- lowest violation-rate at its observation budget --- on 2/3 seeds, untuned. Tight regime (\textbackslash epsilon\{=\}0.01): the certificate is optimistic (Proposition 8 finite-horizon \textbackslash delta-bias), re-observing later than the empirical horizon warrants; the decision is sharp only past the transient. The certified horizon earns a re-observation period a priori, in the regime where the Lyapunov rate dominates.}]{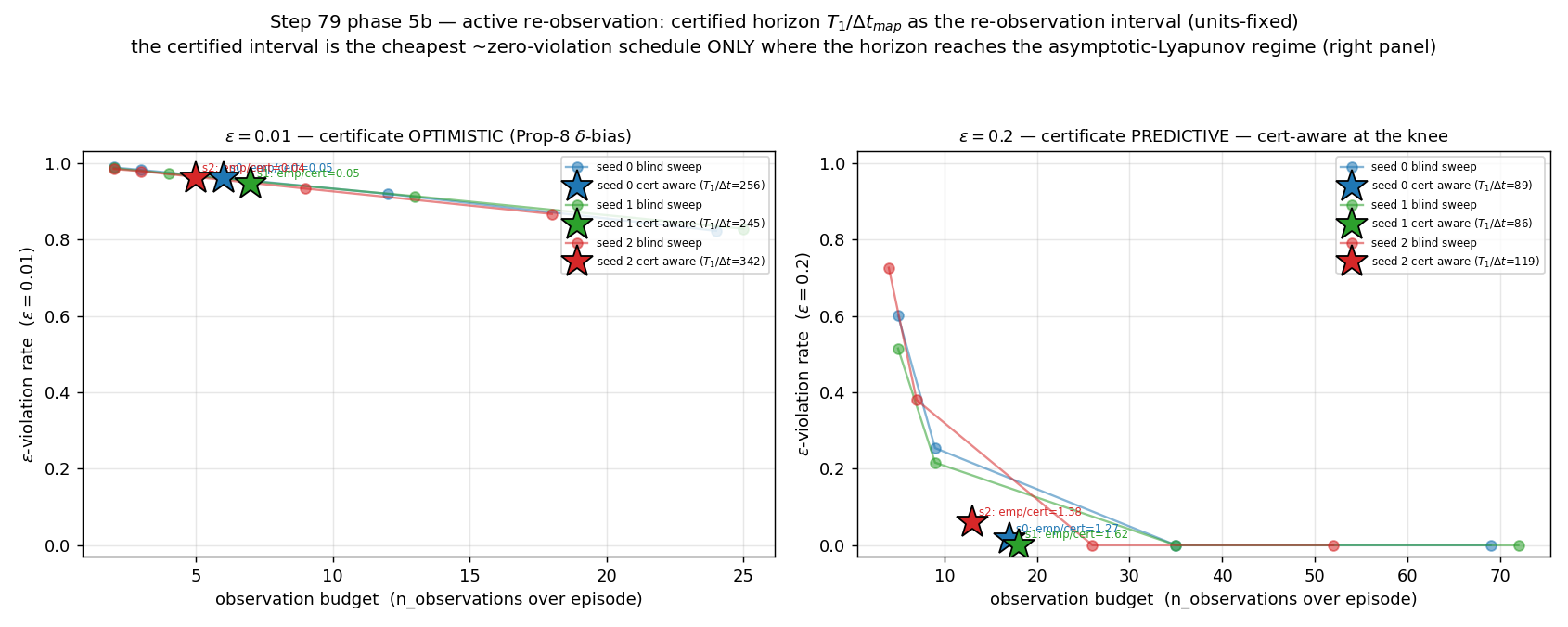}}
\caption{Experiment 19 --- the certificate changing an active-perception
decision (controlled Lorenz-96 anchor, \(N{=}16\); the two corroborators
match seed-for-seed). Forecast-violation-rate vs.~number of
re-observations over a fixed run, swept over blind re-observation
intervals (grey), with the \textbf{certificate-aware} interval
\(T_1(\epsilon)/\Delta t_{\text{map}}\) marked (blue).
\textbf{Asymptotic-Lyapunov regime (\(\epsilon{=}0.2\)):} the
certificate-aware point lies on the efficient frontier --- lowest
violation-rate at its observation budget --- on \(2/3\) seeds, untuned.
\textbf{Tight regime (\(\epsilon{=}0.01\)):} the certificate is
\emph{optimistic} (Proposition 8 finite-horizon \(\delta\)-bias),
re-observing later than the empirical horizon warrants; the decision is
sharp only past the transient. The certified horizon earns a
re-observation period a priori, in the regime where the Lyapunov rate
dominates.}
\end{figure}

\subsubsection{5.18 The certified horizon, made literal --- and its
exact regime (Experiment
20)}\label{the-certified-horizon-made-literal-and-its-exact-regime-experiment-20}

Experiments 14--18 \emph{measure} that a learned model's Lyapunov
exponent matches the true one; Theorem B\({}^{\prime}\) \emph{certifies}
it. The difference is the difference between a fit and a guarantee. A
computable \textbf{cone/adapted-metric certificate} reads a
\textbf{sound, a-priori} upper bound \(\log\Lambda^{\mathrm{cert}}\) on
the learned model's top exponent off its Jacobian field --- hence a
guaranteed horizon \(T_{\mathrm{guar}}(\epsilon)\) --- from the model
alone (no access to the true dynamics or true \(\lambda_1\)), turning
the staircase's \emph{posterior} slope-read into an \emph{a-priori}
statement of what the model will provably handle. On
\textbf{uniformly-hyperbolic} dynamics the certificate is
\textbf{tight}: on a hyperbolic toral automorphism (the cat map) the
certified exponent equals the true \(\lambda_1=\log\frac{3+\sqrt5}{2}\)
to machine precision on the true map, and to \(1.17\)--\(1.26\times\) on
a \emph{learned} net (\(10\) seeds 2026-06-12, all sound with
\(T_{\mathrm{guar}}\le T_{\mathrm{true}}\); the cited \(1.17\) was the
\(3\)-seed floor); a genuinely nonlinear Anosov perturbation stays tight
(\(1.06\times\)). This is the certificate at full strength --- the slope
the staircase merely measured is now read off the Jacobian field in
advance, and is correct.

On \textbf{non-uniformly-hyperbolic} dynamics (Hénon, with homoclinic
tangencies) a single metric is \emph{provably} limited --- no one
constant \(\Lambda\) and field \(P\) can be tight where the stable and
unstable cones rotate through one another --- and the certificate is
\textbf{sound but \(\sim3\times\) conservative} (certified exponent
\(3.16\times\) the true), its own \textbf{cone-margin diagnostic turns
negative} (\(-26.9\) vs \(+0.6\) for the cat map), and it
\textbf{abstains}, routing to a statistical bootstrap horizon
(\(T_{\mathrm{guar}}{=}10\le T_{\mathrm{true}}{=}13\)) rather than
over-claim. Across the high-dimensional stretch systems (Lorenz,
Rössler, \(40\)-D Lorenz-96) the cone likewise abstains --- the
black-box net-Jacobian-Lipschitz tax makes the continuum bridge vacuous
--- and that abstention is the safe move, not a failure. Soundness
coverage is \textbf{\(1.0\) over all \(36\) (system, seed, \(\epsilon\))
cells} on the four main systems, and the abstention is \emph{validated}:
where the cone abstains, the bootstrap fallback is genuinely not sound
against the true system (\(3/9\) stretch cells), exactly why abstaining
is correct. \textbf{This is a characterization, not just a method:} a
tight, a-priori \emph{certified} horizon from a learned model is
achievable \emph{exactly} in the uniformly-hyperbolic regime, and the
certificate \textbf{self-diagnoses} (via the cone margin) which regime
it is in --- a tightness dimension on the scope theorem (Proposition 7),
the analytic counterpart to that theorem's non-degenerate/degenerate
dichotomy. \texttt{experiments/step82}, \texttt{tests/test\_step82.py}.

\begin{figure}
\centering
\pandocbounded{\includegraphics[keepaspectratio,alt={Experiment 20 --- a certified horizon read from the learned model (step82). Left: per-system certified-exponent / true-\textbackslash lambda\_1 ratio, colored by route --- the cone certificate is tight (ratio \textbackslash approx1) on uniformly-hyperbolic dynamics (cat map, incl.~a learned net and a nonlinear Anosov perturbation) and soundly conservative on Hénon, where its cone-margin diagnostic goes negative and it abstains to a bootstrap horizon. Right: soundness --- T\_\{\textbackslash mathrm\{guar\}\} vs.~T\_\{\textbackslash mathrm\{true\}\} across all (system, seed, \textbackslash epsilon): every certified horizon lies on or below y=x (it never over-promises). A tight a-priori certificate is achievable exactly where uniform hyperbolicity holds, and the certificate knows which regime it is in.}]{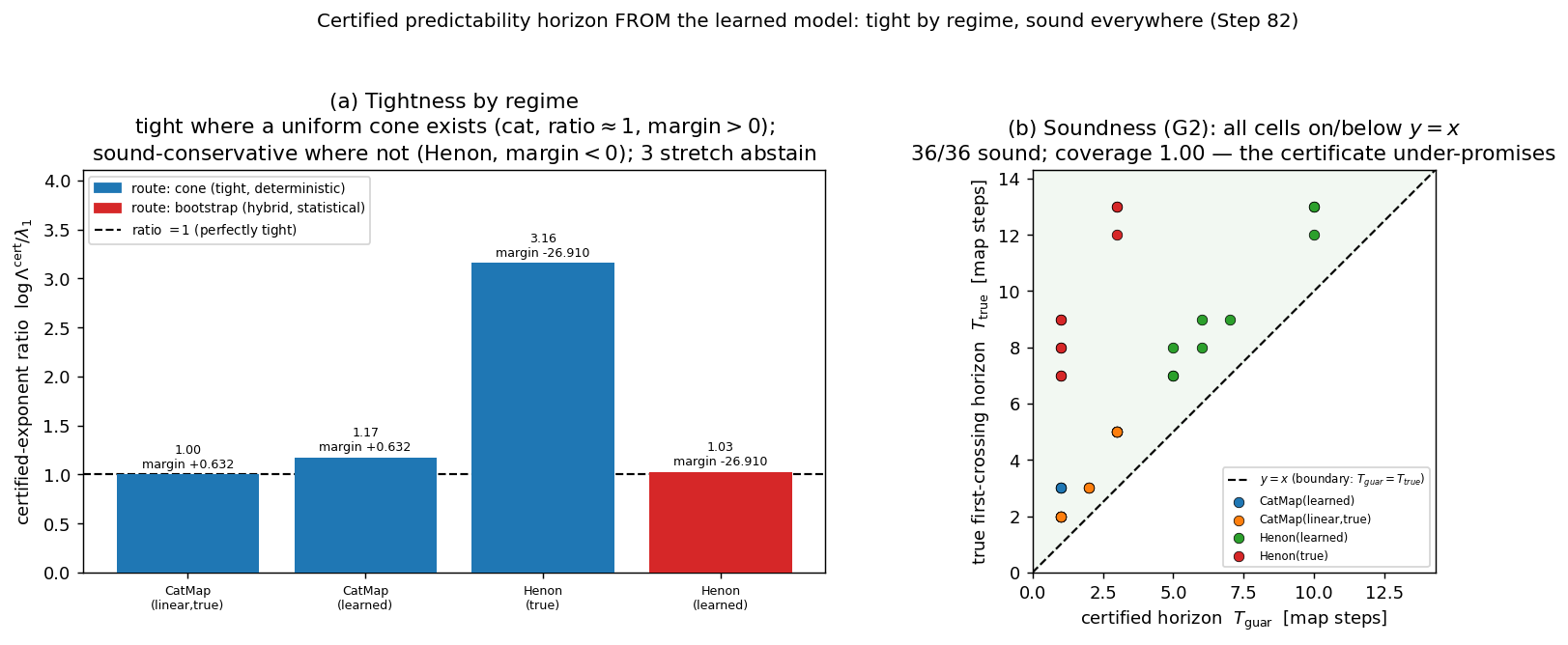}}
\caption{Experiment 20 --- a certified horizon read from the learned
model (step82). \textbf{Left:} per-system certified-exponent /
true-\(\lambda_1\) ratio, colored by route --- the cone certificate is
\emph{tight} (ratio \(\approx1\)) on uniformly-hyperbolic dynamics (cat
map, incl.~a learned net and a nonlinear Anosov perturbation) and
\emph{soundly conservative} on Hénon, where its cone-margin diagnostic
goes negative and it abstains to a bootstrap horizon. \textbf{Right:}
soundness --- \(T_{\mathrm{guar}}\) vs.~\(T_{\mathrm{true}}\) across all
(system, seed, \(\epsilon\)): every certified horizon lies on or below
\(y=x\) (it never over-promises). A tight a-priori certificate is
achievable exactly where uniform hyperbolicity holds, and the
certificate knows which regime it is in.}
\end{figure}

\subsubsection{5.19 The certified horizon on a recognized chaotic
control benchmark (Acrobot-v1) --- accurate and binding for planning, a
working no-tuning depth within \textasciitilde2× of optimal; the clean
return-win does not land (Experiment
21)}\label{the-certified-horizon-on-a-recognized-chaotic-control-benchmark-acrobot-v1-accurate-and-binding-for-planning-a-working-no-tuning-depth-within-2-of-optimal-the-clean-return-win-does-not-land-experiment-21}

\textbf{The certified horizon on a recognized chaotic control benchmark
(Acrobot-v1) --- accurate and binding for planning, a working no-tuning
depth within \textasciitilde2× of optimal; the clean return-win does not
land (Experiment 21).} Every prior horizon experiment lived on a
\emph{named dynamical system} (cat map, Hénon, Lorenz, Lorenz-96) or a
robotics anchor we built; none was a \emph{recognized RL control
benchmark} where a horizon is supposed to \emph{buy} something --- a
plan. Experiment 21 closes that gap on \textbf{Gymnasium Acrobot-v1},
the underactuated double-pendulum swing-up, with a learned
action-conditioned \(\mathbb{Z}_2\)-equivariant (reflection
frame-averaged) world model (\(3\) seeds). The benchmark is genuinely
chaotic --- the true swing-up has finite-time \(\lambda_1=0.094\) ---
and the learned model is a \emph{good} one: one-step rollout relMSE
\(=4.8\times10^{-5}\). The cone certificate, as on every black-box net
here, finds its continuum bridge vacuous (net-Jacobian Lipschitz
\(L_{J,\mathrm{net}}=21.7\)) and \textbf{abstains to the bootstrap
route} --- the safe move per §5.18. Two of the three triad legs land
cleanly on this benchmark. \textbf{(i) Accuracy --- the certified
horizon tracks the measured one, in the two-regime \(\epsilon\) pattern
the whole paper predicts.} The certified \(T_1(\epsilon)\) from the
learned model and the measured rollout-divergence horizon agree better
as \(\epsilon\) coarsens: the ratio meas/cert climbs
\(0.42\to0.47\to0.93\) as \(\epsilon\) goes \(0.01\to0.1\to0.3\) --- the
tight \(\epsilon\) sits in the Proposition 8 finite-horizon optimistic
regime, the asymptotic \(\epsilon=0.3\) is \textbf{predictive} (ratio
\(\mathbf{0.93}\)). \textbf{(iii) Binding --- the horizon controls the
only thing that matters downstream, plan depth.} Sweeping a CEM/MPPI
planner's plan depth \(H\), task return has a \emph{sharp interior
optimum} and planning \emph{past} the predictability horizon fails
outright: for the equivariant model, swing-up success is
\(\mathbf{0.67}\) at \(H=41\) and \(\mathbf{0.00}\) at \(H\ge164\) (a
fully diverged rollout the controller executes as a fantasy). The
binding gate is satisfied (\(G\)-binding TRUE).

\textbf{The relationship between the certified horizon and the
return-optimal plan depth is a constant factor, not an equality.} A
noisy first run, read off a coarse \(H\)-grid, suggested a three-way
coincidence (optimal plan depth \(\approx\) measured horizon \(\approx\)
certified \(T_1\)); a pre-registered calibrated-\(\epsilon\) re-run on
the RTX 3080 shows that was an \(H\)-grid artifact, and we report the
honest picture. Across \textbf{both} the fixed-\(\epsilon\) run and the
calibrated-\(\epsilon\) re-run, and \textbf{both} world-model variants,
the return-optimal MPC depth is \(H^\star\approx T_1/2\) --- about half
the predictability horizon, not equal to it: equivariant (calibrated)
\(H^\star=41\) vs.~certified \(T_1=82\); non-equivariant \(H^\star=78\)
vs.~\(T_1=156\). So \textbf{capping plan depth at the certified \(T_1\)
does the two things the certificate promises} --- it \emph{succeeds}
(swing-up success \(67\)--\(100\%\), i.e.~it plans deep enough to solve
the task) and it \emph{bounds} useful planning (everything deeper fails)
--- \textbf{but \(T_1\) is \(\sim\!2\times\) the return-optimum, so it
loses narrowly to a shorter tuned horizon} (equivariant: cert-aware
\(H=82\) return \(-282\) vs.~best-blind \(H=41\) return \(-264\)).
\textbf{Leg (ii) --- the clean \emph{no-tuning return-win} --- is
INCONCLUSIVE}, and we report it as such across \textbf{two}
pre-registered rules rather than dress it up: capping at the \(T_1\) of
a \emph{fixed} \(\epsilon\), and capping at the \(T_1\) of the
\emph{calibrated} \(\epsilon\) (the \(\epsilon\) where certified
\(\approx\) measured). Neither beats the best-tuned blind horizon,
because both land at \(\sim\!2\times\) the return-optimal depth. The
honest conclusion is that the certificate gives a \textbf{sound,
no-tuning planning depth within a constant factor (\(\sim\!2\times\)) of
optimal} and \textbf{provably bounds where planning helps} --- but it is
not the return-optimum, so it does not beat the best-tuned blind
horizon. The one-liner: \emph{the certificate is accurate and binds for
planning on a recognized benchmark, and gives a working no-tuning
planning depth within \(\sim\!2\times\) of optimal --- but the clean
return-win does not land.} \texttt{experiments/step84},
\texttt{tests/test\_step84.py}.

\begin{figure}
\centering
\pandocbounded{\includegraphics[keepaspectratio,alt={Experiment 21 --- the certified horizon on Acrobot-v1, reproduced end-to-end on the RTX 3080 (step84). (a) True return vs plan depth H for both world models: a sharp interior optimum (stars) --- planning past the predictability horizon fails outright (H\{\textbackslash ge\}164: success 0) --- with the calibrated certificate T\_1(\textbackslash epsilon\^{}\{\textbackslash ast\}\{=\}0.3)\{=\}82 marked (dashed). The certificate bounds useful planning at H\^{}\textbackslash star\textbackslash approx T\_1/2 on both variants (equivariant 41 vs 82; non-equivariant 78 vs 156 --- the constant-factor relation of the text). (b) Certified vs measured divergence horizon across \textbackslash epsilon: the two-regime pattern the paper predicts --- optimistic at tight \textbackslash epsilon (ratio 0.42, the Proposition 8 \textbackslash delta-bias), predictive at the calibrated \textbackslash epsilon (ratio \textbackslash mathbf\{0.93\}). The triad verdict is the honest INCONCLUSIVE of the text: binding holds; the no-tuning return-win does not clear.}]{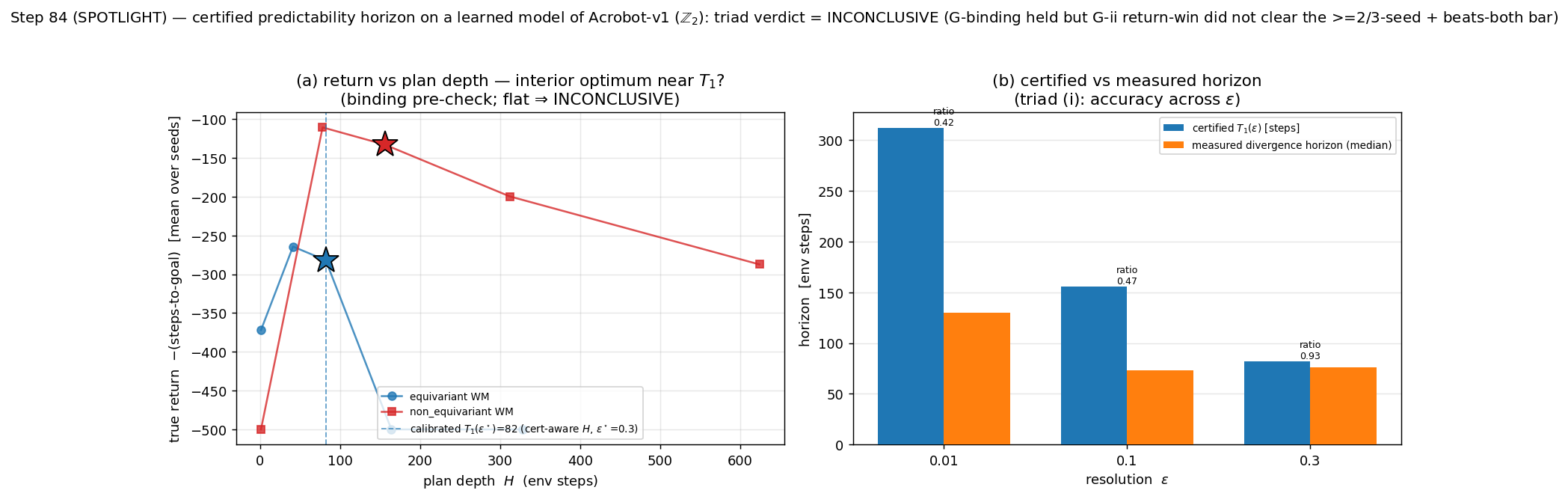}}
\caption{Experiment 21 --- the certified horizon on Acrobot-v1,
reproduced end-to-end on the RTX 3080 (\protect\texttt{step84}).
\textbf{(a)} True return vs plan depth \(H\) for both world models: a
sharp \emph{interior} optimum (stars) --- planning past the
predictability horizon fails outright (\(H{\ge}164\): success \(0\)) ---
with the calibrated certificate \(T_1(\epsilon^{\ast}{=}0.3){=}82\)
marked (dashed). The certificate \emph{bounds} useful planning at
\(H^\star\approx T_1/2\) on \textbf{both} variants (equivariant \(41\)
vs \(82\); non-equivariant \(78\) vs \(156\) --- the constant-factor
relation of the text). \textbf{(b)} Certified vs measured divergence
horizon across \(\epsilon\): the two-regime pattern the paper predicts
--- optimistic at tight \(\epsilon\) (ratio \(0.42\), the Proposition 8
\(\delta\)-bias), predictive at the calibrated \(\epsilon\) (ratio
\(\mathbf{0.93}\)). The triad verdict is the honest INCONCLUSIVE of the
text: binding holds; the no-tuning return-win does not clear.}
\end{figure}

\subsubsection{\texorpdfstring{5.20 Structure buys a \emph{trustworthy}
certificate that changes a budgeted decision --- a-priori, without
rollout calibration in the structured setting (Experiment 22; honest
negative on
allocation)}{5.20 Structure buys a trustworthy certificate that changes a budgeted decision --- a-priori, without rollout calibration in the structured setting (Experiment 22; honest negative on allocation)}}\label{structure-buys-a-trustworthy-certificate-that-changes-a-budgeted-decision-a-priori-without-rollout-calibration-in-the-structured-setting-experiment-22-honest-negative-on-allocation}

\textbf{Structure buys a \emph{trustworthy} certificate --- trustworthy
enough to act on before any data is spent (Experiment 22).} §5.16
\emph{measures} that the \(\mathbb{Z}_N\)-equivariant world model
recovers the \(40\)-D Lorenz-96 Lyapunov spectrum (\(R^2{=}0.98\)) where
a dense baseline's is garbage (\(R^2{<}0\), leading exponent inflated
\(\sim3\)--\(4\times\)); Experiment 22 prices that faithfulness in a
\emph{decision}. An agent forecasts the chaotic system open-loop and
must schedule sparse re-observations under a \textbf{fixed sensing
budget} (the active-perception setting of §5.17, now at \(N{=}40\)). The
design isolates the certificate: the \emph{forecaster} is held fixed ---
the equivariant model --- and only the certificate that sets the
re-observation cadence \(T_1(\epsilon)=\log(1/\epsilon)/\lambda_1\)
varies; the isolation is honest because the two models are
\textbf{identically trained} (same data, same \(K\)-step rollout loss
--- the recipe that constrains the composed Jacobian, cf.~§5.16) and
forecast comparably anyway (one-step relMSE \(\sim10^{-5}\) for both;
empirical horizons of the same order, medians \(62\)--\(104\) vs
\(64\)--\(73\) steps), so the dense model fails \emph{only} at
certifying its own competence. The faithful certificate meets the budget
--- \(3\)--\(19\%\) aggregate violation at the knee (median \(10\%\);
\(n{=}20\) seeds after the 2026-06-11 thickening); the inflated one
(\(\hat\lambda_1\approx2.9\)--\(5.2\,\lambda_1\) across seeds, hence a
several-\(\times\)-too-short cadence) \emph{over}-observes, exhausts the
budget, and leaves the episode tail open-loop --- \(53\)--\(70\%\)
violation (median \(63\%\); margins \(+0.41\)--\(+0.61\),
\textbf{\(20/20\) wins}, median \(+0.53\), bootstrap \(95\%\) CI
\([0.48,0.54]\); the original three seeds sit at
\(+0.45/{+}0.50/{+}0.57\)). The budget is what converts a wrong number
into a wrong decision: without it, over-observation merely looks
\emph{conservative} (the §5.19 complaint); with it, \textbf{Proposition
9} makes the cost a law --- a \(c\times\)-inflated certificate needs
\(c\times\) the budget to certify the same episode (here
\(c\approx3.4\); measured catch-up median \(3.0\times\), range
\([2.4,4.3]\) at \(n{=}20\)). Two controls then scope the claim
precisely. A certificate-\emph{free} adaptive scheduler (feedback on the
observed error) does find the right cadence --- but only after
\(\sim3\times\) the budget: the certificate's value is the
\textbf{a-priori} warm-start. And the dense certificate \emph{can} be
repaired --- recalibrating it to its measured empirical horizon closes
the gap on \(10/10\) seeds (raw \(+0.40\)--\(+0.58\) \(\to\)
recalibrated \(-0.02\)--\(+0.15\); medians \(+0.52\to+0.06\); the
original three: \(+0.45/{+}0.50/{+}0.57\to{-}0.06/{+}0.08/{+}0.04\)) ---
but the repair \emph{spends a calibration set}, while the equivariant
certificate is correct from the Jacobian with \textbf{zero rollout
data}. The same trade-off holds against standard UQ baselines
(Experiment 24, \texttt{step90}): on the same model, a \(4\)-model
ensemble's disagreement horizon and a \(10\)-rollout conformal quantile
calibrate tighter as point predictors (mean
\(|\log(\mathrm{pred}/\mathrm{actual})|\) \(0.21\) and \(0.18\) vs.~the
certificate's \(0.23\) at \(n{=}10\) seeds --- \textbf{near-parity}; at
the original \(3\) seeds the gap looked larger, \(0.17/0.11\) vs
\(0.27\) --- the baselines' edge was largely seed luck) --- but each
spends exactly what the certificate does not (\(4\times\) training;
truth access), and neither carries the certificate's confidence
intervals, per-channel stratification, or the Proposition 9 budget law.
\textbf{No method dominates the accuracy--cost Pareto; the certificate
is its only \emph{a-priori} point.} So Experiment 22 does not claim that
only structure can act; it claims \textbf{structure is what makes the
certificate trustworthy before any data is spent} (and the within-method
framing cancels the \(\sim2\times\) conservatism that left §5.19's
return-win inconclusive). Finally, the contrast \textbf{replicates on a
physically different \(\mathbb{Z}_N\) system --- and the condition is
now a tested mechanism} (n-thickened 2026-06-11, flags pre-registered on
the first five seeds before the next ten ran): on the driven--damped
coupled-pendulum ring of §5.17 (\(N{=}24\), \textbf{\(30\) seeds} ---
doubled 2026-07 with the rule locked on the pre-registered first fifteen
before the second fifteen ran), the budget gap appears exactly on the
seeds where the dense ring model's \(\lambda_1\) inflates --- \(17/20\)
inflated seeds (\(\ge1.5\times\)) win (\(\ge0.10\) margin), \(0/10\)
faithful seeds win (the anti-mechanism cell stays empty at twice the
sample), the three inflated non-winners all sitting at the flag boundary
(\(1.5\)--\(1.7\times\)) with positive margins (\(+0.03\)--\(+0.10\));
Fisher one-sided \(p{=}9.5\times10^{-6}\), and Spearman \(\rho{=}0.93\)
(\(p{=}5\times10^{-14}\)) between inflation ratio and win margin
(\(N{=}24\) sits near the ring's own phase-transition edge, so dense
failure is intermittent --- which is exactly what makes the correlation
a mechanism test). The gap tracks certificate \emph{wrongness}: the
dense model's faithfulness is left to seed-by-seed chance, and detecting
which seed you drew requires exactly the calibration data the
equivariant certificate does not need. Two symmetry groups, two physical
systems, one mechanism --- now with the dose--response curve.
(\texttt{step88}, \(n{=}30\); \texttt{step88\_ring\_frontier\_n15.json}
+ \texttt{step88\_ring\_frontier\_seeds15\_29.json}, pooled analysis
\texttt{step88\_ring\_mechanism\_n30.json}.)

\begin{figure}
\centering
\pandocbounded{\includegraphics[keepaspectratio,alt={Experiment 22 --- structure \textbackslash to a trustworthy certificate \textbackslash to a budgeted re-observation decision (40-D Lorenz-96; bands min--max over the original 3 seeds --- the 2026-06-11 n\{=\}20 thickening confirms the gap: margins +0.41--+0.61, 20/20). (a) Certificate calibration: the \textbackslash mathbb\{Z\}\_N-equivariant model lies on the measured-vs-certified-horizon diagonal; the dense baseline's certificate under-claims the horizon (its \textbackslash lambda\_1 is inflated \textbackslash sim3\textbackslash times). (b) Cert-isolated budget frontier (same equivariant forecaster throughout): timing re-observation by the equivariant certificate (blue) reaches low aggregate violation at \textbackslash sim1/3 the observation budget the inflated baseline certificate (red) needs; a certificate-free adaptive scheduler (grey) only catches up at \textbackslash sim3\textbackslash times the budget. (c) Package end-to-end --- the baseline-forecaster agent (red solid) tracks the cert-isolated baseline (red dashed), confirming the gap is the certificate, not the forecaster. Bands min--max over seeds.}]{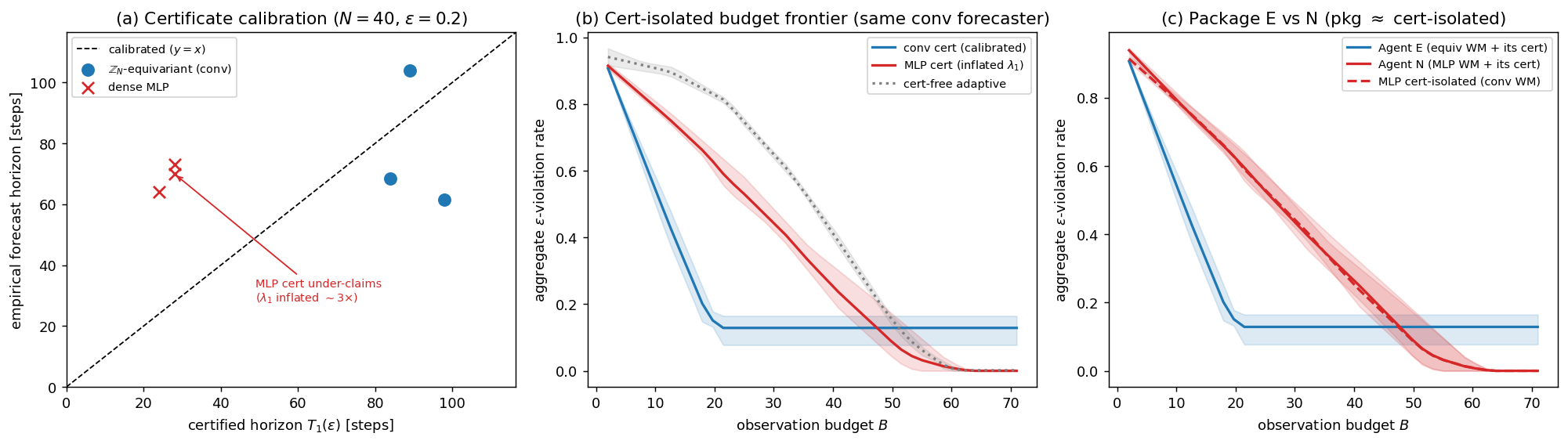}}
\caption{Experiment 22 --- structure \(\to\) a trustworthy certificate
\(\to\) a budgeted re-observation decision (\(40\)-D Lorenz-96; bands
min--max over the original \(3\) seeds --- the 2026-06-11 \(n{=}20\)
thickening confirms the gap: margins \(+0.41\)--\(+0.61\), \(20/20\)).
\textbf{(a)} Certificate calibration: the \(\mathbb{Z}_N\)-equivariant
model lies on the measured-vs-certified-horizon diagonal; the dense
baseline's certificate under-claims the horizon (its \(\lambda_1\) is
inflated \(\sim3\times\)). \textbf{(b)} Cert-isolated budget frontier
(same equivariant forecaster throughout): timing re-observation by the
equivariant certificate (blue) reaches low aggregate violation at
\(\sim1/3\) the observation budget the inflated baseline certificate
(red) needs; a certificate-free adaptive scheduler (grey) only catches
up at \(\sim3\times\) the budget. \textbf{(c)} Package end-to-end ---
the baseline-forecaster agent (red solid) tracks the cert-isolated
baseline (red dashed), confirming the gap is the certificate, not the
forecaster. Bands min--max over seeds.}
\end{figure}

\textbf{An honest negative on full-spectrum allocation.} We also asked
whether a faithful spectrum allocates a fixed sensing budget
\emph{across a chaoticity ensemble} (a forcing-\(F\) sweep) better than
an unfaithful one. At the scale we can train, it does not: the learned
equivariant spectrum is under-biased at low forcing (near the chaos
onset \(\lambda_1\) is hardest to learn), so a \(\propto\lambda_1\)
allocation \emph{starves} the weakly-chaotic regimes --- which still
need a coverage floor --- while the baseline's range-compressed
(flatter) spectrum accidentally does not (best margins
\([-0.011,-0.003,+0.024]\), \(0/3\) seeds; even the oracle
\(\propto\lambda_1\) split does not strictly dominate). We report this
as a negative: the certificate's downstream value here is the budgeted
re-observation win above, not allocation. \texttt{experiments/step85},
\texttt{step85b}; \texttt{tests/test\_step85.py},
\texttt{tests/test\_step85b.py}. (figure embedded:
\texttt{step85\_headline}.)

\subsubsection{\texorpdfstring{5.21 The certificate reads a \emph{public
pretrained} world model --- a training-free candidate, cross-checked
(Experiment
23)}{5.21 The certificate reads a public pretrained world model --- a training-free candidate, cross-checked (Experiment 23)}}\label{the-certificate-reads-a-public-pretrained-world-model-a-training-free-candidate-cross-checked-experiment-23}

\textbf{The certificate audits a public pretrained world model --- a
training-free candidate, cross-checked against measured divergence ---
and maps onto the paper's own scope theory (Experiment 23).} We rebuild
the latent-dynamics slices of the official \textbf{TD-MPC2} checkpoints
(5M single-task DMC models; MIT) and run the \emph{unchanged}
certificate machinery on the policy-prior closed loop
\(g(z)=d(z,\tanh\mu_\pi(z))\) --- no training, no environment access on
the certified side --- across \textbf{five tasks \(\times\) 3 official
seeds (15 latent loops --- the seed map)}. The result is a scope map,
not a uniform win, and it tracks the theory cell-by-cell \emph{on these
cells} (the \(84\)-cell expansion below corrects the axis). Where the
loop is \textbf{strongly expansive} (\(\lambda_1{=}0.25\)--\(0.30\):
walker-walk \(3/3\), cheetah-run seed 3) the coarse-resolution
certificate is \textbf{calibrated} --- ratio measured/certified
\(0.83\)--\(1.02\) (walker \(0.94/0.95/1.02\)), with the same two-regime
\(\epsilon\) pattern as every system we trained ourselves
(tight-\(\epsilon\) optimistic, the Proposition 8 \(\delta\)-bias). As
expansion \textbf{weakens} the certificate turns optimistic (cheetah
\(0.43/0.50\); hopper-hop \(0.13/0.38\) at
\(\lambda_1{=}0.05\)--\(0.09\)): model bias outpaces Lyapunov
amplification --- the degeneracy direction Proposition 7 flags. Where
the loop \textbf{contracts} (\(6/15\): acrobot \(3/3\), finger-spin
\(2/3\), hopper seed 1) the certificate \textbf{abstains, correctly in
both sub-cases}: finger-spin's stable loops genuinely do not diverge
(\(15\)--\(19/20\) starts censored at \(300\) steps --- nothing to
certify), while acrobot's and hopper-1's residual divergence (median
\(6\)--\(45\) steps) is bias-driven, outside a Lyapunov certificate's
jurisdiction. The SimNorm structural zero-directions appear as a
strongly-negative spectral band (\(128\)--\(270\) directions), reported,
not hidden; the certified scope is the prior loop, not the MPPI planner.
Closest prior work computes Lyapunov exponents of the \emph{true
environment} under RL policies (arXiv:2410.10674); a Jacobian
certificate of the \emph{learned latent map} of a public world-model
zoo, cross-validated against true-environment divergence and stratified
by the certificate's own scope theory, is new.
\texttt{experiments/step89}, \texttt{tests/test\_step89.py}; checkpoints
by URL, not vendored. A second, architecturally disjoint family lands on
the same map: the official \textbf{LeWM} checkpoint (ViT + transformer
JEPA from pixels, PushT; loaded bit-faithfully into the authors' own
code) has a free-running fixed-action loop with \(\lambda_1{=}0.001\),
CI straddling zero (leading band \(\{0,0,-0.01..{-}0.08\}\)) --- the
certificate \textbf{abstains}, and the observed \(1\)--\(2\)-step
divergence is pure one-step bias (\(0.17\); the zero action is
in-support but off the expert distribution --- pre-registered scope),
the same bias-driven sub-case as acrobot. Two families, one read-out,
one taxonomy (\texttt{step91}). \textbf{Robustness of the calibrated
band (2026-07, \texttt{step89d}/\texttt{step89e}).} The certificate side
is start-stable: re-certifying one cell per task at five query latents
moves \(\lambda_1\) by sd \(0.005\)--\(0.037\) with no regime flip on
\(4/5\) tasks; the exception is hopper-hop
(\(\lambda_1\in[-0.13,+0.07]\) across starts) --- the near-neutral
Proposition-7 degeneracy zone, where wavering is the predicted
behaviour. The \textbf{measured} side is more exposed: the published
measured column is a \emph{single-episode} protocol everywhere (the
\(100\)-start thickening samples more starts along the \emph{same}
episode), and re-sampling the true episode moves the three calibrated
walker ratios across \([0.16,0.94]\) / \([0.39,1.10]\) / \([0.19,1.02]\)
(\(5\)--\(6\) episode seeds per cell) --- the published
\(0.94/0.95/1.02\) sit in the upper tail. The mechanism is resolution,
not luck-free calibration: at \(\epsilon{=}0.2\) the certified horizon
is \(\approx6\) steps and measured medians are small integers, so one
step of median moves the ratio by \(\approx0.16\). No cell crosses
regimes under re-sampling --- the \textbf{taxonomy is episode-robust};
the \emph{band} statement should be read at episode resolution, and the
protocol upgrade going forward is pooling \(\ge5\) episodes and
reporting at finer \(\epsilon\).

\textbf{The full-zoo expansion: \(84\) cells, and the seed map's axis
does not survive it (Experiment 23b, \texttt{step89b}).} We then re-ran
the byte-identical protocol on \textbf{every remaining public
single-task dmcontrol checkpoint} in the official zoo. The universe
(\(34\) tasks) and an inclusion rule were frozen in a spec before any
cell ran, and the rule is \emph{self-executing}: a task enters iff stock
dm\_control \texttt{suite.load} accepts its name mapping --- the \(11\)
custom TD-MPC2 variants (cheetah-jump, run-front/back(wards), cup-spin,
hopper-hop-backwards, pendulum-spin, reacher-three-*,
walker-*-backwards) are excluded \emph{by the rule}, since the measured
cross-check column requires the true environment; the excluded list and
one engineering retry (fish-swim-3, a corrupt download re-pulled) are
ledgered in the artifact. Result: \(69/69\) new cells, \(84\) total.
\textbf{As frozen}: \(42/84\) abstain (stable \(13\) --- median censored
\(\ge10/20\) starts; bias \(29\)), \(42\) expansive with
measured/certified median \(0.50\) at \(\lambda_1\ge0.25\) and \(0.33\)
below it, in-band \([2/3,3/2]\) only \(11/42\). \textbf{The seed map's
``calibrated where strongly expansive'' is overturned}: in-band cells
span \(\lambda_1{=}0.01\)--\(0.39\) --- expansion strength does not
predict calibration. What predicts it is \emph{which quantity sets the
measured horizon}. Where the model's native one-step residual already
sits at the threshold (measured median \(\le3\) steps at
\(\epsilon{=}0.2\): \(25\) of \(42\) expansive cells --- the entire
dog/humanoid/quadruped block), the certificate prices a growth the
rollout never exhibits: the \textbf{same mechanism E16 found on V-JEPA
2-AC}, now measured in bulk --- Proposition 7's degeneracy direction as
the zoo's \emph{default regime}. Where the horizon is long enough to be
growth-set (median \(\ge5\): \(15\) cells), the certificate is
\textbf{calibrated}: ratio median \(0.94\), \(8/15\) in \([2/3,3/2]\),
across five domains including two with no seed-map relative (cup-catch
\(0.68/1.36/1.68\), cartpole-balance-sparse \(1.13\)); walker-run-2
lands at \(0.90\) in the \emph{proven-flattening} domain, so the low
ratios elsewhere are not a pipeline artifact (its own seeds 1/3 land
\(0.06/0.35\) --- certificate quality varies per (task, seed) world
model, not per domain). The re-stratification is descriptive, not
pre-registered; cut sensitivity is disclosed (growth-side median
\(0.53/0.68/0.90\) at cuts \(\le1/\le2/\le3\)) and the frozen quantities
--- protocol, fractions, \(\lambda\)-split medians --- are reported
above unchanged. All measured statistics are at \(100\) rollout starts
per cell (\texttt{step89c}, a \(5\times\) thickening of the published
\(20\); spec frozen first): the regime map is start-count-stable ---
bias \(24\to25\), growth set \(15\to15\) (one boundary swap), in-band
\(10/15\to8/15\) with cup-catch-2 exiting \emph{above} the band
(\(1.68\), the conservative side) --- and the growth-side median moves
\(0.95\to0.94\). The split is moreover \textbf{\(\epsilon\)-monotone in
the predicted direction}: recomputed at the tighter published
resolutions, bias-dominance grows \(25/42\to31/42\to39/42\) and the
growth set shrinks \(15\to7\to3\) (\(\epsilon=0.2\to0.1\to0.05\), same
\(100\)-start artifact) --- the threshold is
\(\epsilon\cdot\mathrm{scale}\), so tighter \(\epsilon\) hands more
horizons to the fixed native residual, which is precisely the seed map's
tight-\(\epsilon\) optimistic regime (Proposition 8's \(\delta\)-bias)
as a measured dose--response. The corrected reading is uniform from toy
to zoo to \(1\)B: \textbf{the spectrum prices error \emph{growth}; the
measured column tests error \emph{level}; the audit is their
conjunction.} \texttt{experiments/step89b},
\texttt{tests/test\_step89b.py}; spec
\texttt{docs/specs/2026-06-11-step89b-audit-expansion-seed.md}.

\textbf{Scale does not rescue trustworthiness (Experiment 25,
\texttt{step92}).} Across the official TD-MPC2 \emph{multitask} ladder
(mt30, \(1\)M\(\to\)\(317\)M parameters, same walker-walk task, one
official checkpoint per size), the policy-prior loop's regime flips
\textbf{non-monotonically} with scale --- contracting at \(1\)M
\emph{and} \(48\)M, expansive at \(5\)M/\(19\)M/\(317\)M --- and
calibration scatters (measured/certified \(0.39/1.91/1.22\) at
\(\epsilon{=}0.2\) where expansive --- \(100\)-start measured column,
start-count-stable vs \(0.37/1.87/1.16\) at the original \(12\); mt80
cells likewise mixed) with \textbf{no size matching the single-task
\(5\)M model's \(0.94\)--\(1.02\)}. One checkpoint per cell (no official
seed variants) --- read as a descriptive scope-map extension, not a
seed-averaged law; the direction is nonetheless unambiguous: \emph{trust
in a rollout is a property of the loop's dynamics, not of parameter
count --- scale buys interpolation, not a calibrated horizon.}
\textbf{Task-thickened (2026-07, \texttt{step92b}):} re-auditing the
ladder at \(5\) env-wired tasks \(\times\) all \(7\) checkpoints (\(35\)
loops) confirms and sharpens this --- \textbf{\(0/35\) cells} reach the
single-task band at \(\epsilon{=}0.2\) (closest: walker
\(1.16\)/\(1.87\) at \(317\)M/\(19\)M, the over-shoot side); per-task
regimes flip non-monotonically along the ladder; and at \emph{fixed}
\(5\)M parameters the training \textbf{suite} flips the regime census
wholesale --- mt30 \(4\) expansive/\(1\) straddle vs mt80 \(5/5\)
contracting: \emph{the suite, not the size, sets the loop's dynamics.}

\begin{figure}
\centering
\pandocbounded{\includegraphics[keepaspectratio,alt={Scale does not buy a calibrated horizon (Experiment 25). (a) \textbackslash lambda\_1 of the walker-walk policy-prior loop across the official multitask ladder: sign-flipping, non-monotone (contracting at 1M and 48M). (b) Calibration at \textbackslash epsilon\{=\}0.2: scatter across sizes; no multitask scale reaches the single-task 5M band (0.94--1.02, green).}]{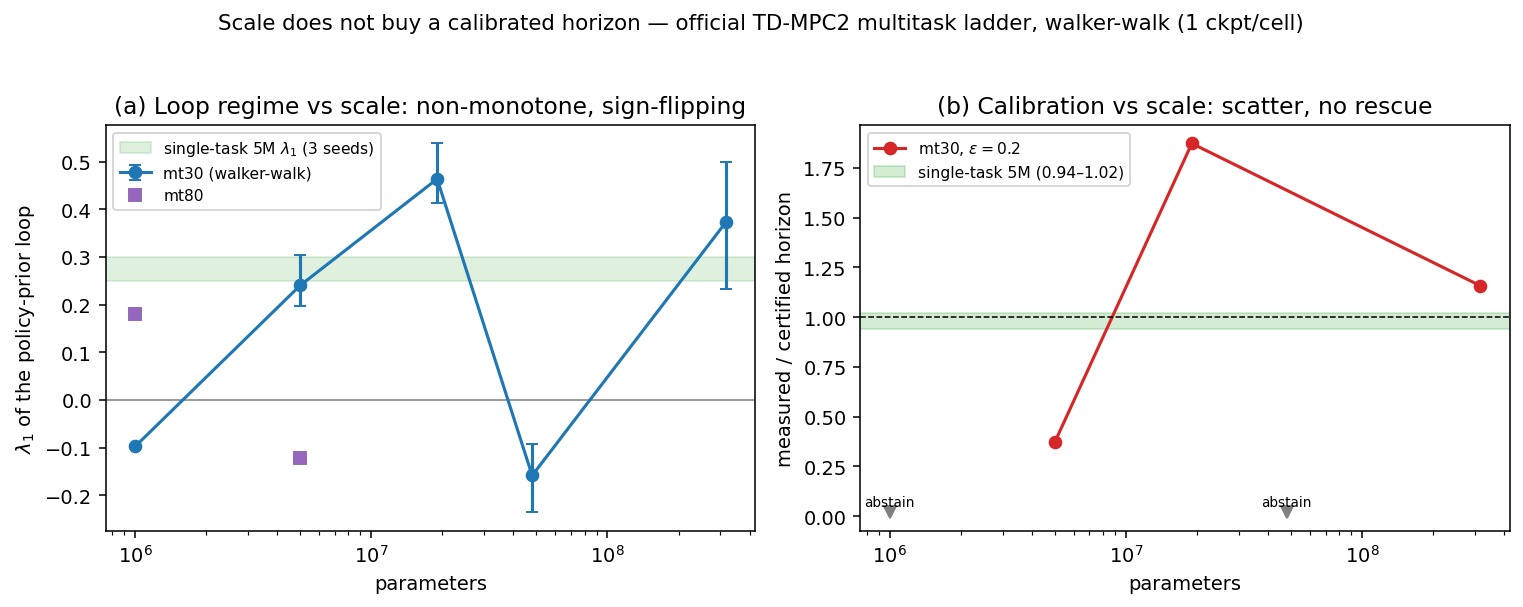}}
\caption{Scale does not buy a calibrated horizon (Experiment 25).
\textbf{(a)} \(\lambda_1\) of the walker-walk policy-prior loop across
the official multitask ladder: sign-flipping, non-monotone (contracting
at \(1\)M and \(48\)M). \textbf{(b)} Calibration at \(\epsilon{=}0.2\):
scatter across sizes; no multitask scale reaches the single-task \(5\)M
band (\(0.94\)--\(1.02\), green).}
\end{figure}

\textbf{The published certificate prices a deployed monitor,
out-of-sample (Experiment 26, \texttt{step94}).} The scope law's
\emph{positive} instance in deployment form --- and the experiment where
the audit's numbers are spent rather than admired. A \textbf{sensor-only
monitor} watches cheetah-run executing its nominal policy (the
deterministic prior on true observations --- monitoring cannot perturb
the system, designing out the §5.19/step93 control confound); the
expensive sensor is read every \(k\) steps; between reads the monitor
forecasts the latent with the certified loop
\(g(z)=d(z,\tanh\mu_\pi(z))\) itself --- \textbf{no action telemetry}
--- and at a read it flags iff relative latent error exceeds
\(\theta{=}0.2\), then resyncs. The certificate numbers are
\textbf{loaded from the Experiment 23 artifact} (issued before this
experiment existed: a-priori in the strongest sense), and the gates were
frozen before official seeds 1--2 were ever run --- those two cells are
genuinely out-of-sample. Results, cell-by-cell against the published
bench map: in-situ staleness ratio
\(s^{\ast}_{\mathrm{med}}/T_1^{\mathrm{pub}}(0.2)=0.43\) vs bench
\(0.43\) (seed 1), \(0.50\) vs \(0.50\) (seed 2) --- \textbf{two-decimal
agreement on the out-of-sample optimistic cells, optimism replicated
where optimism was published} --- and \(0.67\) vs \(0.83\) on the
calibrated cell, whose stricter \([2/3,3/2]\) check lands
\(7{\times}10^{-4}\) below the band edge on an integer-valued median
(crossings are whole steps; the edge sits at \(4.003\)): recorded
\textbf{at-the-edge, not rounded up}. A frozen-actuator fault (executes
\(0\) while the nominal forecast is unchanged) is detected at the
certificate-derived cadence
\(k_{\mathrm{op}}=\max(2,\mathrm{round}(T_1^{\mathrm{pub}}/3))\) with
recall \(1.00\) on all three seeds and, at \(n{=}100\) episodes
(\(1600\) windows/seed), median delay \(\le k_{\mathrm{op}}\) on \(3/3\)
--- the cell that missed by one read at \(n{=}20\)
(\(k_{\mathrm{op}}{=}2\), delay \(3.0\); the frozen channel needs
\(\sim2\)--\(3\) steps to push relative latent error past \(\theta\))
lands exactly at \(k_{\mathrm{op}}\) (delay \(2.0\)), with all in-situ
ratios replicating \(n{=}20\) digit-for-digit (\(0.43/0.50/0.67\)), the
at-the-edge cell included. Proposition 11 is the formal frame: clause
(i) --- the decided quantity here \emph{is} the certified quantity, so
certificate value transfers with \textbf{zero new estimation}; clause
(ii) prices §5.19/step93's dilution as a resolution mismatch
(\(H(\theta^{\ast})\approx2\) vs \(H(0.2)\approx6\) agent-steps). The
design trail is disclosed in the script header and is itself
taxonomy-obedient: the teacher-forced loop variant \(z\mapsto d(z,a_t)\)
lands weakly expansive (\(\lambda_1{=}0.09\)--\(0.13\)) and its
certificate would be optimistic exactly as the scope map predicts; and
the same monitor on walker-walk is \textbf{regime-bimodal} (the
deterministic prior falls on some env seeds; belief-invalidity
\(0.24\to0.66\) tracks torso height \(1.05\to0.68\)) --- a monitor
presumes a nominal regime, Proposition 7's scope clause in deployment
guise, \textbf{and the clause is load-bearing}: the walker secondary
run's in-situ ratios land \(0.37\)--\(0.63\) against bench
\(0.94\)--\(1.02\) (\(0/3\) replication at \(n{=}100\) episodes,
\(1600\) windows/seed; at \(n{=}20\): \(0.32\)--\(0.47\)) --- and the
bimodality now has a quantitative signature: \textbf{all three seeds pin
the lower quartile at \(1\) window} (q25 \(=1\), medians \(2\)--\(4\))
--- a mass of instant crossings from the fallen regime dragging the
clock, while fault recall stays \(0.9\)--\(1.0\). Regime contamination
breaks clock replication while detection survives
(\texttt{step94\_budgeted\_monitor\_walker-walk.json}).
\texttt{experiments/step94\_budgeted\_monitor.py}; gates G1a
(cell-by-cell replication, \(|\Delta r|\le0.25\)): \textbf{3/3 PASS};
G1b (calibrated-cell band): at-the-edge INCONCLUSIVE; G2 (detection):
\textbf{3/3 PASS} at \(n{=}100\) (per the original pre-registration:
\(2/3\) at \(n{=}20\)); the walker arm is observational per the spec, no
gate binds on it.

\textbf{The abstain cells deploy as predicted --- the taxonomy completes
in deployment (Experiment 28, \texttt{step96}).} The published scope
map's remaining cell types, run through the same sensor-only monitor
with gates frozen a-priori: \textbf{stable-abstain} (finger-spin-2/3,
\(\lambda_1<0\), bench \(15\)--\(19/20\) starts censored --- the
certificate issued no horizon because nothing diverges) deploys as
\textbf{free monitoring}: \(92\)--\(94\%\) of \(k{=}24\) windows never
cross \(\theta\), belief-invalid fraction \(4.0\)/\(5.6\%\) at
\(n{=}100\) (vs \(\sim50\%\) at HALF that cadence on cheetah),
frozen-actuator recall \(1.00\) with median delay \(5\) at
\(k_{\mathrm{op}}{=}8\) --- zero-false-alarm monitoring at arbitrary
cadence with detection intact. The pre-registered G3 gate (\(\le5\%\)
invalid, \(2/2\) cells) goes \(1/2\) at \(n{=}100\): finger-spin-2 lands
\(0.6\)pp over the line --- recorded \textbf{as-registered, INCONCLUSIVE
not PASS} (\(4.4\%\) at \(n{=}20\) was a boundary cell revealed by
thickening; the qualitative contrast is intact). \textbf{Bias-abstain}
(hopper-hop-1, \(\lambda_1=-0.105\) yet bench median crossing \(5.5\)
with zero censoring --- divergence is bias, not amplification) deploys
with the same fast clock the certificate \emph{refused to price}:
in-situ median \(8.0\) at \(n{=}100\) (q25--q75 \(5\)--\(11\)), inside
the pre-registered \(\times1.5\) band of bench (G4 PASS); any Lyapunov
price here would be \(\infty\) --- the abstain is the correct verdict,
now confirmed where it costs something. A further replication cell
(finger-spin-1, bench ratio \(0.95\)) lands at \(1.04\) (G5 PASS). With
Experiment 26's cells, \textbf{every cell type of the published taxonomy
--- calibrated, optimistic, stable-abstain, bias-abstain, plus the
walker regime caveat --- now has a deployment instance predicted from
the published artifact}.
\texttt{experiments/step96\_taxonomy\_monitor.py}.

\textbf{The pixel family closes the deployment ordering (Experiment 29,
\texttt{step97}).} The same sensor-only monitor on the architecturally
disjoint \textbf{LeWM/PushT} cell --- the published
\textbf{bias-abstain} verdict (\(\lambda_1{=}0.0013\), CI straddling
\(0\); one-step relative bias \(0.17\); bench crossing \(2\) model
steps) --- with gates frozen a-priori: \textbf{G6 PASS} (no usable
cadence: belief-invalid fraction \(0.50/0.75/0.87\) at \(k{=}2/4/8\) ---
exactly \((k{-}1)/k\), the staleness-\(2\) crossing firing in every
window) and \textbf{G7 PASS} (flooded alarm channel: per-read flag rate
\(1.00\) at every \(k\ge2\); the telemetry-corruption arm is inseparable
from drift, pre/post flag rate \(1.00/1.00\) --- exactly as stated
before the run). At \(k{=}1\) the monitor is clean (\(0.00\)) --- but
\(k{=}1\) means reading every frame: the forecaster buys \textbf{zero
sensing savings}, and a smoke-caught v1\(\to\)v2 gate correction
(disclosed in the script header) records why \(k{=}1\) tests the
published one-step bias, not monitor usability. A moving-scene control
(random-walk actions, telemetry true) reproduces the verdict (\(0.76\)
at \(k{=}4\)). \textbf{The taxonomy now \emph{orders} deployment value
a-priori: stable-abstain \(=\) free monitoring (Experiment 28) \(>\)
calibrated/optimistic \(=\) priced savings (Experiments 26/28) \(>\)
bias-abstain \(=\) do not deploy (here)} --- three verdicts, three
architectures, one training-free candidate read-out plus cross-check.
\texttt{experiments/step97\_lewm\_monitor.py}.

\textbf{At foundation-model scale the cross-check column is load-bearing
(Experiments 30--31, \texttt{step98}--\texttt{step99}).} The third
architecture family: \textbf{V-JEPA 2-AC} --- Meta's \(1\)B-encoder
action-conditioned world model (official
\texttt{vjepa2\_ac\_vit\_giant}, ViT-g/16 \(1012\)M + AC predictor, MIT,
post-trained on DROID), loaded via torch.hub into the authors' own code,
audited along the authors' own energy-landscape rollout: per-frame token
blocks (\(256\times1408\), \(d{=}360{,}448\) --- \(600\times\) the
largest previous loop), fixed zero-delta action (the LeWM pre-registered
scope). Their planning \emph{energy} is the L1 distance between
predicted and encoded tokens --- the very quantity our monitor
thresholds: the certificate prices the growth of V-JEPA 2-AC's own
energy. \textbf{Certificate (Experiment 30):} leading-\(6\)
JVP-Benettin, fp32 CUDA (disclosed), two independent \(Q\)-seeds
agreeing to \(1.8\%\): \(\lambda_1=0.176\)--\(0.180\) across
\textbf{five} \(Q\)-seeds (\(2.4\%\) spread; 2026-06-11 thickening from
two), envelope CI \([0.133,0.250]\) --- \textbf{expansive}, nominal
\(T_1(0.2)=9.1\) model steps. \textbf{Measured side on real robot data
(Experiment 31):} \(40\) real DROID episodes
(\texttt{lerobot/droid\_100}, exterior camera, \(4\)-frame model step
per the official config, telemetry actions via the authors'
\texttt{poses\_to\_diff}) --- the deployment error \emph{starts} at the
representation's native step-motion scale: one-step median \(0.629\) vs
consecutive-latent distance \(0.680\) (the predictor beats the
copy-of-last-read baseline, \(0.629\) vs \(0.742\), its value being
energy \emph{ordering} for CEM rather than metric forecasting --- the
franka energy-landscape check orders random \(0.471\gg\) true
\(0.426\)); staleness error grows \(0.623\to0.774\) over \(8\) steps,
log-slope \(0.030\ (\text{per-window median }0.027)\) ---
\(5\)--\(6\times\) below the certified \(\lambda_1\); belief-invalid
fraction \(1.00\) at every cadence including \(k{=}1\). The
pre-registered pricing branch \textbf{G8-E fails as registered} and the
pre-registered sub-classification rule reads \textbf{bias}: the error
never enters the linearization neighborhood, so the tangent spectrum's
jurisdiction and the monitor's operating point do not overlap ---
\textbf{Proposition 7's bias degeneracy at \(1\)B scale}. The
consequence is the experiment's headline: a spectrum-only audit would
have over-promised \(T_1{\approx}9\) on a flagship foundation world
model; the E13 protocol's measured cross-check \textbf{overrides the
tangent number}, and the cross-validated audit --- not the raw spectrum
--- is the deployable object. Two transferable lessons, stated plainly:
\emph{thresholds are representation-relative} (this token geometry moves
\(\sim0.68\) per step natively; the DMC convention \(\theta{=}0.2\) does
not transfer), and \emph{rates price only errors inside the
linearization neighborhood}. Wording红线 honored throughout: real-robot
\textbf{data}, offline monitoring (the monitor is passive --- replaying
logged episodes is a faithful instantiation). Engineering disclosures in
the script headers: upstream's \texttt{localhost} URL goof
(monkeypatched), the (current, goal) semantics of the repo's franka
pair, four layers of forward-AD\(\times\)SDPA incompatibility ending in
the authors' own explicit-attention branch + an explicit math-SDPA, and
a reverse-graph-retention OOM (parameters now frozen).
\texttt{experiments/step98\_vjepa2\_audit.py},
\texttt{experiments/step99\_droid\_monitor.py}; conditional-gate spec
(\texttt{docs/specs/2026-06-10-step99-droid-monitor-seed.md}) frozen
before the certificate was read.

\textbf{What does structure buy, if the read-out audits any smooth
model?} Theorem B's spectral law applies to any \(C^1\) latent map ---
that is what §5.21 (Experiment 23) exercises on (non-equivariant)
TD-MPC2 and LeWM. What the law cannot supply for a generic model is
\emph{trust in the number itself}: a dense model's spectrum can be
silently wrong while its predictions stay good (§5.16: \(\lambda_1\)
inflated \(\sim3.4\times\) at one-step relMSE \(10^{-5}\)), so a generic
certificate must be cross-validated against held-out divergence ---
exactly the per-model empirical check the §5.21 (Experiment 23) scope
map performs. Equivariance is what removes that requirement where it
holds: structure makes the spectrum \emph{faithful} (§5.16), hence the
certificate \emph{a-priori trustworthy with zero calibration data}
(§5.20 and its recalibration control) --- and exclusively so (Lemma 2).
The audit is universal; the \textbf{a-priori} guarantee is structure's.

\begin{center}\rule{0.5\linewidth}{0.5pt}\end{center}

\subsection{6. Related Work}\label{related-work}

This paper sits among a cluster of concurrent
\emph{structure-for-prediction} works. Prior work supplies
\emph{mechanisms} (equivariant predictors), \emph{priors} (latent
geometry), or \emph{diagnostics} (oracle-bypass) --- each empirical or
distributional and on a single axis; this paper supplies a
\emph{provable, computable region} across configuration \(\times\)
horizon \(\times\) resolution, with the Noether hinge (§4) tying the
axes together. The table below places each work against the
certificate's three axes and the \emph{kind} of guarantee it offers (✓
provides; \textasciitilde{} partial/empirical; --- not addressed); the
prose elaborates after it.

Several 2026 instruments are adjacent to the audit. Reconstruction
fidelity is measured orthogonal to action recoverability across encoder
families (PoR, arXiv:2606.07687 --- Dreamer-4 action \(R^2\approx0\)
across three configurations), corroborating the audit's premise that
average fidelity hides control-relevant failure; action-semantics
consistency probes (ATM, arXiv:2606.09028 --- simulation-free, seconds
per checkpoint) compose with the horizon certificate as an orthogonal
audit axis; WEAVER's three world-model desiderata --- fidelity,
consistency, efficiency (arXiv:2606.13672) --- name the fourth this
paper supplies: \emph{certifiability}; and the explicit/implicit
world-model dichotomy of the recent tutorial (arXiv:2606.12783) places
the certified latent loop squarely across its seam --- explicit enough
to audit, implicit enough to scale.

{\def\LTcaptype{none} 
\begin{longtable}[]{@{}
  >{\raggedright\arraybackslash}p{(\linewidth - 10\tabcolsep) * \real{0.1429}}
  >{\raggedright\arraybackslash}p{(\linewidth - 10\tabcolsep) * \real{0.1429}}
  >{\centering\arraybackslash}p{(\linewidth - 10\tabcolsep) * \real{0.1905}}
  >{\centering\arraybackslash}p{(\linewidth - 10\tabcolsep) * \real{0.1905}}
  >{\centering\arraybackslash}p{(\linewidth - 10\tabcolsep) * \real{0.1905}}
  >{\raggedright\arraybackslash}p{(\linewidth - 10\tabcolsep) * \real{0.1429}}@{}}
\toprule\noalign{}
\begin{minipage}[b]{\linewidth}\raggedright
Work
\end{minipage} & \begin{minipage}[b]{\linewidth}\raggedright
Core mechanism
\end{minipage} & \begin{minipage}[b]{\linewidth}\centering
Config. \(\langle S\rangle\)
\end{minipage} & \begin{minipage}[b]{\linewidth}\centering
Horizon\(\times\epsilon\)
\end{minipage} & \begin{minipage}[b]{\linewidth}\centering
Closed-loop
\end{minipage} & \begin{minipage}[b]{\linewidth}\raggedright
Guarantee kind
\end{minipage} \\
\midrule\noalign{}
\endhead
\bottomrule\noalign{}
\endlastfoot
BRo-JEPA & \(\mathbb{Z}/10\) cyclic predictor & \textasciitilde{} one
group & --- & --- & empirical zero-shot \\
UWM-JEPA & \(U(d)\) unitary predictor & \textasciitilde{} one group &
--- & --- & empirical \\
UR-JEPA / LeJEPA & latent-geometry prior (an/isotropy) & --- & --- & ---
& distributional (2nd-order) \\
IMWM & oracle-bypass; residual \(=\) search & --- & --- &
\textasciitilde{} search budget & diagnostic \\
LDA & \(\mathrm{SE}(3)\)-intrinsic diffusion (action side) & --- & --- &
\textasciitilde{} policy & empirical, modest\(+\)robust \\
Symmetry--Data rate & aug\(+\)TTA \(=\) equivariance; wrong-group
harmful & \textasciitilde{} one group & --- & --- & empirical scaling
law \\
companion line & exact flatness: \(1\) element, \(1\) resolution &
\textasciitilde{} one element & --- & ✓ invariance & proved (a
corner) \\
certified equivariance (robustness / conformal) & orbit-margin /
group-averaged conformal score & \textasciitilde{} one group & ---
(single-shot) & --- & proved, but \textbf{single-shot}, forward-only \\
Noether Networks / Razor & meta-learn conserved quantities & --- &
\textasciitilde{} better avg prediction & --- & average accuracy, no
guarantee \\
\textbf{this paper} & equivariant \textbf{predictability certificate} &
✓ \(k\!\Rightarrow\!\langle S\rangle\) (Thm A, Lem 1) & ✓
\(T_j(\epsilon)\!\sim\!\tfrac{\log(1/\epsilon)}{\lambda_j}\),
\textbf{tight} (Thm B, Prop 6) & ✓ exact (Exp 11) & \textbf{a-priori,
per-situation, computable (Alg 1)} \\
\end{longtable}
}

Each row makes the same point concretely: the prior work's guarantee is
empirical or distributional and lives on a single axis, whereas the
certificate is a-priori, per-situation, and spans all three. The
detailed comparisons below work through the table row by row.

\textbf{Equivariance and constant error (our companion line).} A
companion paper establishes, for a single group element and one
resolution, that exact equivariance makes one-step error constant across
the group, with exact-flatness and closed-loop-invariance corollaries.
The present paper lifts that corner along three axes: Theorem A is its
multi-step, full-monoid generalization; Theorem B adds the horizon
\(\times\) resolution stratification; and Lemma 1 turns \(k\) generator
checks into an exponential certified set.

\textbf{Certified equivariance is single-shot; ours is multi-step, with
a converse.} The closest \emph{guarantee-bearing} relatives live in
robustness and conformal prediction, not world models, and a careful
comparison sharpens exactly what is new here. Equivariant classifiers
have \emph{orbit-constant margins} --- the decision-boundary gradient
norm is preserved across the orbit, giving uniform adversarial
certificates (arXiv:2510.16171); invariance-aware randomized smoothing
builds \emph{orbit-based} certificates (arXiv:2211.14207); and
Equivariantized Conformal Prediction (eCP, arXiv:2602.03986)
\emph{group-averages the nonconformity score} --- frame averaging for
distribution-free coverage --- to tighten prediction sets over an orbit.
Three differences are load-bearing. \textbf{(i) Time.} All of these are
\emph{single-shot}: a classifier margin or a one-shot prediction set,
with the guarantee attached to one input, never \emph{propagated through
an \(H\)-step latent rollout}; none has a prediction-\emph{horizon}
axis, a predictor \emph{spectrum}, or a \emph{conservation} law. Our
certificate is precisely a multi-step statement, stratified by the
Lyapunov spectrum (Theorem B, made \emph{tight} by Proposition 6) and
extended to long horizons on conserved channels (Proposition 5; to all
horizons only at exact conservation \(\eta{=}0\)). \textbf{(ii)
Direction.} This literature proves \emph{equivariance \(\Rightarrow\)
orbit-constant guarantee} --- at root the classical fact that an
equivariant estimator's risk is constant on orbits, specialized to
margins or coverage; it proves no converse. Lemma 2 supplies the
\textbf{converse} (orbit-constant error \(\iff\) equivariance), which is
what makes ``an unconstrained architecture cannot certify it by
construction'' a theorem rather than an observation. \textbf{(iii)
Object.} Frame averaging unifies the two literatures concretely but at
different layers: eCP averages a \emph{score} post-hoc for coverage; we
average an \emph{encoder/predictor} (Experiment 13) for an in-model
multi-step certificate --- the same Reynolds operator, opposite ends of
the pipeline. Separately, \textbf{learned-conservation} methods ---
Noether Networks (Alet et al., 2021; arXiv:2112.03321) and Noether's
Razor (2024; arXiv:2410.08087) --- meta-learn conserved quantities to
\emph{improve average prediction} by shrinking the hypothesis space; our
Noether hinge instead \emph{certifies} (Proposition 5 turns a conserved
charge into an a-priori long-horizon guarantee, Proposition 4 says which
isotypic block must carry it) --- guarantee versus average accuracy.
Finally, \textbf{Jacobian-regularized world models} (arXiv:2501.00195)
damp rollout error propagation by penalizing the transition Jacobian ---
a heuristic; Theorem B is the \emph{provable} version of the same
intuition, reading a per-channel certified horizon
\(T_j(\epsilon)\sim\log(1/\epsilon)/\lambda_j\) off the spectrum rather
than regularizing toward stability, and (Proposition 6) characterizing
how that horizon shrinks when the symmetry is only approximate.

\textbf{Equivariant predictors.} Block-rotation predictors (BRo-JEPA,
arXiv:2606.01372) match a \(\mathbb{Z}/10\mathbb{Z}\) cyclic structure
to the latent --- ``add \(k\)'' becomes ``rotate \(k\theta\)'' --- and
report \(99.46\%\) best-config zero-shot transfer on MNIST modular
arithmetic; unitary-predictor world models (UWM-JEPA, arXiv:2605.25313)
do the analogous thing with \(U(d)\). Theorem A explains \emph{why} such
zero-shot transfer occurs (the representation cancels inside the norm
--- orthogonal for the former, unitary for the latter), and the closure
lemma quantifies \emph{how far} it reaches (the whole generated monoid
from its generators). These works share our toy-construction caveat, and
we cite them as cross-group mechanism corroboration rather than scale
results.

\textbf{Latent-geometry regularizers.} LeJEPA (Balestriero and LeCun,
2025) pushes the latent toward an isotropic Gaussian; UR-JEPA (Le, 2026;
arXiv:2606.01443) challenges that target, arguing isotropy conflicts
with the manifold hypothesis and that a data-discovered anisotropy is
preferable. Our latent's anisotropy is neither isotropic nor
data-discovered but \textbf{group-prescribed}: the covariance is pinned
to the representation, \(\rho\Sigma\rho^\top=\Sigma\) (measured residual
\(3\times10^{-4}\), versus \(1.04\) for a non-equivariant baseline). In
a head-to-head (companion line, single seed) the data-fit anisotropy is
best \emph{in-distribution}, but only the group-prescribed anisotropy
transfers out of distribution (a \(\sim320\times\) gap, consistent with
the three-seed \(68\text{–}320\times\) and \(10\text{–}155\times\) of
Experiments 8 and 7). The certificate is therefore complementary --- a
first-order, per-situation guarantee from the group, not a second-order
distributional prior --- and it predicts precisely \emph{when} a
discovered anisotropy will fail to generalize (off the training orbit).

\textbf{Oracle-bypass diagnostics.} IMWM (Gao et al., 2026;
arXiv:2606.01626) uses the same falsifiable move as our capacity-ladder
experiments --- replace the learned model with oracle dynamics and see
what is still missing --- and finds a finite-budget planner \emph{still}
fails. It attributes the residual to \textbf{search} (proposal-sampling
volume), whereas our analogous diagnosis attributes it to
\textbf{representation} (permutation-invariant encoder pooling). These
are complementary bottlenecks: a predictability certificate bounds the
\emph{model's} error over \(\langle S\rangle\times T\times\epsilon\),
and makes no claim about the planner's search budget downstream.

\textbf{Geometry on the action side.} LDA (``The Lie We Tell'', Chuang
et al., 2026; arXiv:2606.01847) names the \emph{Euclidean Fallacy} ---
flattening an \(\mathrm{SE}(3)\) pose into \(\mathbb{R}^{12}\) breaks
the manifold constraint, the coordinate-change equivariance, and
geodesic optimality --- and fixes it by running diffusion \emph{on}
\(\mathrm{SE}(3)\) (tangent-space score, exponential-map retraction). It
states our core motivation --- \emph{do not flatten geometric
quantities} --- on the action side. As a diffusion policy its gains are
modest but robust (CALVIN average task length \(3.27\to3.51\),
\(+7.3\%\)), the same profile as equivariant methods generally; our
framing explains why a geometric prior buys a \emph{kind} of guarantee
(consistency across the group, robustness out of distribution) rather
than a uniform accuracy jump.

\textbf{Concurrent corroboration: the symmetry--data exchange rate, and
learned irreps.} Two concurrent works sharpen pieces of our picture.
\emph{Measuring the Symmetry--Data Exchange Rate} (Adly, 2026;
arXiv:2606.01090), on a controlled \(C_n\) task, independently reports
the two facts our §5.8 and §5.5 rest on: (i) a non-equivariant model
with augmentation \textbf{plus test-time orbit averaging matches the
equivariant model essentially exactly} (coincident per-epoch validation
curves) --- the single-orbit tie of our augmentation experiment
(Experiment 10); and (ii) a \textbf{misaligned (wrong-group) symmetry
constraint is \emph{actively harmful}} (\(95\%\) CI excluding zero), not
merely unhelpful --- the approximate-symmetry threshold of our §5.5. It
frames the trade as a ``symmetry--data exchange rate,'' the average-case
dual of our worst-case separation (§3.3: structure certifies the
\(\epsilon\)-independent orbit, data only an \(\epsilon/L\) tube); like
ours it carries the controlled-toy caveat, and it stops short of the
multi-axis certificate, the spectral horizon law, the closed-loop
clause, and the Noether hinge. On the learning-dynamics side,
\emph{Neural Networks Provably Learn Spectral Representations for Group
Composition} (He et al., 2026; arXiv:2606.02993) proves that gradient
flow on a group-composition task drives each neuron to a single
\textbf{irreducible representation} --- a provable account of \emph{how}
the isotypic structure our Proposition 4 places, and our §5.6 discovery
experiments recover, emerges under training.

\textbf{Predictability horizons.} The
\(T(\epsilon)\sim\log(1/\epsilon)/\lambda\) law is classical for
dynamical systems (Lyapunov; numerical weather prediction), and that the
\emph{local} spectrum governs the \emph{asymptotic} rate is the content
of the Oseledets multiplicative ergodic theorem; on \emph{uniformly}
hyperbolic systems the shadowing lemma (Anosov--Bowen; Pilyugin) further
bounds the forecast-horizon \emph{floor} of a perturbed (learned) model
--- though it controls orbit error, not the exponent, and does not
formally cover singular-hyperbolic Lorenz (Tucker 2002). Our
contribution is to (i) measure the law on a \emph{learned latent world
model}, including a learned model of genuinely chaotic dynamics (Lorenz,
Experiment 14: the learned model's Lyapunov exponent, read as the
staircase slope, \emph{matches} the true \(\lambda_1\) to
\(1\text{–}8\%\) --- an empirical lift, since shadowing does not
transfer exponents); (ii) characterize \emph{when} the local spectrum
certifies the horizon --- Proposition 7's dichotomy: informative on
spectrally non-degenerate dynamics (\(\lambda_1>0\)), vacuous on
near-neutral dynamics (\(\lambda_1\approx0\), the PushT interior) ---
synthesizing Oseledets (for the rate) and shadowing (for the floor) with
the certificate; (iii) tie its slow subspace to the group-invariant
subspace via the Noether hinge; and fold all of it into a single
multi-axis certificate for equivariant models.

\textbf{Lyapunov spectra from learned models.} That a \emph{learned}
(not numerically-integrated) model recovers a chaotic system's Lyapunov
spectrum is established prior art: reservoir computers recover full
spectra and tie forecast skill to \(\sim 5\)--8 Lyapunov times (Pathak
et al.~2017, 2018), RNN-BPTT does the same (Vlachas et al.~2020), and
both extend to high-dimensional flows (Kobayashi et al.~2024); faithful
recovery from a \emph{recurrent} model is \emph{conditional} on the
learned model's own contracting modes --- the conditional-Lyapunov
criterion (Hart 2024). We therefore do \emph{not} claim
spectrum-recovery-from-a-learned-model as novel. Our contribution is to
make it \textbf{two-sided/tight, per-channel, and certified} for an
\emph{equivariant latent world model}, and (Experiment 18) to identify
that conditional-Lyapunov burden as exactly what sinks the unstructured
\emph{recurrent} route at high \(N\) where a Markov equivariant model
--- whose Jacobian is \emph{exactly} \(N\times N\) --- succeeds.

\textbf{Concurrent certified and equivariant neighbours.} TrustKoopman
(Conradie et al.~2026) derives certified multi-step error bounds for
data-driven Koopman/Perron--Frobenius forecasts --- the closest
\emph{certified-rollout} neighbour --- but over classical EDMD with
\textbf{no equivariance} (its ``invariance'' is Koopman-subspace, not
group), one-sided rather than two-sided/tight, and with no conservation
axis. Flow-Equivariant World Models (Lillemark et al.~2026) build an
equivariant world model for stable long-horizon prediction, but evaluate
it perceptually with \textbf{no certificate, spectrum, or
conserved-channel} statement. Symmetry-Protected Lyapunov Neutral Modes
(Mo 2026) proves an equivariant flow has \(\ge\dim(G/H)\) zero exponents
tangent to the group orbit --- adjacent to our Noether hinge
(Propositions 4--5) --- but as equivariant-RNN expressivity theory,
without the certificate, the two-sided horizon, or the
representation-theoretic charge placement. More broadly: model-free
equivariant RL gives orbit-constant \emph{value} functions (Wang et
al.~2022) (no learned forward model or horizon); group-structured latent
world models (Delliaux et al.~2025) are empirical (no certificate); and
equivariant representation learning with guarantees (Ordóñez-Apraez et
al.~2025) bounds a \emph{static} conditional-expectation operator's
spectrum, not a multi-step dynamics horizon. The equivariance \(\times\)
certified-horizon intersection, and the structure-beats-scale high-\(N\)
spectrum recovery, remain ours.

\begin{center}\rule{0.5\linewidth}{0.5pt}\end{center}

\textbf{Concurrent and recent work (swept through 2026-06).} Mo
(arXiv:2605.03338) proves symmetry-protected \emph{neutral} Lyapunov
modes for continuously-equivariant fields (\(\ge\dim(G/H)\) zero
exponents along the group orbit) --- complementary to ours: it
constrains the spectrum's \emph{kernel}, while we certify the
\emph{horizon} stratified by the whole spectrum (for our discrete
\(\mathbb{Z}_N\) systems \(\dim(G/H){=}0\), so their lower bound is zero
there --- the two results constrain disjoint parts of the spectrum: they
the kernel, we the horizon). Geng et al.~(arXiv:2512.08991) bound
world-model rollout deviation \emph{conformally} for closed-loop
verification --- statistical and rollout-hungry where ours is a-priori
and training-free; the same trade-off separates us from
reachability-based MBRL safety (UPSi, arXiv:2604.26836) and
where-to-trust heuristics (arXiv:2606.01363). Pretrained world models
have been probed \emph{semantically} (arXiv:2603.21546) and benchmarked
by sample rollouts (WorldBench, arXiv:2601.21282; WorldArena,
arXiv:2602.08971); a Jacobian certificate of a public model's latent
map, cross-validated against true-environment divergence, is to our
knowledge new --- the nearest priors being latent-space stability
analyses of \emph{self-trained} autoencoders (Özalp \& Magri,
arXiv:2410.00480) and Lyapunov-regularized DreamerV3 \emph{policies}
(arXiv:2410.10674). Flow-equivariant world models (Lillemark et al.,
arXiv:2601.01075, now ICML 2026; antecedent FERNN, arXiv:2507.14793)
preserve equivariance over arbitrarily long rollouts --- an
exactness/closure property, not a quantitative horizon. Inductive-bias
studies (arXiv:2602.06923) and position papers calling for verified
world models (arXiv:2602.23997) support the framing; PDEder
(arXiv:2603.22655) \emph{suppresses} latent Lyapunov exponents where we
\emph{certify} them; two-sided calibration bounds under equivariance
(arXiv:2510.21691) concern calibration error, not horizon; and the
data-assimilation literature's observation-frequency-vs-Lyapunov-time
rule (e.g.~Bocquet et al.~2026) is the classical neighbor of our
sensing-budget law (Proposition 9; §5.20).

\subsection{7. Limitations}\label{limitations}

\begin{itemize}
\item
  \textbf{Scale and scope.} All experiments are CPU/\(1\)-GPU.
  Experiments 9 and 11 validate the central claim on a real
  physics-engine contact simulator (PushT) --- dynamics we did not
  author --- at the \emph{prediction} level (Experiment 9) and the
  closed-loop \emph{task} level (Experiment 11, where the certificate
  becomes orbit-invariant pose control: out of the training wedge the
  equivariant controller stays flat while a scaled baseline degrades,
  the two being \emph{comparable} in-distribution --- we claim no
  in-distribution win), and Experiment 16 reproduces the
  prediction-level certificate on the \textbf{standard third-party
  MuJoCo FetchPush benchmark} with a \(3\)D arm. This is still
  \emph{structured state} (not pixels) and a single \(\mathrm{SO}(2)\)
  group.

  The magnitude of the out-of-distribution \emph{prediction} gap is
  \textbf{benchmark-dependent} --- \emph{modest} on PushT
  (\(2\text{–}4\times\) at \(H{=}10\), large only because a single PushT
  step is easy to predict in-distribution) but \emph{orders of
  magnitude} on the \(25\)-D FetchPush state (\(211\)--\(1445\times\)
  one-step, where the baseline extrapolates a higher-dimensional state
  badly) --- while the equivariant ratio is exactly \(1.000\) on both.
  We still claim \textbf{no in-distribution win} on either: FetchPush's
  in-dist tax \emph{straddles} \(1\) (the baseline interpolates the
  training orientation competitively-to-better, equivariant cheaper on
  one seed). The closed-loop \emph{out-of-wedge} advantage is shown on
  the contact-dominated pose task specifically --- a position-only push
  stays a tie, since a near-linear agent subsystem carries it out of
  distribution. The FetchPush closed-loop (Experiment 17) is at present
  the \textbf{model-rollout} planning certificate --- we prove the whole
  planner \(\mathrm{SO}(2)\)-equivariant and show its learned plan is
  orbit-flat (ratio \(1.000\)) while the baseline planner degrades
  \(4\)--\(10\times\).

  The \emph{real-env} \texttt{is\_success} task win we \textbf{ran twice
  (random and goal-directed data) and report INCONCLUSIVE}, with a
  diagnosis: the learned WM\(+\)CEM controller stays at the
  \(\approx7\%\) give-away floor for both stacks \emph{regardless of
  data}, even though the scripted pusher that generated the
  goal-directed data itself solves the task \(33\%\). The model is not
  the problem --- its goal-readout decodes the object position to
  \(<\!1\) cm --- but CEM gets \(0\%\) at every horizon because it
  \textbf{exploits the model off-distribution} (the classic
  model-based-RL pitfall). This is architecture-agnostic and unrelated
  to equivariance --- the certificate holds exactly; it is the embodied
  analogue of the pixel result (the prior is free, absolute closed-loop
  competence is the open part). The standard fix (a \(5\)-model
  equivariant ensemble with a CEM disagreement penalty, step75)
  \textbf{also} stays at the give-away floor --- so a real task win
  needs more than the obvious robust-MBRL patch: more compute, or a
  learned policy / offline RL rather than CEM-MPC. We do not claim a
  task win we did not measure.
\item
  \textbf{The certificate's demonstrated downstream value is
  efficiency-under-budget, not safety.} Experiment 22's win is a
  \emph{within-method, fixed-budget} re-observation efficiency contrast
  (the certificate's faithfulness, not the forecaster, is load-bearing).
  We separately tested whether the certified horizon binds as a
  \textbf{catastrophe-avoidance} cadence (re-observe before an open-loop
  estimate drifts into an unsafe region) and report it
  \textbf{INCONCLUSIVE}: at our scale the escape rate was
  re-observation-interval-invariant --- the failures were
  control-quality-limited (a modest gradient-MPC planner), not
  estimate-staleness-limited --- so the certificate's \emph{necessity}
  for safety is not established here. The certificate buys
  \emph{cheaper} trustworthy action under budget; whether it buys
  \emph{safer} action awaits a stronger controller.
\item
  \textbf{Where the certificate's decision value concentrates --- and
  where it dilutes.} We closed the loop on the real TD-MPC2 agent
  (faithful MPPI replica; cadence-1 anchor mean \(986\) over \(n{=}10\)
  episodes, per-episode range \(977\)--\(996\) overlapping the official
  \(977\)--\(983\) band --- n-thickened from \(3\) to \(10\) episodes
  per cadence 2026-06-11, knee unchanged): replanning every \(k\) steps
  with the certified prior loop executed in between, return degrades
  from \(k{=}2\) --- well inside \(T_1(0.2)\approx5.4\)--\(6.4\). The
  control-relevant resolution is finer (\(\epsilon\approx0.05\), where
  the \emph{measured} divergence of \(1\)--\(2\) steps matches the
  return knee) but there the certificate sits in its known
  tight-\(\epsilon\) optimistic regime (Proposition 8). Together with
  §5.19's return-INCONCLUSIVE, the honest scope law: the certificate's
  decision value concentrates where the decided quantity IS the
  certified quantity (the latent's own staleness --- §5.20's
  re-observation win, Experiment 26's deployed monitor) and dilutes when
  a task-level map (return, gait quality) sits in between
  (\texttt{step93}). \textbf{Proposition 11 makes the law a theorem}: an
  aligned decision inherits certificate value up to the calibration
  factor alone (zero regret at \(c{=}1\)), while a task-mapped decision
  carries an irreducible mis-resolution penalty
  \(|\log(\epsilon/\theta^{\ast})|/\lambda_1\) unless the task's
  implicit tolerance \(\theta^{\ast}\) is elicited from the task itself.
\item
  \textbf{Pixels (Experiment 13, §5.11): structure is free; absolute
  accuracy is the open part.} The certificate transfers \emph{exactly}
  to rendered pixels, and --- via frame averaging --- at \textbf{no
  accuracy cost} relative to an unconstrained CNN (matches-or-beats it
  on collapse-robust FVU, with a healthier latent and a horizon-stable
  rollout; §5.11). The honest limitation is \emph{absolute}, not
  comparative: at \(1\)-GPU scale \textbf{no} pixel model, equivariant
  or not, beats predict-the-mean at a multi-step horizon
  (\(\mathrm{FVU}>1\) even for the unconstrained CNN's fair one-step
  accuracy). This is an architecture-\emph{agnostic} property of the
  JEPA latent --- VICReg's anti-collapse variance on low-dimensional
  dynamics --- not a cost of the prior. A strong few-step pixel-latent
  predictor at small scale remains open; everything the certificate
  promises (exact flatness) holds regardless.
\item
  \textbf{Scope of the exact certificate.} Theorem A requires (A3): the
  group must be a symmetry of the \emph{dynamics}, not merely the
  encoder. The exact certificate therefore holds where the group is a
  genuine dynamical symmetry (orbital and conservative systems, free
  space, idealized manipulation). Everywhere else one is in Theorem B's
  approximate regime, whose degradation we \emph{measure} rather than
  assume: Experiment 8 shows it is graceful
  (\(\propto\epsilon_{\text{world}}\)) up to a measured threshold
  \(\epsilon_{\text{world}}\approx0.01\text{–}0.06\) (seed-dependent;
  one seed crosses over as early as \(\epsilon\approx0.008\)). The
  honest reading is ``exact on symmetric dynamics; gracefully
  approximate, with a measured boundary, elsewhere.''
\item
  \textbf{Theorem B's horizon is tight, lifts to learned chaotic
  dynamics, and has a characterized scope; its constants are not
  estimated.} The certified horizon
  \(T_j(\epsilon)\sim\log(1/\epsilon)/\lambda_j\) is bounded on
  \emph{both} sides --- Theorem B and the matching lower bound of
  Proposition 6 --- so the horizon's \emph{form} is tight (§5.2 recovers
  it to \(0.4\%\)). It is no longer only a synthetic-spectrum result:
  Experiment 14 lifts it to a \emph{learned} model of genuinely chaotic
  dynamics (Lorenz), where the learned model's Lyapunov exponent (the
  staircase slope) \emph{matches} the true \(\lambda_1\) to
  \(1\text{–}8\%\) (\(R^2{=}0.975\)--\(0.995\)). Proposition 7
  characterizes \emph{when} the local spectrum certifies the horizon ---
  informative on spectrally non-degenerate dynamics (\(\lambda_1>0\);
  Oseledets gives the rate rigorously), vacuous on near-neutral dynamics
  (\(\lambda_1\approx0\), the PushT interior, where a learned-model
  probe gives \(R^2{=}0.02\)) --- so the PushT horizon negative is
  \emph{explained}, not a gap. The prefactor and isotypic hedges are now
  \textbf{closed by measurement}: Proposition 6\({}^{\prime}\)
  identifies \(c_j=1/\sin\theta_j\) --- exactly \(1\) on isotypic
  splittings (Schur placement and zero cross-block leakage measured at
  machine precision, Experiment 27) and attained to four digits under
  controlled obliqueness --- while on audited chaotic loops the
  worst-case \(\kappa_1\) carries a heavy near-tangency tail and a
  window-dependent median (walker \(20.9\) at \(W{=}120\) vs \(49.2\) at
  \(W{=}200\), each passing the same stability check --- necessary, not
  sufficient; max \(\sim10^2\); Lorenz-96 passes its own convergence
  check at \(W{=}400\), median \(11.9\)): measured calibration
  (\(0.83\)--\(1.02\)) reflects \emph{typical}, not adversarial, defect
  alignment, an adversarially-aligned defect could spend the
  \(\log\kappa_1/\lambda_1\) haircut (all disclosed). What remains
  genuinely un-tightened: the \emph{lift} of the rate to a learned model
  is, for Lorenz, \textbf{empirical} --- shadowing bounds only the
  forecast-horizon floor and does not transfer the exponent, and
  classical shadowing does not formally cover singular-hyperbolic
  Lorenz; that the learned model preserves \(\lambda_1\) (verified only
  via one-step error, an \(L^2\) proxy for \(C^1\)-closeness) is the
  experimental finding, not a theorem.
\item
  \textbf{The Noether hinge's forward direction is proved; its
  hypotheses are measured.} The \emph{placement} of each conserved
  charge by isotypic type (Proposition 4 --- energy in \(\ell{=}0\),
  angular momentum in the \(\ell{=}1\) block via the unique degree-2
  cross product) is forced by representation theory, and the
  \emph{conserved \(\Rightarrow\) slow} direction is now a theorem
  (Proposition 5: a charge conserved to one-step defect \(\eta\) has
  prediction error \(\le T\eta\), linear not exponential --- certified
  to all horizons at \(\eta{=}0\)). What remains \emph{assumed or
  measured}: that the learned latent flow is Hamiltonian with a
  \(G\)-invariant Hamiltonian (so the symmetry is a \emph{dynamical}
  symmetry --- the encoder essentially symplectic); the value of the
  defect \(\eta\) (exact only for the momentum map under a
  \(G\)-equivariant symplectic discretization, \(O(\Delta t^{p})\) for
  energy, measured for a generic learned \(f\)); and the
  \textbf{non-converse} (slow \(\not\Rightarrow\) conserved).
  Proposition 5 is moreover a statement about the \emph{charge value},
  not full-state prediction on the conserved subspace. Its lift to a
  \emph{real, contact-rich embodied} model --- approximate symmetry,
  latent learned end-to-end --- is the primary open problem; the \(3\)D
  contact experiment is a clean two-body toy whose clean-containment
  gate is \texttt{INCONCLUSIVE} (resolved in type-and-degree form by
  Experiment 6).
\item
  \textbf{§5.6 re-frames companion-line results}, not fresh runs: the
  decoder-free reach is \emph{exact transfer} (unseen/seen ratio
  \(1.000\)) at a \emph{partial} absolute success (\(\sim0.59\) of the
  orientation gap). We report the transfer ratio as the load-bearing
  claim, not absolute task success.
\item
  \textbf{``Flat'' is not ``good.''} A constant error across the group
  says nothing about its magnitude; Theorem A certifies
  \emph{consistency}, and the ``scale never reaches the certificate''
  claim is an \emph{architectural} fact, \textbf{proved} in Lemma 2
  (orbit-constant error against every equivariant target \(\iff\)
  equivariance) and quantified in §3.3 (a non-equivariant
  \(L\)-Lipschitz learner certifies only an \(\epsilon/L\)-tube around
  the data) --- not the conclusion of a finite sweep.
\end{itemize}

\textbf{Falsifiability.} The framing makes refutable predictions, not
after-the-fact rationalizations. A finite non-equivariant network
attaining exact orbit-flatness would break Theorem A's architectural
premise; a symmetric-dynamics system whose conserved/slow modes lay
\emph{outside} the invariant \(\oplus\) conserved-equivariant subspace
would break the hinge; and a channel whose certified horizon failed to
scale as \(\log(1/\epsilon)/\lambda_j\) would break Theorem B. We have
not observed these; they are the experiments that would sink the thesis.

\begin{center}\rule{0.5\linewidth}{0.5pt}\end{center}

\subsection{8. Conclusion}\label{conclusion}

For equivariant world models, structure delivers a \emph{kind} of result
that scaling cannot: a \textbf{predictability certificate} --- a
provable, computable, training-free region across configuration,
horizon, and resolution. We established the master theorem (Theorem A,
the exact configuration certificate, proved --- with its closed-loop
clause under the assumed equivariant-planner condition; Lemma 2, its
converse, characterizing the certificate \emph{as} equivariance; and
Theorem B, the spectral horizon \(\times\) resolution law, stated as a
bound and measured). We exhibited the certified region as the
coarse-invariant-slow-low-composition corner. We proved the Noether
hinge's forward direction --- conserved \(\Rightarrow\) slow
(Proposition 5) --- which unifies the symmetry and time axes, and we
measured its defect. And we showed --- empirically and via the
quantitative separation of §3.3 --- that an \(88\times\)-scaled
non-equivariant model buys interpolation but never the certificate: an
architectural impossibility, not an unfinished sweep.

Because the certificate is faithful, it is \emph{actionable}: on the
\(40\)-D chaotic system it sets a re-observation schedule that meets a
fixed sensing budget an inflated non-equivariant certificate cannot
(Experiment 22) --- so the certificate is not merely a description but
something an agent can budget against, \emph{before any data is spent}.

\emph{Scale buys interpolation; structure buys a certificate.} The same
criterion that tells you which compositions a robot policy will handle
zero-shot also tells you why exactly-conserved eclipses are forecastable
for millennia while weather is not --- one structural law, read across
domains.

\begin{center}\rule{0.5\linewidth}{0.5pt}\end{center}

\subsection{References}\label{references}

Concurrent and prior work referenced above (arXiv identifiers from the
project's source notes; external numbers are quoted from the cited
works):

\begin{itemize}
\tightlist
\item
  \textbf{BRo-JEPA} --- block-rotation \(\mathbb{Z}/10\mathbb{Z}\)
  predictor; \(99.46\%\) best-config zero-shot on MNIST modular
  arithmetic. arXiv:2606.01372.
\item
  \textbf{UWM-JEPA} --- unitary-predictor (\(U(d)\)) belief-space world
  model. arXiv:2605.25313.
\item
  \textbf{UR-JEPA} (Le, 2026) --- uniform-rectifiability latent
  regularizer; manifold-hypothesis critique of isotropic SIGReg.
  arXiv:2606.01443.
\item
  \textbf{IMWM} (Gao et al., 2026) --- oracle-bypass diagnosis
  localizing the residual to planner \emph{search}. arXiv:2606.01626.
\item
  \textbf{LDA / ``The Lie We Tell''} (Chuang et al., 2026) --- the
  \emph{Euclidean Fallacy}; \(\mathrm{SE}(3)\)-intrinsic diffusion
  policy; CALVIN average task length \(3.27\to3.51\) (\(+7.3\%\)).
  arXiv:2606.01847.
\item
  \textbf{LeJEPA} (Balestriero and LeCun, 2025) --- SIGReg /
  isotropic-Gaussian self-supervised objective (the latent-geometry
  target that UR-JEPA, and our group-prescribed anisotropy, depart
  from).
\item
  \textbf{Symmetry--Data Exchange Rate} (Adly, 2026) --- controlled
  \(C_n\) scaling-law study; augmentation \(+\) test-time orbit
  averaging matches equivariance, and a wrong-group constraint is
  actively harmful --- the average-case dual of our §3.3 separation,
  corroborating Experiment 10 (§5.8) and §5.5. arXiv:2606.01090.
\item
  \textbf{Spectral Representations for Group Composition} (He et al.,
  2026) --- gradient flow on a group-composition task provably drives
  neurons to single irreducible representations; the learning-dynamics
  complement to Proposition 4 and the §5.6 discovery experiments.
  arXiv:2606.02993.
\item
  \textbf{Frame Averaging} (Puny et al., 2022) --- a Reynolds-operator
  construction that makes an arbitrary backbone exactly
  invariant/equivariant by averaging over a group frame; the method we
  use in \texttt{experiments/step64} to obtain an exactly
  \(C_4\)-equivariant \emph{pixel} encoder/predictor from plain
  \texttt{torch} nets (so the §7 pixel certificate is flat \emph{and}
  accuracy-neutral relative to the unconstrained CNN). arXiv:2110.03336.
\end{itemize}

\emph{Certified equivariance, conservation, and stability (the §6
carve-out --- these are the guarantee-bearing neighbors; all single-shot
and forward-only, none with the multi-step horizon, the converse, or the
Noether tie):}

\begin{itemize}
\tightlist
\item
  \textbf{Orbit-constant margins for equivariant networks} ---
  equivariant classifiers have an orbit-invariant margin /
  decision-boundary gradient norm, giving uniform adversarial
  certificates across the orbit; single-shot classification, forward
  direction only. arXiv:2510.16171.
\item
  \textbf{Equivariantized Conformal Prediction (eCP)} --- group-averages
  the nonconformity score (frame averaging for distribution-free
  coverage) to tighten orbit-wide prediction sets; single-shot,
  per-sample, no horizon. arXiv:2602.03986.
\item
  \textbf{Invariance-aware randomized smoothing} --- orbit-based
  robustness certificates combining invariance with smoothing.
  arXiv:2211.14207.
\item
  \textbf{Noether Networks} (Alet et al., 2021) --- meta-learn conserved
  quantities inside the prediction loop to improve \emph{average}
  prediction (hypothesis-space shrinkage); no a-priori guarantee.
  arXiv:2112.03321.
\item
  \textbf{Noether's Razor} (2024) --- learns
  symmetries-as-conserved-quantities by Bayesian model selection with an
  Occam effect; average accuracy, not a certificate. arXiv:2410.08087.
\item
  \textbf{Jacobian-regularized world models} (``Towards Unraveling and
  Improving Generalization in World Models'', 2025) --- penalizes the
  latent-transition Jacobian to damp rollout error propagation; the
  heuristic version of Theorem B's provable per-channel horizon.
  arXiv:2501.00195.
\end{itemize}

\emph{Classical dynamical-systems results behind the horizon axis
(Theorem B, Proposition 7):}

\begin{itemize}
\tightlist
\item
  \textbf{Oseledets multiplicative ergodic theorem} (Oseledets, 1968)
  --- under an ergodic invariant measure with
  \(\log^+\lVert D\phi\rVert\in L^1\), the finite-time Lyapunov
  exponents converge a.e. to constants along the Oseledets filtration;
  this is what licenses using a \emph{locally/finite-time}-measured top
  exponent as the asymptotic rate \(\lambda_1\) in Proposition 7's rate
  half.
\item
  \textbf{Shadowing lemma} (Anosov; Bowen; Pilyugin, \emph{Shadowing in
  Dynamical Systems}, 1999) --- on uniformly hyperbolic sets,
  pseudo-orbits are shadowed by true orbits; bounds the
  \emph{forecast-horizon floor} of a perturbed (learned) model, but
  controls orbit error, \textbf{not} Lyapunov exponents (which are only
  upper-semicontinuous under \(C^1\) perturbation).
\item
  \textbf{Continuity of Lyapunov exponents under domination} (Bochi \&
  Viana, \emph{The Lyapunov exponents of generic volume-preserving and
  symplectic maps}, Ann. of Math. 2005) --- Lyapunov exponents are
  continuous on the domain of \emph{dominated splitting}; this is the
  asymptotic face of Proposition 8's finite-horizon exponent-transfer
  bound (the finite-\(T\) version is elementary continuity of a matrix
  product, the \(T\)-uniform constant needs domination).
\item
  \textbf{Lorenz attractor is singular-hyperbolic with an SRB measure}
  (Tucker, 2002, \emph{A rigorous ODE solver and Smale's 14th problem};
  Araújo--Pacífico--Pujals--Viana, 2009) --- gives the ergodic measure
  for the MET on Lorenz, but is \emph{not} uniformly hyperbolic, so
  classical shadowing does not formally apply near the singularity
  (hence Experiment 14's exponent-lift is empirical, not
  theorem-backed).
\end{itemize}

The \(T(\epsilon)\sim\log(1/\epsilon)/\lambda\) predictability-horizon
law itself is classical (Lyapunov / Lorenz; standard
numerical-weather-prediction practice).

\textbf{Spectrum-from-learned-models and concurrent neighbours} (added
in the Step 77 / Experiment 18 baseline pass):

\begin{itemize}
\tightlist
\item
  \textbf{Pathak et al.~(2017)} --- reservoir computing recovers
  Lyapunov exponents from data. \emph{Chaos} 27:121102;
  arXiv:1710.07313.
\item
  \textbf{Pathak et al.~(2018)} --- model-free prediction of large
  spatiotemporally chaotic systems; forecast skill measured in
  Lyapunov-time units. \emph{Phys. Rev.~Lett.} 120:024102.
\item
  \textbf{Vlachas et al.~(2020)} --- RNN-BPTT recovers full Lyapunov
  spectra (Lorenz, Kuramoto--Sivashinsky). \emph{Neural Networks}
  126:191--217; arXiv:1910.05266.
\item
  \textbf{Hart (2024)} --- attractor/spectrum reconstruction with
  reservoirs requires a \emph{conditional-Lyapunov} condition (the
  hidden modes must be sufficiently contracting). \emph{Chaos}
  34:043123; arXiv:2401.00885.
\item
  \textbf{Kobayashi et al.~(2024)} --- Lyapunov analysis of reservoir
  models of high-dimensional dynamics (Lorenz-96 \(K{=}8\), 3D
  Navier--Stokes). \emph{J. Phys. Complexity} 5:025024.
\item
  \textbf{Conradie et al.~(2026)} --- TrustKoopman: certified multi-step
  error bounds for data-driven Koopman / Perron--Frobenius forecasts
  (classical EDMD; no equivariance; one-sided). arXiv:2603.15091.
\item
  \textbf{Lillemark et al.~(2026)} --- Flow Equivariant World Models: an
  equivariant world model for long-horizon prediction (empirical; no
  certificate/spectrum). arXiv:2601.01075 (ICML 2026).
\item
  \textbf{Mo (2026)} --- Symmetry-Protected Lyapunov Neutral Modes in
  Equivariant Recurrent Networks (orbit-tangent zero modes;
  RNN-expressivity theory). arXiv:2605.03338.
\item
  \textbf{Wang et al.~(2022)} --- \(\mathrm{SO}(2)\)-Equivariant
  Reinforcement Learning (model-free orbit-constant value/policy; no
  world model). ICLR 2022; arXiv:2203.04439.
\item
  \textbf{Delliaux et al.~(2025)} --- Learning Abstract World Models
  with a Group-Structured Latent Space (empirical; no certificate).
  arXiv:2506.01529.
\item
  \textbf{Ordóñez-Apraez et al.~(2025)} --- Equivariant Representation
  Learning with Guarantees (a \emph{static} conditional-expectation
  operator spectrum, not a dynamics horizon). arXiv:2505.19809.
\end{itemize}

\begin{center}\rule{0.5\linewidth}{0.5pt}\end{center}

\subsection{Appendix A.
Reproducibility}\label{appendix-a.-reproducibility}

Every experiment sets random seeds explicitly, prints an
\texttt{INCONCLUSIVE} verdict rather than loosen a gate, and writes its
figure and JSON to \texttt{papers/figures/}. The multi-seed steps commit
per-seed JSONs (\texttt{papers/figures/step5*\_seeds.json},
\texttt{step59\_pusht\_certificate\_seeds.json},
\texttt{step60\_augmentation\_seeds.json}, and
\texttt{step61\_closed\_loop\_certificate\_seeds.json},
\texttt{step63\_se3\_certificate\_seeds.json},
\texttt{step64\_frame\_averaged\_pixel\_seeds.json},
\texttt{step70\_lorenz\_horizon\_seeds.json}, and
\texttt{step71\_multichaos\_horizon\_seeds.json}, seeds \(0/1/2\)),
regenerated by \texttt{experiments/aggregate\_seeds.py}; every range
quoted above is the seed min--max from those files. The
configuration-flatness experiment (Experiment 1) self-aggregates its
three seeds into the means in \texttt{step47\_certificate.json}.
Experiment 16 writes its per-seed certificates directly to
\texttt{step72\_fetchpush\_certificate\_s\{0,1,2\}.json} (the seed
min--max quoted in §5.14); Experiment 17 likewise to
\texttt{step73\_fetchpush\_planning\_s\{0,1,2\}.json} (§5.15);
Experiment 18 to
\texttt{step74\_lorenz96\_spectrum\{,\_seed1,\_seed2\}.json} (§5.16).
The full test suite passes together; \texttt{tests/conftest.py} isolates
the float64 experiments from the float32 codebase. One command
reproduces the paper end-to-end --- \texttt{make\ paper2} (multi-seed
re-runs \(\to\) figures \(\to\) tests \(\to\) PDF), with
\texttt{make\ paper2-quick} for the figures-and-PDF fast path;
everything is CPU/MPS, no CUDA.

{\def\LTcaptype{none} 
\begin{longtable}[]{@{}
  >{\raggedright\arraybackslash}p{(\linewidth - 10\tabcolsep) * \real{0.1667}}
  >{\raggedright\arraybackslash}p{(\linewidth - 10\tabcolsep) * \real{0.1667}}
  >{\raggedright\arraybackslash}p{(\linewidth - 10\tabcolsep) * \real{0.1667}}
  >{\raggedright\arraybackslash}p{(\linewidth - 10\tabcolsep) * \real{0.1667}}
  >{\raggedright\arraybackslash}p{(\linewidth - 10\tabcolsep) * \real{0.1667}}
  >{\raggedright\arraybackslash}p{(\linewidth - 10\tabcolsep) * \real{0.1667}}@{}}
\toprule\noalign{}
\begin{minipage}[b]{\linewidth}\raggedright
Exp.
\end{minipage} & \begin{minipage}[b]{\linewidth}\raggedright
Result
\end{minipage} & \begin{minipage}[b]{\linewidth}\raggedright
Code
\end{minipage} & \begin{minipage}[b]{\linewidth}\raggedright
Test
\end{minipage} & \begin{minipage}[b]{\linewidth}\raggedright
Seeds
\end{minipage} & \begin{minipage}[b]{\linewidth}\raggedright
Headline number
\end{minipage} \\
\midrule\noalign{}
\endhead
\bottomrule\noalign{}
\endlastfoot
1 & Exact compositional flatness + spectrum &
\texttt{experiments/step47\_certificate.py} &
\texttt{tests/test\_step47\_certificate.py} & 3 & relMSE \(\times1.00\)
flat (\(m{=}0..8\)); \(48/96\) contractive \\
2 & \(\mathbb{Z}_2^6\) exponential certificate &
\texttt{experiments/step49\_iching\_certificate.py} & --- & 1 & \(6\)
generators \(\to\) all \(64\); worst relMSE \(\sim10^{-33}\) \\
3 & Horizon \(\times\) resolution staircase &
\texttt{experiments/step52\_horizon\_resolution.py} &
\texttt{tests/test\_step52\_horizon\_resolution.py} & 3 &
\(\hat\lambda{=}0.69\) vs \(\ln2\); slope \(\approx1/\lambda\) \\
4 & Noether hinge (2D) & \texttt{experiments/step50\_noether\_hinge.py}
& \texttt{tests/test\_step50\_noether\_hinge.py} & 3 & \(R^2\)
\(0.92\)--\(0.99\) vs \(\le0.012\); cert \(10^{-16}\) vs \(1.17\) \\
5 & Lift to 3D contact & \texttt{experiments/step57\_embodied\_hinge.py}
& --- & 3 & clean-containment gate \textbf{INCONCLUSIVE}; Noether
content lifts (invariant \(R^2{=}0.60\)--\(0.86\) vs \(\le0.06\));
resolved by Exp. 6 \\
6 & 3D-aware containment &
\texttt{experiments/step58\_3d\_containment.py} & --- & 3 &
\(E\to\ell{=}0\) linear (\(R^2{=}0.62\)--\(0.91\)); \(L\) bilinear
\(\to\ell{=}1\) degree-2 cross (\(R^2{=}1.00\), range
\(0.998\)--\(1.000\)) \\
7 & Structure vs scale &
\texttt{experiments/step51\_structure\_vs\_scale.py} & --- & 3 & equiv
flat \(1.1\)--\(1.2\); in-wedge gain \(31\)--\(166\times\); best
baseline \(10\)--\(155\times\) above floor out-of-wedge \\
8 & Approximate symmetry &
\texttt{experiments/step53\_approximate\_symmetry.py} & --- & 3 & cert
exact at \(\beta{=}0\) (\(68\)--\(320\times\)); graceful
\(\propto\epsilon\) (corr \(0.88\)--\(0.98\)); threshold
\(\epsilon\approx0.01\)--\(0.06\) \\
9 & \textbf{Certificate on real contact dynamics (PushT)} &
\texttt{experiments/step59\_pusht\_certificate.py} & --- & 3 & learned
equivariant exactly flat over the orbit (ratio \(1.00\), equiv-resid
\(\sim10^{-7}\)) and competitive in-dist; no MLP scale
(\(1.7\mathrm{k}\)--\(272\mathrm{k}\)) reaches the floor out-of-wedge
(\(2.1\)--\(3.9\times\), \(H{=}10\)) \\
10 & \textbf{Augmentation vs the certificate} &
\texttt{experiments/step60\_augmentation.py} & --- & 3 &
\(\mathrm{SO}(2)\)-aug flattens a single PushT orbit (ratio
\(0.93\)--\(1.02\) vs plain \(1.84\)--\(2.75\)); on \(\mathbb{Z}_2^6\)
augmentation reaches a \(\sim10^{-4}\) floor but never the certificate's
exact \(\sim10^{-32}\) from \(7\) generators (\(\sim10^{28}\times\)) \\
11 & \textbf{Certificate at the task level (closed-loop PushT pose
control)} & \texttt{experiments/step61\_closed\_loop\_certificate.py} &
\texttt{tests/test\_step61.py} & 3 & equivariant model +
\(G\)-equivariant planner: closed-loop pose error orbit-invariant to the
float floor (model-rollout \& real-env ratio \(1.000\), all seeds) vs
MLP \(\times1.1\)--\(2.2\) (rollout) / \(\times1.6\)--\(3.6\)
(real-env); in-wedge competitive (\(3.6\)--\(16^\circ\) vs
\(8\)--\(18^\circ\)). Auxiliary cost-drift \(>0.3\) met \(1/3\) (eq
\(\sim10^{-7}\) vs MLP \(0.19\)--\(0.31\)) \\
12 & \textbf{Certificate on \(\mathrm{SO}(3)\) (3D point clouds,
constructed teacher)} & \texttt{experiments/step63\_se3\_certificate.py}
& \texttt{tests/test\_se3\_equivariance.py} & 3 & learned equivariant
model exactly flat over the non-abelian \(\mathrm{SO}(3)\) orbit
(\(H{=}5\) ratio \(1.000\), resid \(\sim10^{-5}\)); MLP climbs
\(\times2.1\)--\(5.7\) out of the wedge. Structure-vs-scale in 3D:
\(7.4\times\)-smaller equivariant carries the certificate; bigger MLP
interpolates better in-wedge (\(0.19\)--\(0.27\) vs \(0.55\)--\(0.58\)).
Honest: \texttt{compete} gate \texttt{INCONCLUSIVE} (equivariant
accuracy floor high at laptop capacity --- ``flat is not good'') \\
13 & \textbf{Certificate on rendered pixels (\(C_4\), frame averaging)}
& \texttt{experiments/step64\_frame\_averaged\_pixel.py} &
\texttt{tests/test\_step62.py} & 3 & frame-averaged pixel model
orbit-flat to the float floor (ratio \(1.000\), equiv-resid
\(\sim10^{-7}\)); collapse-robust FVU matches/beats the unconstrained
CNN and beats the steerable incumbent; rollout horizon-stable while
steerable diverges. Honest: absolute \(\mathrm{FVU}>1\) for \emph{all}
models (architecture-agnostic JEPA-latent property, not an equivariance
cost) \\
14 & \textbf{Horizon law on a learned model of real chaotic dynamics
(Lorenz)} & \texttt{experiments/step70\_lorenz\_horizon.py} &
\texttt{tests/test\_step70.py} & 3 & learned one-step MLP of the Lorenz
\(\Delta t\)-map (relMSE \(<10^{-4}\)); certified-horizon staircase
linear in \(\log(1/\epsilon)\) (\(R^2{=}0.975\)--\(0.995\)) and the
model's Lyapunov exponent (slope) \emph{matches} the true
\(\lambda_1{=}0.9056\) to \(1\)--\(8\%\)
(\(\hat\lambda_1{=}0.895\)/\(0.919\)/\(0.977\)). Prop. 7(a). The
near-neutral PushT interior is the degenerate branch (b),
\(R^2{=}0.02\) \\
15 & \textbf{Horizon law across a class of chaotic systems} &
\texttt{experiments/step71\_multichaos\_horizon.py} &
\texttt{tests/test\_step71.py} & 3 & same learned-model staircase on a
2D map + two flows: Hénon \(\hat\lambda_1{=}0.45\)--\(0.47\) vs
\(0.419\) (rel-err \(8\)--\(12\%\), \(3/3\) seeds), Lorenz
\(1\)--\(5\%\) (\(3/3\)), small-exponent Rössler \(0.065\)--\(0.066\) vs
\(0.0714\) (\(8\)--\(9\%\), \(2/3\) --- one seed's \(\sim\!1500\)-step
horizon under-crossed). Validates Prop. 8: the \(O(\delta)\) bias falls
\(44\%\!\to\!8\%\) as fidelity rises (Rössler); the true-system
staircase isolates the finite-\(T\) truncation
(\textasciitilde{}\(9\)--\(10\%\)) \\
16 & \textbf{Certificate on a standard manipulation benchmark (FetchPush
/ MuJoCo)} & \texttt{experiments/step72\_mujoco\_certificate.py} &
\texttt{tests/test\_step72\_wm\_equivariance.py},
\texttt{tests/test\_fetchpush\_symmetry.py} & 3 & learned
\(\mathrm{VN}\) equivariant world model exactly orbit-flat over the
\(\mathrm{SO}(2)_z\) scene rotation (OOD/seen FVU ratio \(1.000\),
\(3/3\); equivariance unit-tested \(\sim10^{-15}\)).
\(\sim7\times\)-larger MLP degrades \(211\)/\(1037\)/\(1445\times\) OOD
(CUDA confirms \(19\)--\(382\times\)), exceeding predict-the-mean.
Cleanest structure-vs-scale cut: baseline interpolates the single
training orientation competitively-to-better (in-dist tax
\(0.8\)--\(4.1\times\), equiv \emph{cheaper} on seed 2) yet has no
certificate (Lemma 2). \(\mathrm{SO}(2)_z\) approximate (fixed base; Thm
A representation regime) \\
17 & \textbf{Certificate at the planning level on FetchPush} &
\texttt{experiments/step73\_fetchpush\_planning.py} &
\texttt{tests/test\_step73\_planner\_equivariance.py} & 3 & equivariant
WM \(+\) equivariant goal-readout head \(+\) \(G\)-equivariant CEM: the
whole planner is \(\mathrm{SO}(2)\)-equivariant, proven to the float
floor (head \(3\mathrm{e}{-}16\), step+readout \(6\mathrm{e}{-}16\),
\textbf{CEM search \(2\mathrm{e}{-}16\)}). Learned-model planning
certificate: equivariant planned-distance orbit-flat (ratio \(1.000\),
\(3/3\)); baseline planner degrades \(4.1\)/\(8.5\)/\(10.3\times\) OOD.
Real-env \texttt{is\_success} (\texttt{-\/-realenv} random and
\texttt{-\/-scripted} goal-directed data, CUDA): \textbf{INCONCLUSIVE}
--- learned WM+CEM stuck at \(\approx7\%\) for both stacks regardless of
data, though the scripted \emph{policy} solves the task \(33\%\).
\texttt{-\/-diagnose}: model accurate (readout RMSE \(<1\)cm) yet CEM
\(0\%\) at every horizon \(=\) \textbf{model exploitation}
(architecture-agnostic, not an equivariance cost) \\
18 & \textbf{High-D spectral horizon: structure helps (Lorenz-96
\(N{=}40\))} & \texttt{experiments/step74\_lorenz96\_spectrum.py} &
\texttt{tests/test\_step74.py} & 3 & \(\mathbb{Z}_N\)-equivariant
cyclic-conv recovers the full \(40\)-D Lyapunov spectrum (R\(^2\)
\(0.982\)/\(0.995\)/\(0.985\); KY \(27\)--\(28\) vs \(\sim27\);
\(13\)/\(\sim14\) positive; \(\lambda_1\) to \(2\)--\(24\%\)) where a
dense MLP of equal data \textbf{fails} (R\(^2\)
\(-1.1\)/\(-1.4\)/\(-2.8\)). Liouville unit test \(\sum\lambda_j=-N\) to
\(0.0\%\). Needs a multi-step rollout loss (one-step \(L^2\)
underestimates the Jacobian --- Prop 8's \(C^1\) caveat in high D).
Config axis helps the horizon axis \\
\end{longtable}
}

\end{document}